\documentclass{article}

\usepackage{natbib}
\usepackage{fullpage}
\usepackage{mathpazo}

\usepackage{dsfont}
\usepackage{graphicx}
\usepackage{xcolor}
\usepackage{amsmath}
\usepackage{amsfonts}

\usepackage[textsize=scriptsize]{todonotes}

\DeclareMathOperator*{\argmin}{arg\,min}

\usepackage[utf8]{inputenc} 
\usepackage[T1]{fontenc}    
\usepackage{hyperref}       
\usepackage{url}            
\usepackage{booktabs}       
\usepackage{amsfonts}       
\usepackage{nicefrac}       
\usepackage{microtype}      
\usepackage{xcolor}         
\usepackage{subcaption}

\title{What Can Transformers Learn In-Context? \\
A Case Study of Simple Function Classes}

\def\AND{%
        \end{tabular}\hfil\linebreak[4]\hfil%
        \begin{tabular}[t]{c}\ignorespaces%
}

\date{}
\author{
  Shivam Garg\thanks{Equal contribution.} \\
  Stanford University\\
  \texttt{shivamg@cs.stanford.edu} \\
  \and
  Dimitris Tsipras\footnotemark[1] \\
  Stanford University\\
  \texttt{tsipras@stanford.edu} \\ 
  \AND
  Percy Liang\\
  Stanford University\\
  \texttt{pliang@cs.stanford.edu} \\ 
  \and
  Gregory Valiant \\
  Stanford University\\
  \texttt{valiant@stanford.edu}
}

\begin{document}

\maketitle

\begin{abstract}
 In-context learning refers to the ability of a model to condition on a prompt sequence consisting of in-context examples (input-output pairs corresponding to some task) along with a new query input, and generate the corresponding output.
Crucially, in-context learning happens only at inference time without any parameter updates to the model. 
While large language models such as GPT-3 exhibit some ability to perform in-context learning, it is unclear what the relationship is between tasks on which this succeeds and what is present in the training data.
To make progress towards understanding in-context learning, we consider the well-defined problem of training a model to in-context learn a function class (e.g., linear functions): that is, given data derived from some functions in the class, can we train a model to in-context learn ``most'' functions from this class?
We show empirically that standard Transformers can be trained from scratch to perform in-context learning of linear functions---that is, the trained model is able to learn unseen linear functions from in-context examples with performance comparable to the optimal least squares estimator.
In fact, in-context learning is possible even under two forms of distribution shift: (i) between the training data of the model and inference-time prompts, and (ii) between the in-context examples and the query input during inference.
We also show that we can train Transformers to in-context learn more complex function classes---namely sparse linear functions, two-layer neural networks, and decision trees---with performance that matches or exceeds 
task-specific learning algorithms.
\footnote{Our code and models are available at \url{https://github.com/dtsip/in-context-learning}.}
\end{abstract}

\section{Introduction}
Large language models such as GPT-3~\citep{brown2020language} are able to perform \emph{in-context learning}:
given a prompt containing examples from a task (input-output pairs) and a new query input, the language model can generate the corresponding output.
For example, these models are able to produce English translations of French words after being prompted on a few  such translations, e.g.:
\[\underbrace{\ \text{maison} \to \text{house},\ \text{chat} \to \text{cat},\ \text{chien} \to \ }_\text{prompt}
\underbrace{\ \text{dog}\ }_\text{completion}.\]
This capability is quite intriguing as it allows models to adapt to a wide range of downstream tasks on-the-fly---i.e., without the need to perform any parameter updates after the model is trained~\citep{brown2020language,lieber2021jurassic, rae2021scaling, black2022gpt}.
However, it is unclear to what extent these models have developed the ability to learn \emph{new tasks} from in-context examples alone as opposed to simply indexing into a vast set of known tasks from the training data (e.g., see~\citet{min2022rethinking}).
\footnote{The term ``in-context learning'' has also been used to refer to a more general notion of learning from a prompt~\citep{olsson2022context}. In this work, we focus on the standard notion which refers to learning a task/function given in-context examples~\citep{brown2020language}.}

To make progress towards understanding in-context learning, we consider the well-defined problem of learning a \emph{function class} from in-context examples.
That is, we say that a model can in-context learn a function class $\mathcal{F}$ if, for ``most'' functions $f \in \mathcal{F}$, the model can approximate $f(x_{\text{query}})$ for a new query input $x_{\text{query}}$ by conditioning on a prompt sequence $(x_1, f(x_1), \ldots,x_k, f(x_k), x_{\text{query}})$ containing in-context examples and the query input.

Formally, let $D_\mathcal{X}$ be a distribution over inputs and $D_{\mathcal{F}}$ be a distribution over functions in  $\mathcal{F}$. A prompt $P$ is a sequence $(x_1, f(x_1), \ldots, x_k, f(x_k), x_{\text{query}})$ where inputs (i.e., $x_i$ and $x_{\text{query}}$) are drawn i.i.d. from $D_\mathcal{X}$ and $f$ is drawn from $D_{\mathcal{F}}$. We say that a model $M$ can in-context learn the function class $\mathcal{F}$ up to $\epsilon$, with respect to  $(D_{\mathcal{F}}, D_\mathcal{X})$, if it can predict $f(x_\text{query})$ with an average error
\begin{equation}
\label{eq:in-context_loss}
\mathds{E}_{P}\left[\ell\left(M\left(P\right), f\left(x_{\text{query}}\right)\right) \right] \leq \epsilon,
\end{equation}
where $\ell(\cdot, \cdot)$ is some appropriate loss function, such as the squared error. 

Within this framework, we can now concretely ask: 
\begin{center}
    \emph{Can we train a model to in-context learn a certain function class?}
\end{center}
Note that, here, being able to in-context learn a function class is a property of model $M$ alone, independent of how it was trained.
\emph{Training} such a model can be viewed as an instance of \emph{meta-learning}~\citep{schmidhuber1987evolutionary, naik1992meta,thrun2012learning}, a general paradigm for learning a model or method that can learn from data.

\begin{figure}[htb]
    \centering
    \includegraphics[width=\textwidth]{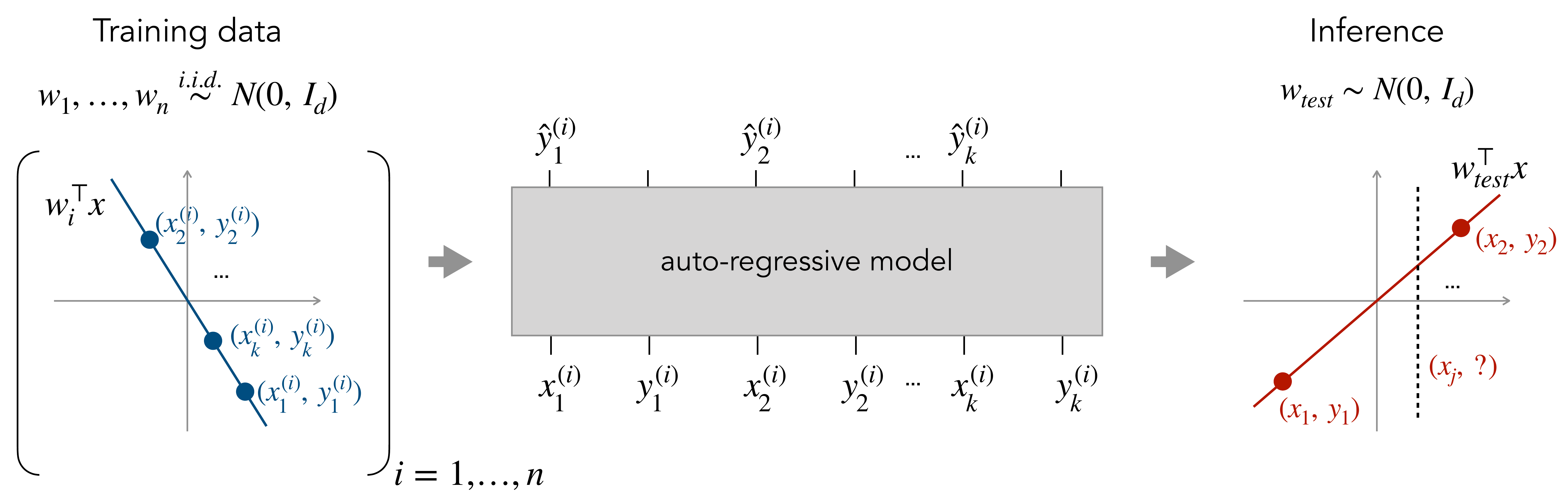}
    \caption{\emph{Can we train a model that in-context learns a function class (here linear functions)?}
    We train Transformers by repeatedly sampling a random function $f$ from that class, as well as random inputs $x_1, \ldots, x_k$ and training the model to predict each $f(x_i)$ given the prompt $x_1, f(x_1), \ldots , x_{i-1}, f(x_{i-1}), x_i$ (wrt squared loss).
    Then, during inference, we evaluate the model's ability to predict accurately on new, \emph{unseen} functions.
    }
    \label{fig:setting}
\end{figure}

We empirically study this question,  focusing on Transformer models~\citep{vaswani2017attention,radford2018improving}---the architecture behind recent large language models---trained from scratch to in-context learn a range of simple, well-defined function classes (e.g. linear functions).
Specifically, we sample prompts containing in-context examples (input-output pairs) generated using functions in a given class and train models to predict the function value at the corresponding query inputs. (see illustration in Figure~\ref{fig:setting}). Our findings are as follows.

\paragraph{Transformers can in-context learn linear functions.}
We show empirically that we can train a standard Transformer from scratch to in-context learn the class of linear functions, with respect to
the input distribution $D_{\mathcal{X}}$ being an isotropic Gaussian in 20 dimensions, and  $D_{\mathcal{F}}$ being the distribution over linear functions with weight vectors drawn from an isotropic Gaussian (the model was trained on prompts generated from the same distributions $D_{\mathcal{X}}$ and $D_{\mathcal{F}}$).
Specifically, the trained model achieves error comparable to the optimal least squares estimator, suggesting that it encodes an effective learning algorithm, at least for the distribution used to generate the training prompts.

\paragraph{Generalization to out-of-distribution prompts.}
To understand the extent to which the trained model encodes an algorithm that works beyond the training distribution, we consider in-context learning under two types of distribution shifts: (a) a shift between the prompts encountered during training and inference (e.g., training on prompts without any noise in the in-context example outputs but testing with noisy outputs), (b) a shift between the in-context examples and the query input during inference (e.g., in-context examples lie in one orthant and the query input lies in another). 
We find that the  performance of our model is quite robust to such shifts, indicating that it has learned to perform linear regression with some generality.

\paragraph{More complex function classes.}
We also consider the function classes of 3-sparse linear functions, two-layer ReLU neural networks with 100 hidden units, and decision trees of depth 4, all with 20 dimensional inputs.
We show that we can again train Transformer models that can in-context learn these classes (with respect to isotropic Gaussian inputs and appropriately defined distributions over functions). 
For sparse linear functions, the trained model is able to exploit sparsity, obtaining performance better than least squares and comparable to Lasso.
For neural networks, the corresponding model is able to obtain performance comparable to neural networks of the same architecture trained using gradient descent on in-context examples. Moreover, it is also able to in-context learn linear functions.
For decision trees, the trained model can learn unseen trees with as few as 100 in-context examples, whereas greedy learning and tree boosting algorithms are unable to achieve competitive performance (for the distribution of prompts studies here).
Note that learning these function classes requires involved algorithms (e.g., gradient descent with the Lasso objective), and our results show that Transformers can encode algorithms with similar performance in a single forward pass.

\paragraph{Role of model capacity and problem dimension.}
Finally, we explore how the ability of Transformers to in-context learn linear functions scales with model capacity and problem dimensionality. We find that increasing the capacity of the model improves performance significantly, and also allows the model to in-context learn higher-dimensional functions.
Moreover, increasing the capacity often significantly improves performance with distribution shifts, even when the absolute improvement in the standard error is small.

\section{Training models for in-context learning}
\label{sec:setting}
We now describe a general methodology for training a model that can in-context learn a function class $\mathcal{F}$ with respect to a distribution $D_{\mathcal{F}}$ over functions, and $D_\mathcal{X}$ over inputs. To do so, we start by constructing random training prompts as follows.
For each prompt, we first sample a random function $f$ from the class according to $D_{\mathcal{F}}$, then create a set of random inputs $x_1, \ldots, x_{k+1}$ drawn independently from  $D_{\mathcal{X}}$, and finally evaluate $f$ on these inputs to produce the prompt
$P = (x_1, f(x_1), \ldots, x_{k+1}, f(x_{k+1}))$.
For example, in the case of linear functions, inputs could be drawn from the isotropic Gaussian distribution $N(0, I_d)$, and a random function chosen by sampling weight vector $w$ from $N(0, I_d)$ and setting $f(x) = w^\top x$.

Now, given such prompts, we train a model to predict $f(x_i)$ for a given $x_i$ based on a set of preceding in-context examples.
Concretely, let $P^{i}$ denote the prompt prefix containing $i$ in-context examples (the first $i$ input-output pairs) and the $(i+1)^{\text{th}}$ input: $P^i = (x_1, f(x_1), x_2, f(x_2), \ldots, x_{i}, f(x_{i}), x_{i+1})$. 
Then, we train a model $M_\theta$ parameterized by $\theta$ aiming to minimize the expected loss over all the prompt prefixes:
\begin{equation}
\min_\theta \ 
\mathds{E}_{P}\left[\frac{1}{k+1}
\sum_{i=0}^k \ell\left(M_\theta\left(P^i\right), f\left(x_{i+1}\right)\right) 
\right],
\label{eq:training_objective}
\end{equation}
where $\ell(\cdot, \cdot)$ is an appropriately chosen loss function.
Below, we describe how this general methodology can be implemented for a concrete model family (see Appendix~\ref{app:setup} for additional details).

\paragraph{Model structure.}
We use a decoder-only Transformer architecture~\citep{vaswani2017attention} from the GPT-2 family~\citep{radford2019language}. 
Our model consists of 12 layers, 8 attention heads, and a 256-dimensional embedding space (9.5M parameters).
This architecture takes as input a sequence of vectors in its embedding space and predicts the next vector in the sequence within the same space (in language modeling, these vectors correspond to input tokens).
We apply this architecture to our prompt format of $(x_1,f(x_1),\ldots,x_{k+1},f(x_{k+1}))$ as follows.
We map each prompt output $f(x_i)$ to the same dimension as prompt inputs $x_i$ by appending zeros, and map the prompt inputs and outputs into the latent embedding space of the Transformer through a (learnable) linear transformation.
We then use another (learnable) linear transformation to map the vector predicted by the model to a scalar.
Note that the Transformer architecture allows us to compute the prediction ($M_\theta(P^i)$) for all prompt prefixes in a single forward pass.

\paragraph{Training.}
We train the model according to the training objective in~\eqref{eq:training_objective} using squared error as the loss function. 
We do so by sampling a batch of random prompts at each training step and then updating the model through a gradient update (we use a batch size of 64 and train for 500k total steps). 
This training is done from scratch, that is, we do \emph{not} fine-tune a pre-trained language model, nor do we train on actual text.

\paragraph{Curriculum learning.}
Many natural function classes contain functions of varying complexity. We exploit this by training our model using a curriculum \citep{bengio2009curriculum, elman1993learning, sanger1994neural, wu2020curricula}, where we train on a simpler distribution of functions in the beginning (e.g., linear functions with weight vectors restricted to a low-dimensional subspace) and gradually increase the function complexity.
This speeds up training drastically, 
often allowing us to train models that would be significantly more expensive to train without a curriculum (see Section \ref{sec:factors} for details).

\section{In-context learning of linear functions}
\label{sec:linear_functions}
In the previous section, we describe a general methodology for training Transformer models to in-context learn a class of functions.
Here, we focus on a simple function class---namely linear functions---and study how well models trained using our methodology can in-context learn this class.

\paragraph{Prompt distribution.}
We consider the class of linear functions $\mathcal{F} = \left\{f \mid  f(x) = w^\top x,\  w\in\mathbb{R}^d\right\}$, in $d$ dimensions where $d=20$.
We sample $x_1, \ldots, x_k$, $x_\text{query}$, and $w$ independently from the isotropic Gaussian distribution $N(0, I_d)$.
We then compute each $y_i=w^\top x_i$ and construct the prompt as $P = (x_1,\ y_1, \ x_2,\ y_2, \ldots,\ x_k,\ y_k,\ x_{\text{query}})$.


\paragraph{Baselines.} To contextualize the performance of our trained model, we compare it to other learning algorithms: 
(a) the least squares estimator, computing the minimum-norm linear fit to the in-context examples $(x_i,\ y_i)$,
(b) $n$-Nearest Neighbors, averaging the $y_i$ values for the $n$ nearest neighbors of $x_{\text{query}}$,
(c) averaging the values $y_ix_i$ to estimate $w$ and compute the inner product of this estimate with $x_\text{query}$.
Least squares is the optimal estimator for this problem and thus serves as a lower bound to the best error one can achieve.
The other two baselines are consistent (but sub-optimal) estimators that are easier to compute and thus provide an estimate of the performance achieved by simple approaches.
See Appendix~\ref{app:baselines} for more details.

\subsection{Transformers can in-context learn linear functions}
\label{sec:linear_functions_1}
\begin{figure}[htp]
    \centering
    \includegraphics[width=.65\textwidth]{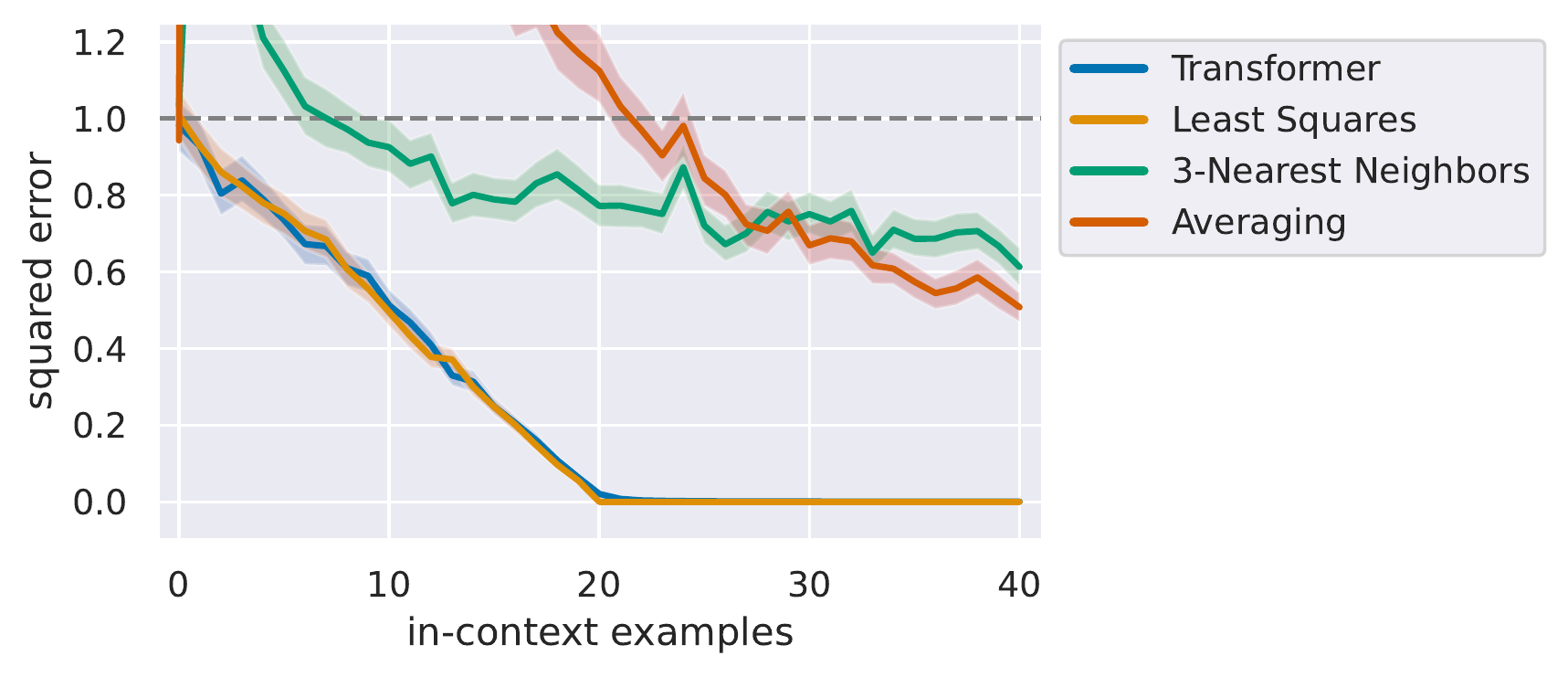}
        \caption{
    \emph{Evaluating the trained Transformer on in-context learning linear functions.}
    We plot the normalized squared error of the Transformer $((M(P) - w^\top x_{\text{query}} )^2/d)$, along with the relevant baselines, as a function of the number of in-context examples.
    Transformer's error decreases at a rate comparable to least squares.
    When the number of in-context examples reaches the problem dimension $d$ (here 20), least squares achieves $0$ error while the Transformer achieves an error of $0.02$, improving to $0.0006$ at $2d$ in-context examples.
    While the simple baselines obtain better-than-trivial error (zero estimator, dashed line), their performance is relatively poor.
    (Error averaged over 1280 prompts. 90\% confidence intervals over 1000 bootstrap trials.)}
    \label{fig:core}
\end{figure}

We show the in-context learning ability of the resulting model along with the relevant baselines in Figure \ref{fig:core}.
The trained Transformer is able to in-context learn the class of linear functions with respect to the prompt distribution specified above,
performing comparably to the optimal least squares estimator for any number of in-context examples considered.
While the simpler baselines achieve non-trivial error, they are far worse, indicating that the trained model encodes a  more complex algorithm.

\paragraph{Can memorization of training prompts explain model performance?} 
Note that the probability of the model encountering a training prompt similar to the one used for testing is astronomically low---the prompt inputs alone lie in a 800-dimensional space when predicting with $2d$ in-context examples ($d=20$).
Moreover, even considering the possibility that the model encountered a similar \emph{weight vector} during training cannot explain its performance.
That is, the model encounters 32 million random weight vectors during training and even using the best of these vectors would lead to an expected error of around $0.2$ (computed empirically, see Appendix \ref{app:memorization} for details).
 However, the model is able to achieve an error of less than $0.001$ for a prompt with $2d$ in-context examples. Further, in Section \ref{sec:factors}, we show that the model is able to obtain a similar error even when trained on prompts generated using only $10,000$  distinct weight vectors, in which case the best weight vector seen during training would yield an even worse error of around $0.5$. Thus, the model cannot be relying on memorization of training prompts or weight vectors, and instead encodes an algorithm capable of in-context learning linear functions that are very different from those seen during training.

\subsection{What functions is the model learning in-context?}
Recall that the goal of our model is: given the prompt $P=(x_1,\ w^\top x_1, \ldots, x_k,\ w^\top x_k,\ x_{\text{query}})$, output $w^\top x_\text{query}$.
Thus, if we fix the prefix given by the $k$ in-context examples, we can view the output of the model  as a function  $\hat{f}_{w,  x_{1:k}}(x_{\text{query}})$, that approximates $w^\top x_{\text{query}}$.
When $k < d$ (fewer in-context examples than dimensions), the ground truth cannot be recovered perfectly  and the ideal model should approximate $(\text{proj}_{x_{1:k}}(w))^\top x_{\text{query}}$, where $\text{proj}_{x_{1:k}}(w)$ is the projection of $w$ onto the subspace spanned by $x_1,\ldots,x_k$. Here, we will evaluate how accurately the model approximates this.

\begin{figure}
    \centering
    \begin{subfigure}[b]{0.69\textwidth}
        \centering
        \includegraphics[width=.97\textwidth]{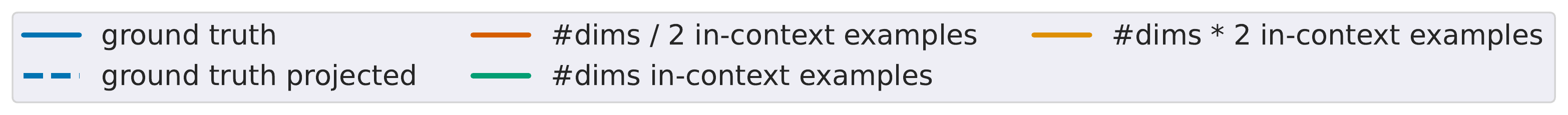}
    
        \includegraphics[height=0.3\textwidth]{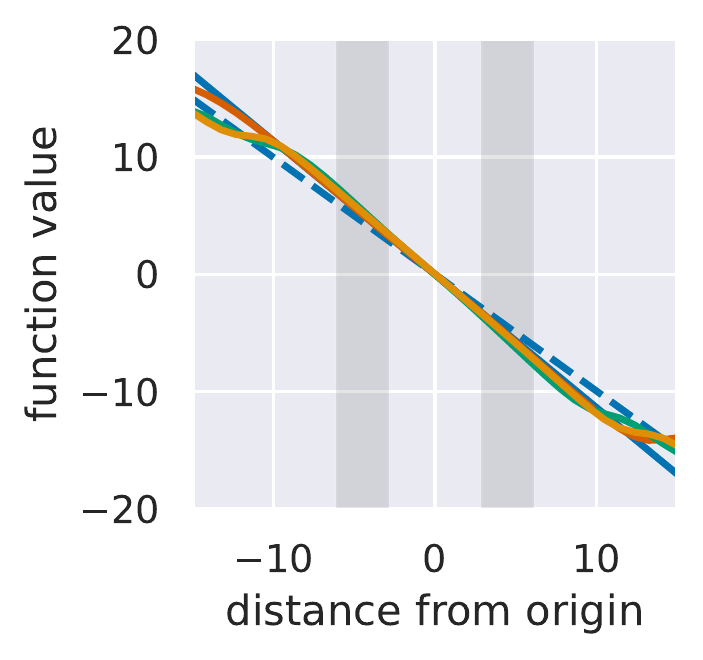}
        \includegraphics[height=0.3\textwidth]{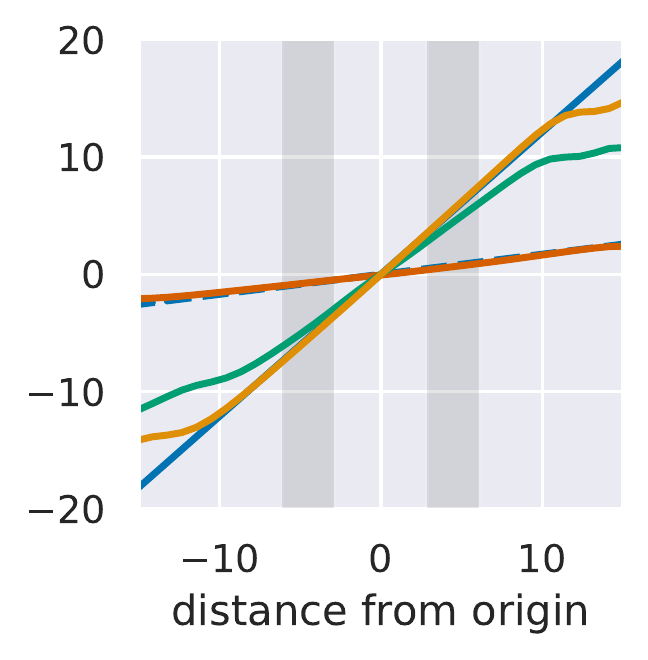}
        \includegraphics[height=0.3\textwidth]{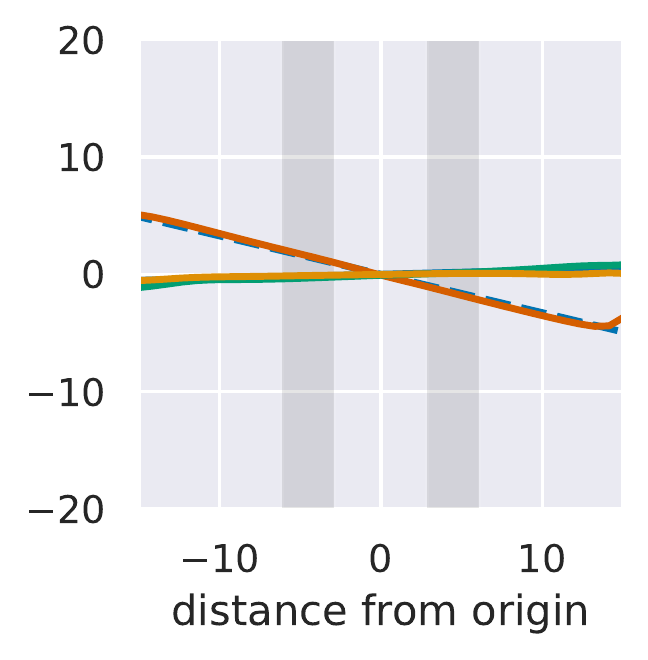}
        \caption{function visualizations}
        \label{fig:random_lines}
    \end{subfigure}
    \begin{subfigure}[b]{0.29\textwidth}
        \centering
        \includegraphics[width=\textwidth]{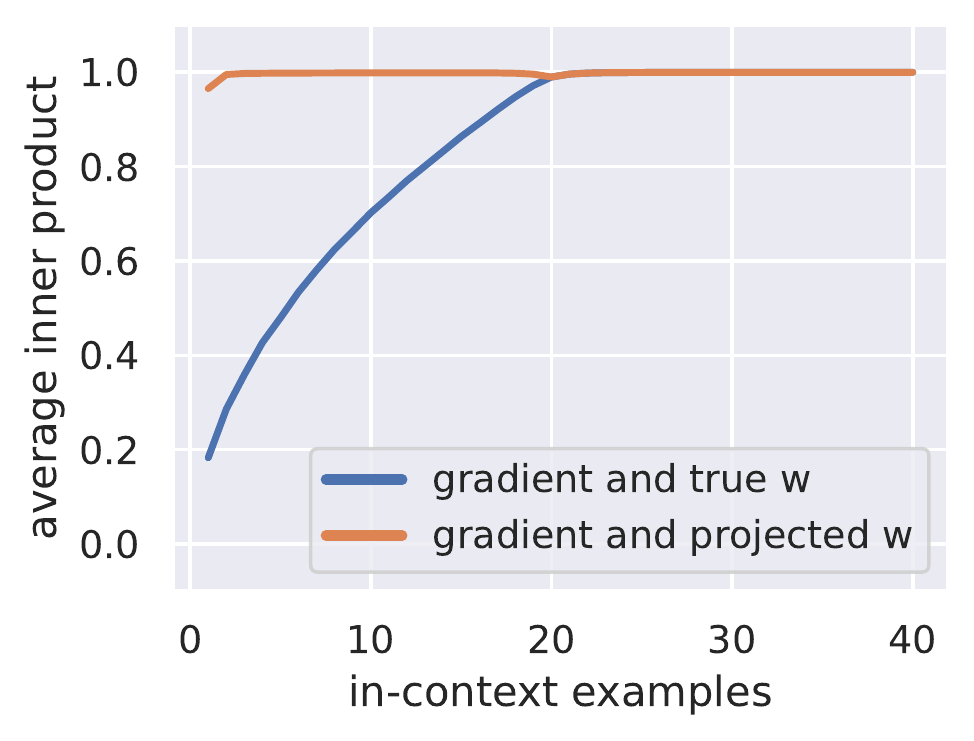}
        \caption{gradients}
        \label{fig:gradients}
    \end{subfigure}

    \caption{\emph{Understanding the prefix-conditioned function}. 
    (a) We plot the model prediction as we fix the in-context examples and vary the query input along a random direction (for three random prompts).
    The shaded regions denote the intervals in which the norm of a randomly training input lies with probability 0.99. When the scale of the query input is close to this range, the model prediction is close to the ground truth linear function (or its projection to the space of in-context inputs when $k < d$).
    (b) We compute the gradient of the model prediction with respect to the query input, and plot its (normalized) inner product with the true $w$ and projected $w$, averaged over 1280 random prompts. The gradient aligns almost perfectly with $w$ when $k \geq d$, and with projected $w$ for all $k$, indicating that the model locally aligns with the ground truth.
    }
    \label{fig:local}
\end{figure}

\paragraph{Visualizing along a random direction.}
For a randomly sampled fixed prefix, we visualize $\hat{f}_{w,  x_{1:k}}(x_{\text{query}})$  as we vary the query input along a random direction $x$ (Figure~\ref{fig:random_lines}).
That is, we pick a random unit vector $x$, and evaluate $\hat{f}_{w,  x_{1:k}}(\lambda x)$ as we vary  $\lambda$, the distance of the query input from origin. 
We observe that  $\hat{f}_{w,  x_{1:d}}(\lambda x)$ and $\hat{f}_{w,  x_{1:2d}}(\lambda x)$ closely match the ground truth  and $\hat{f}_{w,  x_{1:d/2}}(\lambda x)$ matches the projected ground truth, when the distance from origin is not too large compared to the norm of a typical randomly sampled input. In fact, in Appendix~\ref{app:query}, we show that the model is quite robust to scaling the query input:  the error doesn't increase much as we scale up the query input by a factor of up to 2, or scale down by a factor of up to 16, and degrades slowly after that.

\paragraph{Local correctness.} 
So far, we have seen that the model is able to make predictions close to the ground truth for randomly drawn query inputs and in-context examples.
We will now turn our attention to the local change of $\hat f$ around $x_\text{query}$ by considering the gradient of the function $\hat{f}_{w,  x_{1:k}}(x_{\text{query}})$ with respect to $x_{\text{query}}$ (our model is fully differentiable so we can compute the gradient directly).
Since $\hat{f}$ computed by the model should ideally approximate $\text{proj}_{x_{1:k}}(w)^\top x$, this gradient should lie in the direction of the projected ground truth $\text{proj}_{x_{1:k}}(w)$.
In Figure~\ref{fig:gradients}, we show the inner product  between the gradient and $\text{proj}_{x_{1:k}}(w)$ (both normalized),  averaged over 1280 random prompts, and observe that they align almost perfectly. Since $\text{proj}_{x_{1:k}}(w) = w$ almost surely when $k \geq d$, we observe that the gradient also aligns with $w$ perfectly in this regime. Thus the model is locally correct with respect to changes in the query input.

\section{Extrapolating beyond the training distribution}
\label{sec:distribution_shifts}

In the previous section, we demonstrated that we can train a model to in-context learn linear functions with respect to the distribution of prompts encountered during training.
That is,  we evaluate the in-context learning ability of the model with respect to distributions $D_\mathcal{X}$ and $D_\mathcal{F}$ that were also used to train the model.

Here, we evaluate the in-context learning performance of our model on prompt distributions different from the one used for training.
Our overarching goal here is to better understand the learning algorithm encoded by our model by analysing how it responds to different prompt distributions.

Formally, we will refer to the distribution of functions used during training as $D_\mathcal{F}^\text{train}$ and the corresponding distribution of prompt inputs as $D_\mathcal{X}^\text{train}$.
Then, during inference, functions are sampled from a (potentially different) distribution $D_\mathcal{F}^\text{test}$, while prompt inputs from a distribution $D_\mathcal{X}^\text{test}$.
Moreover, deviating again from our analysis so far, we also consider a separate distribution  $D_\text{query}^\text{test}$, from which the query input is sampled, potentially dependent on the rest of the in-context inputs $x_1,\ldots,x_k$ (which are still sampled from   $D_\mathcal{X}^\text{test}$).

Within this framework, we consider the same model as last section, and evaluate its performance on prompts that deviate from those encountered during training, either by
\begin{enumerate}
    \item sampling prompt inputs or functions from a different distribution, that is $D_{\mathcal{X/F}}^\text{train}\neq D_{\mathcal{X/F}}^\text{test}$ or
    \item introducing a mismatch between in-context examples and the query input, that is $D_\text{query}^\text{test} \neq D_\mathcal{X}^\text{test}$.
\end{enumerate}
We describe each such prompt structure below and present a subset of the results in Figure~\ref{fig:distribution} (see Appendix~\ref{app:dist_shift} for additional details and full results). 
Overall, the model performs reasonably accurate in-context learning with respect to these prompt distributions, indicating that it has indeed learnt to perform linear regression to some generality.

Recall that we generate a training prompt $P = (x_1, w^T x_1,   \ldots, x_k, w^T x_k, x_{\text{query}})$ by drawing the prompt inputs ($x_i$ and $x_{\text{query}}$), and the weight vector ($w$) i.i.d.\ from $N(0, I_d)$, with $d = 20$. For all the settings below, except prompt scaling, we normalize the inputs so that their expected squared norm is equal to that of inputs encountered during training. 

\begin{figure}[htp]
    \centering
    \begin{subfigure}[b]{0.33\textwidth}
        \centering
        \includegraphics[height=.75\textwidth]{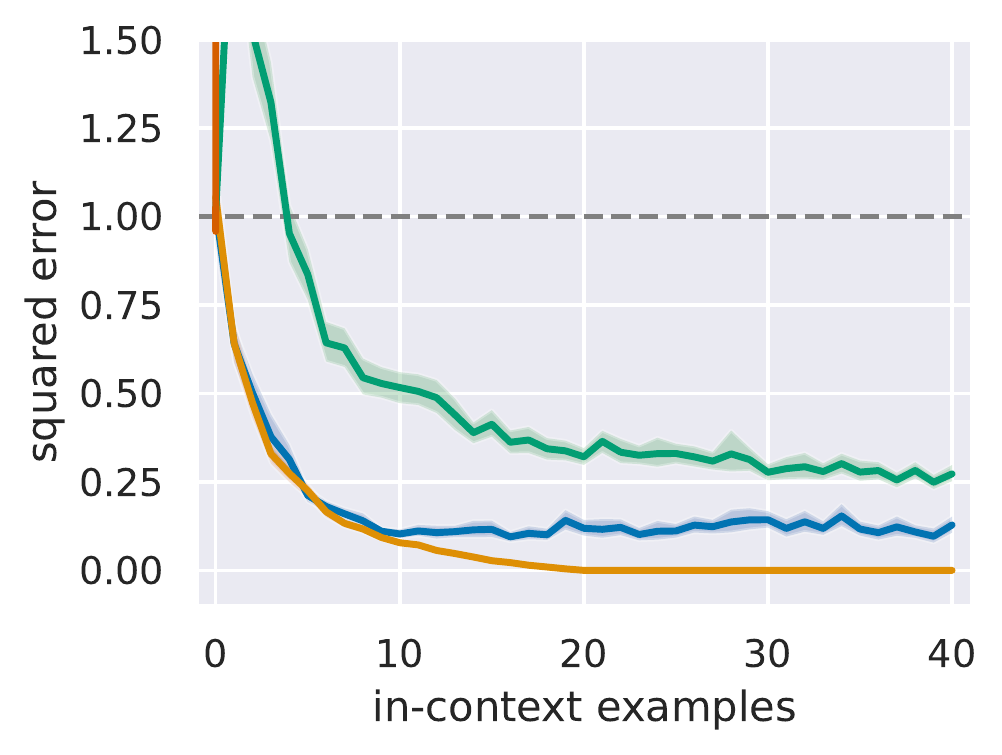}
        \caption{Skewed covariance}
        \label{fig:skewed}
    \end{subfigure}
    \begin{subfigure}[b]{0.33\textwidth}
        \centering
        \includegraphics[height=.75\textwidth]{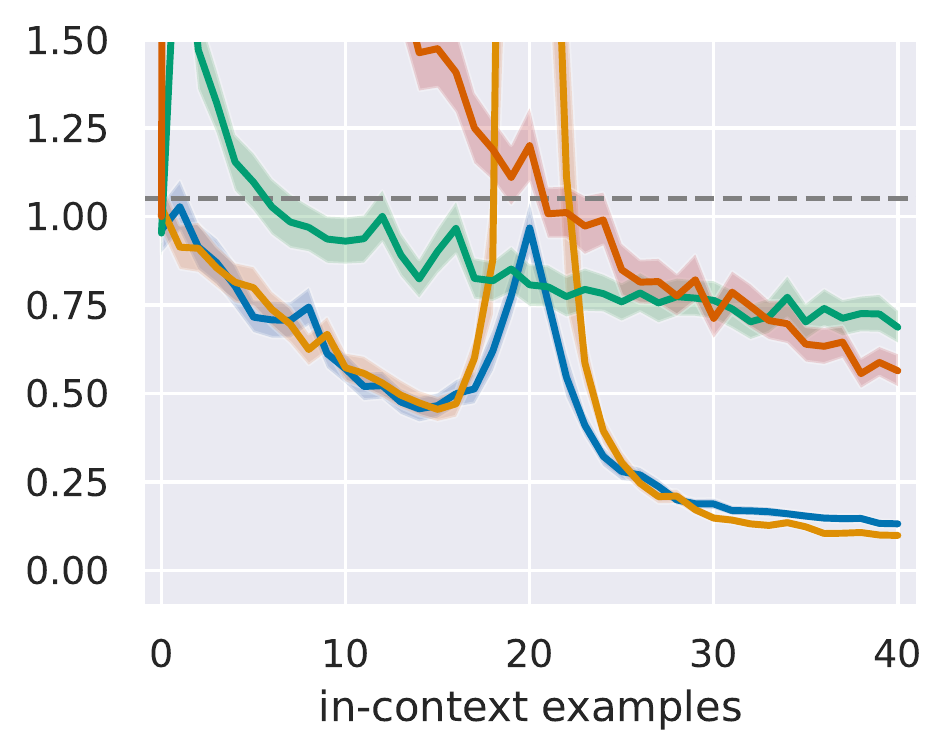}
        \caption{Noisy linear regression}
        \label{fig:noisy}
    \end{subfigure}
    \begin{subfigure}[b]{0.33\textwidth}
        \centering
        \includegraphics[height=.75\textwidth]{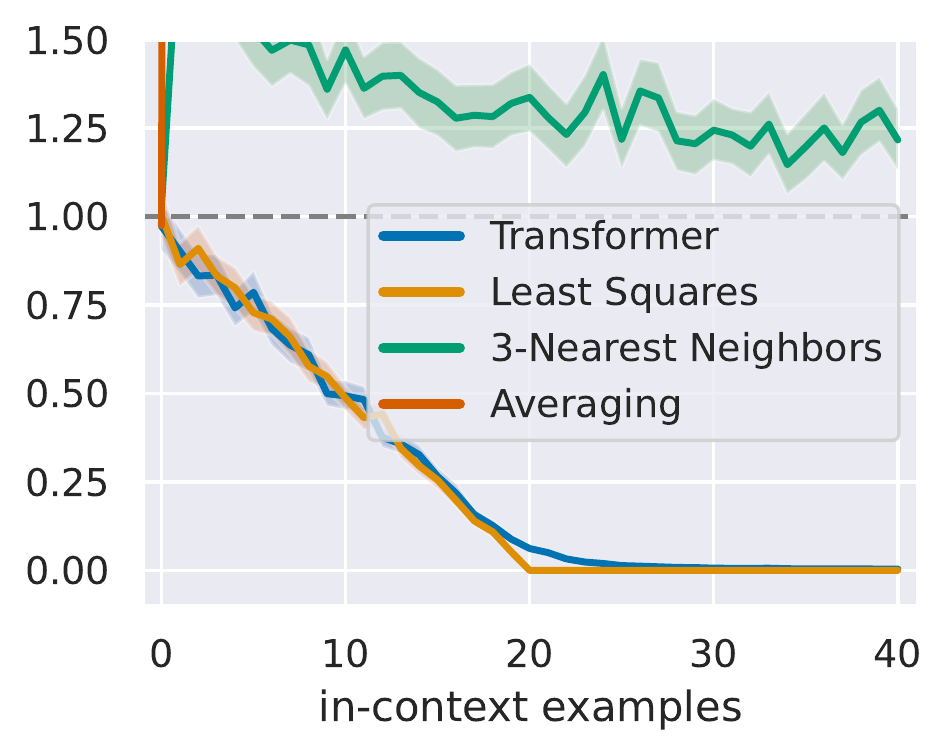}
        \caption{Different orthants}
        \label{fig:random}
    \end{subfigure}
    \caption{\emph{In-context learning on out-of-distribution prompts.}
    We evaluate the trained model on prompts that deviate from those seen during training by: 
    (a) sampling prompt inputs from a non-isotropic Gaussian,
    (b) adding label noise to in-context examples,
    (c) restricting in-context examples to a single (random) orthant.
    In all cases, the model error degrades gracefully and remains close to that of the least squares estimator, indicating that its in-context learning ability extrapolates beyond the training distribution.
    }
    \label{fig:distribution}
\end{figure}

\paragraph{Skewed covariance.} We sample prompt inputs from $N(0, \Sigma)$ where $\Sigma$ is a skewed covariance matrix with eigenbasis chosen uniformly at random and $i^{\text{th}}$ eigenvalue proportional to $\nicefrac{1}{i^2}$.
The model matches the performance of least squares until $k=10$, mimicking the sharp drop in the error in this regime, but its error plateaus afterwards (see Figure \ref{fig:skewed}). 
Thus, it is not perfectly robust to this distribution mismatch but still does relatively well, achieving less than half the error of the nearest neighbor baseline in most cases.

\paragraph{Low-dimensional subspace.} We sample prompt inputs from a random $10$ dimensional subspace. 
In this case, the model achieves low error after 10 in-context examples,  closely matching the behavior of the optimal least squares estimator (the model achieves an error of $0.036$, $0.0014$, and $0.00057$ at 10, 20, and 40 in-context examples respectively)---see Appendix Figure \ref{fig:half_app}.
Crucially, unlike the training prompts, when $k$ is between $10$ and $20$, the prompt inputs are linearly dependent, and a model  achieving low error in this regime indicates that it encodes a valid orthogonalization procedure for these inputs.

\paragraph{Noisy linear regression.} We add noise to each prompt output, that is, the $i^{\text{th}}$ output is equal to $w^T x_i + \epsilon_i$ where $\epsilon_i \sim N(0, 1)$.
The trained model closely tracks the performance of least squares when the number of in-context examples is not close to the input dimension $20$ (see Figure \ref{fig:noisy}). 
Interestingly, the model also exhibits the double descent error curve~\citep{belkin2019reconciling}  that is known to manifest for the least squares estimator~\citep{nakkiran2019more}.
Note that in this noisy setting, the optimal estimator corresponds to solving least squares with appropriate $\ell_2$-regularization. However, since the model was trained on noiseless data, we cannot expect it to learn this.

\paragraph{Prompt scale.} We consider the setting where the prompt scale between training and inference is different. We either scale the prompt inputs or the weight vectors, by a factor $\{\nicefrac{1}{3}, \nicefrac{1}{2},  2, 3\}$.
The model is relatively robust when scaling the weight vector, but not as robust when scaling the prompt inputs, especially for the more extreme scales $\nicefrac{1}{3}$ and $3$.
Specifically, for $40$ in-context examples, the model achieves errors $0.0012,0.0008,0.0016,0.0278$ when scaling the weights, and errors $0.30,0.013,0.043,0.58$ while scaling the inputs, by factors $\nicefrac{1}{3}, \nicefrac{1}{2},  2$ and $3$ respectively (Appendix Figure~\ref{fig:scale_app}). For context, recall that with 40 in-context examples, the least squares  estimator achieves an error of $0$ whereas the model achieves an error of $0.0006$ at the original scale.

\paragraph{Different orthants for in-context  and query inputs.} We fix the sign of each coordinate to be positive or negative for all in-context inputs $x_i$ (at random). 
As a result, all in-context inputs lie in the same orthant, while the query input lies in another orthant with high probability.
The model is not affected by the mismatch between in-context and query inputs and closely match the performance of least squares.
In this case, the model achieves errors $0.062$ and $0.004$ for $20$ and $40$ in-context examples respectively (see Figure \ref{fig:random}), whereas recall that it achieves errors $0.02$ and $0.0006$ on standard prompts. 
This indicates that the model is not relying on some variant of nearest neighbor search as in that case, its error would have been significantly larger (see the 3-nearest neighbor baseline).

\paragraph{Query input orthogonal to in-context inputs.} We sample the query input from the subspace orthogonal to the subspace spanned by in-context example inputs.
Here, there is no information relevant to the query input in the in-context examples and thus the model would ideally predict something close to 0 to minimize the error.
Indeed, the model outputs such a prediction, achieving an error close to $1$ (Appendix Figure~\ref{fig:orthogonal_app}).

\paragraph{Query input matches an in-context example.}
We choose the query input to match one of the in-context examples inputs chosen uniformly at random.
In this case, the model achieves errors $0.001, 0.001, 0.0005$ for $10$, $20$, $40$ examples respectively thus making close to the correct prediction, without being affected by the additional in-context examples present (Appendix Figure~\ref{fig:overlapping_app}).

\section{More complex function classes}
\label{sec:beyond_linear}
We now turn our attention to in-context learning for more complex function classes, namely sparse linear functions, decision trees, and two-layer ReLU neural networks. Here, we are back in the setting where the distribution of prompts during inference is same as that during training (except the setting of neural networks where we evaluate on linear functions as well).
The overall methodology remains the same: we sample random functions from these families and train a Transformer from scratch to approximate these functions given in-context examples. (See Appendix~\ref{app:baselines} for more details and baselines.) 

\begin{figure}[ht]
    \phantom{}
    \hfill
    \begin{subfigure}[b]{0.4\textwidth}
        \includegraphics[height=.6\textwidth]{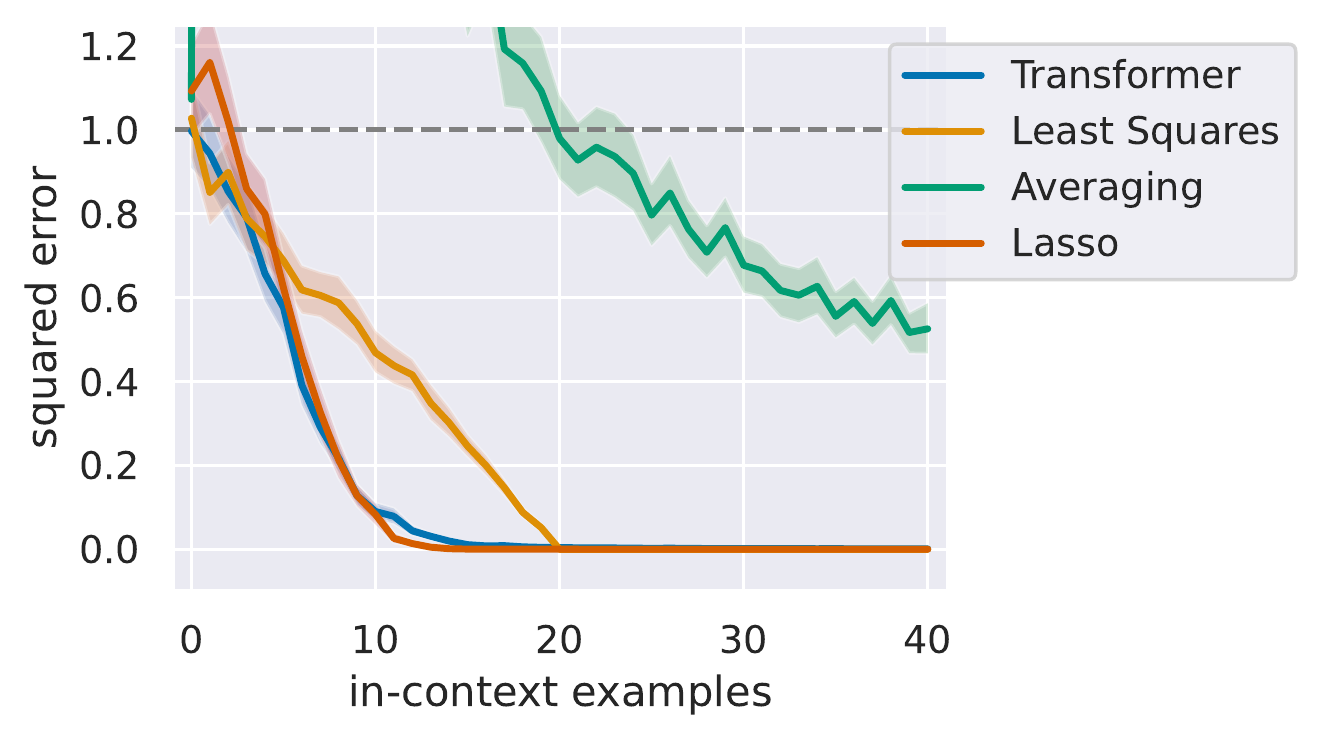}
        \caption{Sparse linear functions}
        \label{fig:sparsity}
    \end{subfigure}
    \hfill
    \begin{subfigure}[b]{0.4\textwidth}
        \centering
        \includegraphics[height=.6\textwidth]{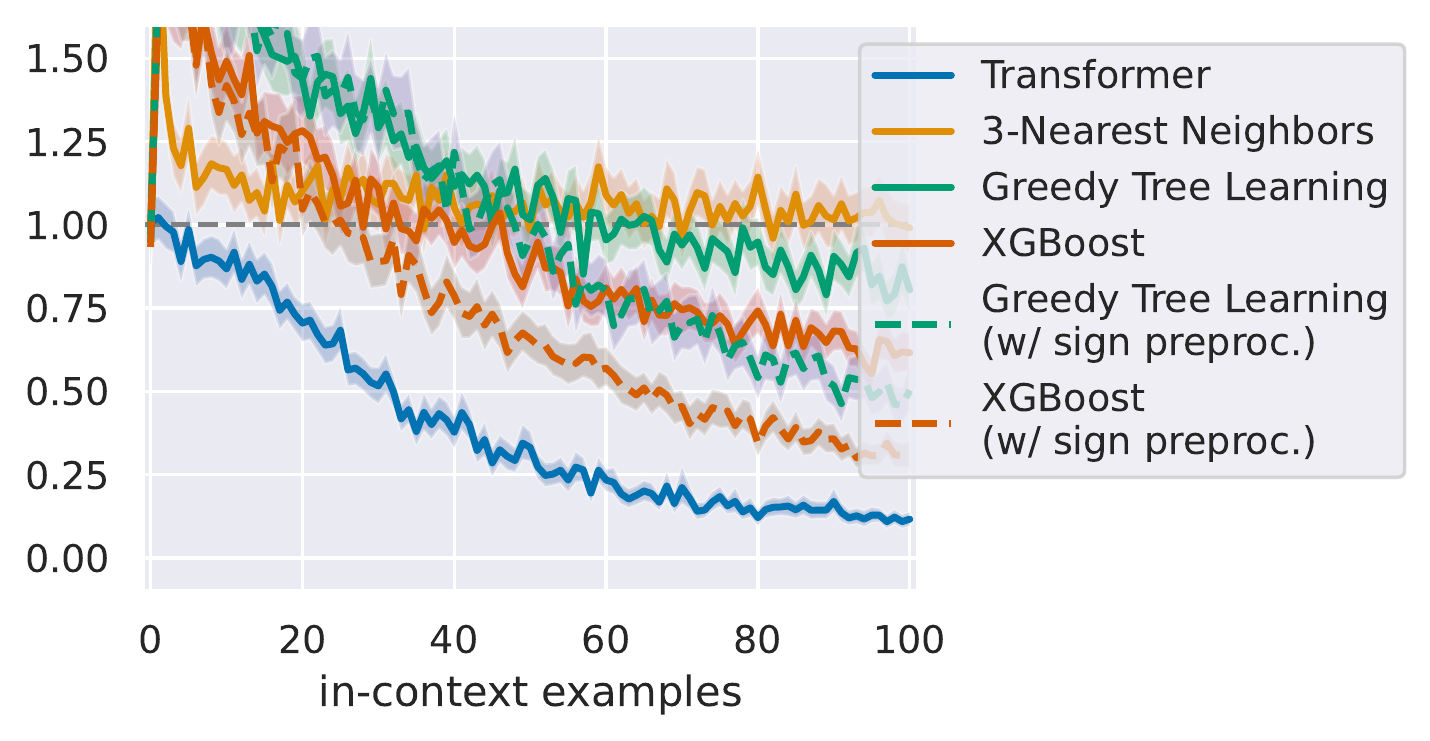}
        \caption{Decision trees}
        \label{fig:decision_trees}
    \end{subfigure}
    \hfill
    \phantom{}
    \\
    \phantom{}
    \hfill
    \begin{subfigure}[b]{0.4\textwidth}
        \includegraphics[height=.6\textwidth]{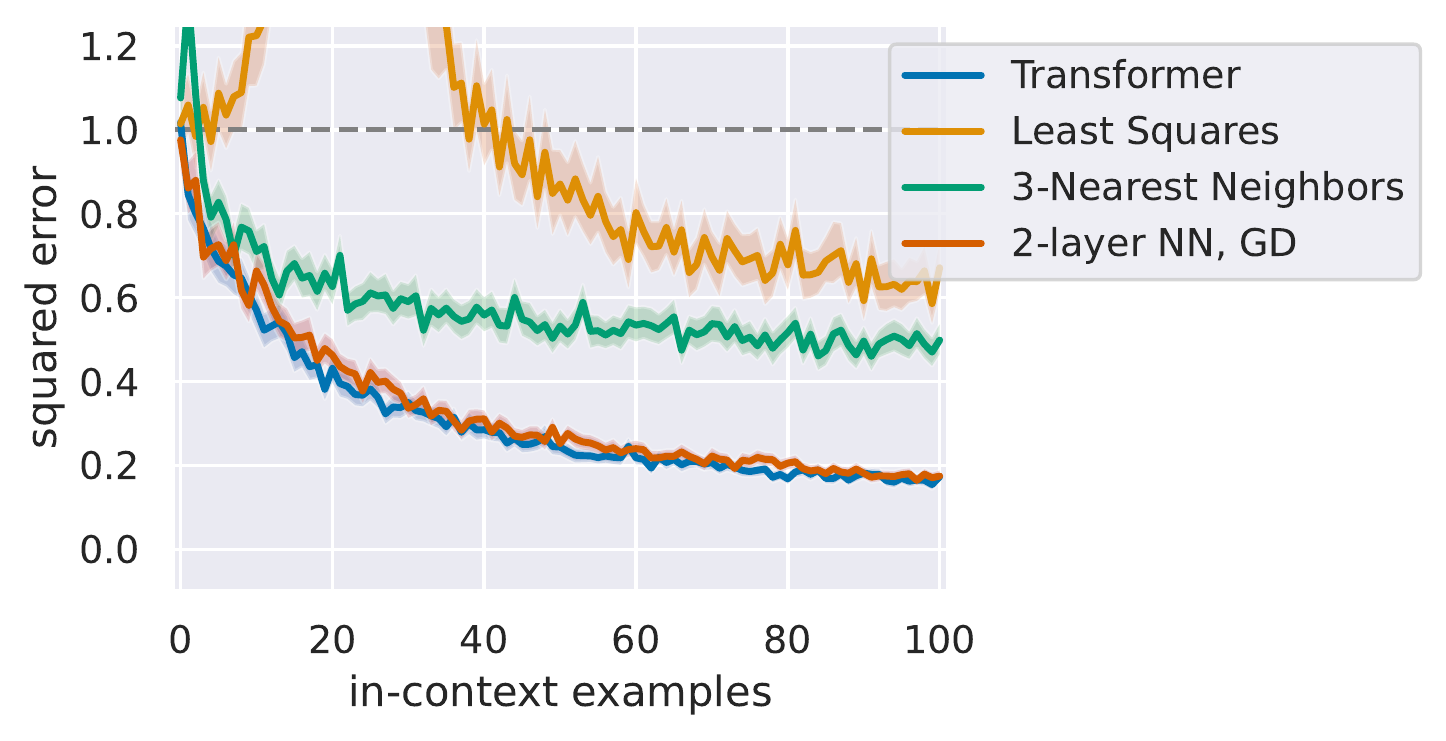}
        \caption{2-layer NN}
        \label{fig:relu}
    \end{subfigure}
    \hfill
    \begin{subfigure}[b]{0.4\textwidth}
        \centering
        \includegraphics[height=.6\textwidth]{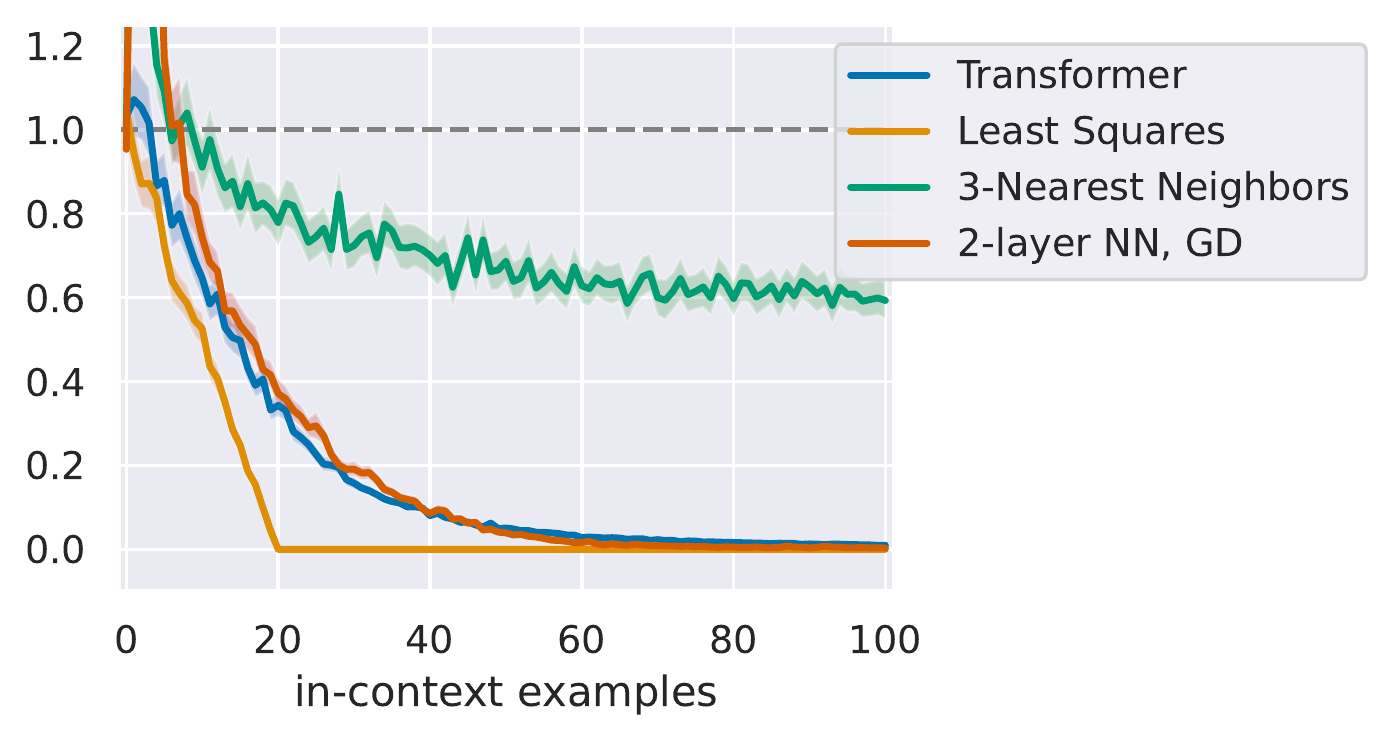}
        \caption{2-layer NN, eval on linear functions}
        \label{fig:relu_lr}
    \end{subfigure}
    \hfill
    \phantom{}
    \caption{\emph{Training a Transformer  to in-context learn more complex function classes. } 
    (a) A Transformer trained on prompts generated using sparse linear functions can in-context learn this class, with error decreasing at a rate similar to Lasso, and significantly better than minimum norm least squares.
    (b) A Transformer trained on prompts generated using random decision trees can in-context learn this class, 
    with much better performance  than greedy tree learning or tree boosting.
    (c) A Transformer trained on prompts generated using random 2-layer ReLU neural networks can in-context learn this class. The error decreases at a rate similar to the baseline which involves training a neural network using a variant of gradient descent with in-context examples as the training data. (d) The same model (from (c)) can in-context learn the class of linear functions. The error decreases at a rate slower than least squares, but comparable to a neural network trained using a variant of gradient descent. In all cases, the errors are normalized so that the trivial zero estimator achieves an error of $1$ (dashed line).
    }
\end{figure}

\paragraph{Sparse linear functions.}
First, we consider functions of the form $f(x) = w^\top x$ where $w \in \mathbb{R}^d$ and has exactly $s$ non-zero coordinates.
To sample a prompt $P = (x_1, f(x_1), \ldots, x_k,f(x_k), x_\text{query})$, we draw prompt inputs $x_i$ and $x_\text{query}$, and a weight vector $w$ from $N(0, I_d)$, and then zero out all but $s$ coordinates of $w$ uniformly at random. 
We choose $d = 20$ and $s = 3$.
In this setting, the least squares estimator is no longer optimal---one can perform better by leveraging the weight vector sparsity. 
One estimator that leverages sparsity is Lasso~\citep{tibshirani1996regression}, which involves solving the least squares objective with an $\ell_1$-norm regularizer for the weight vector.
We plot the performance of our model in Figure \ref{fig:sparsity}, and observe that it is also able to leverage sparsity, nearly matching the performance of Lasso. Our model achieves errors $0.58$ and $0.09$ while Lasso achieves errors $0.62$ and $0.08$ for $k = 5$ and $10$ respectively. Note that, unlike least squares, Lasso does not have a closed form expression and involves iterative minimization of the regularized objective, yet the Transformer is able to achieve comparable performance in a single forward pass.

\paragraph{Decision trees.} 
Next, we consider the class of depth 4 decision trees with 20 dimensional inputs. A function $f$ in this class is represented by a full binary tree (with 16 leaf nodes) where each non-leaf node is associated with a coordinate, and each leaf node is associated with a target value. To evaluate $f$ on an input $x$, we traverse the tree starting from the root node, and go to the right child if the coordinate associated with the current node is positive and go to the left child otherwise (that is, the threshold at each node is $0$). $f(x)$ is given by the value associated with the leaf node reached at the end. To sample a  random prompt $P = (x_1, f(x_1), \ldots, x_k, f(x_k), x_\text{query})$, we draw prompt inputs $x_i$s and $x_\text{query}$ from $N(0, I_d)$, and $f$ corresponds to a tree where the coordinates associated with the non-leaf nodes are drawn uniformly at random from $\{1, 2,  \ldots, d\}$ and the values associated with the leaf nodes are drawn from $N(0, 1)$. 
In Figure \ref{fig:decision_trees}, we show that Transformers can be trained to in-context learn this class, with performance much better than  greedy tree learning and boosting (via XGBoost~\citep{chen2016xgboost}).
With $k=100$ in-context examples, the Transformer achieves an error of $0.12$ whereas greedy learning achieves an error of $0.80$ and XGBoost achieves an error of $0.62$.

Since the decision trees in our function class predict solely based on the sign of each coordinate of $x_i$, we also consider a baseline where we provide the greedy learning and XGBoost algorithms with the signs of each $x_i$ instead. 
This significantly improves their performance---at 100 in-context examples, greedy achieves an error of 0.50 and XGBoost an error if 0.31---but they still perform much worse than the trained Transformer.

Note that, in general, we do not have a good understanding of the space of efficient algorithms for learning decision trees, and the conditions under which known heuristics work \citep{blanc2021decision, brutzkus2020id3}.
At the same time, we found that Transformers can be trained to directly discover such an algorithm for the prompt distribution we considered. This suggests an intriguing possibility where we might be able to reverse engineer the algorithm encoded by a Transformer to obtain new sample efficient algorithms for existing learning problems. 

\paragraph{Two-layer ReLU neural networks.}
Finally, we consider the class of two layer ReLU neural 
networks containing functions of the form $f(x) = \sum_{i=1}^r \alpha_i \ \sigma(w_i^\top x)$, where $\alpha_i \in \mathbb{R}$, $w_i \in \mathbb{R}^d$ and $\sigma(\cdot) = \max(0, \cdot)$ is the ReLU activation function.
To draw a random prompt $P = (x_1, f(x_1), \ldots, x_k, f(x_k), x_\text{query})$, we sample prompt inputs $x_i$s and $x_\text{query}$ from $N(0, I_d)$, along with network parameters $a_i$s and $w_i$s from $N(0, {2/r})$ and $N(0, I_d)$ respectively. We set the input dimension $d$ to $20$ and the number of the hidden nodes $r$ to $100$. 
In Figure~\ref{fig:relu}, we show that Transformers can be trained to in-context learn this class of functions.
In fact, the Transformer performs comparably to the baseline 
 which involves training a two-layer neural network of the same architecture on in-context examples using Adam \citep{kingma2014adam}, a variant of gradient descent
(see Appendix \ref{app:baselines} for details). Specifically, for $k=100$ in-context examples, both the Transformer and the neural network trained on in-context examples achieve an error of $0.17$.

Moreover, the model trained to in-context learn two-layer neural networks is also able to in-context learn linear functions (for which it is not explicitly trained), albeit with a rate slower than least squares, but comparable to a neural network trained on in-context examples generated using a linear function (Figure \ref{fig:relu_lr}). For $k=20$, $50$, and $100$ in-context examples respectively, the Transformer achieves error $0.34$, $0.05$, and $0.01$, and the two-layer network achieves error $0.37$, $0.04$, and $0.003$ (the least squares estimator achieves error $0$ for $k\geq 20$).

\section{Investigating what matters for in-context learning}
\label{sec:factors}
We now return to the setting of training models to in-context learn linear functions and explore different factors that lead to successful in-context learning.

\begin{figure}[htp]
    \hfill
    \begin{subfigure}[b]{0.28\textwidth}
        \centering
        \includegraphics[height=.9\textwidth]{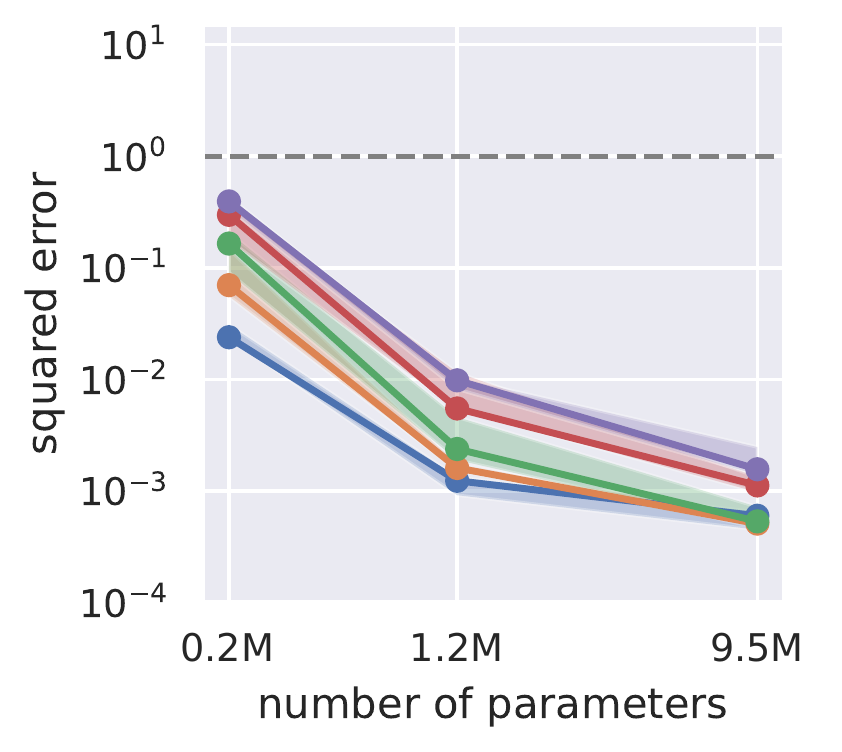}
        \caption{Standard}
        \label{fig:cap_standard}
    \end{subfigure}
    \hfill
    \begin{subfigure}[b]{0.28\textwidth}
        \centering
        \includegraphics[height=.9\textwidth]{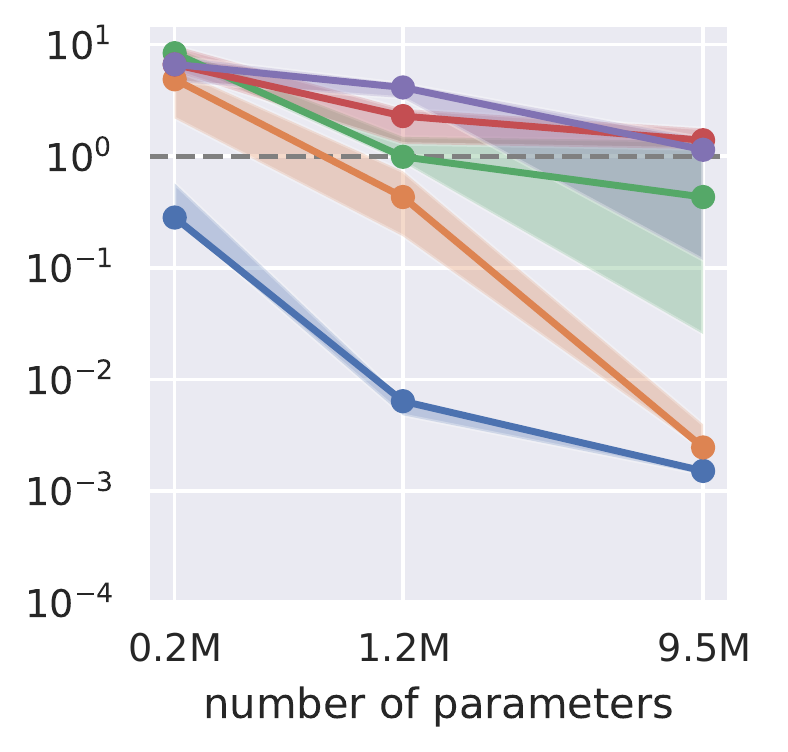}
        \caption{Different orthants}
        \label{fig:cap_random}
    \end{subfigure}
    \hfill
    \begin{subfigure}[b]{0.28\textwidth}
        \centering
        \includegraphics[height=.9\textwidth]{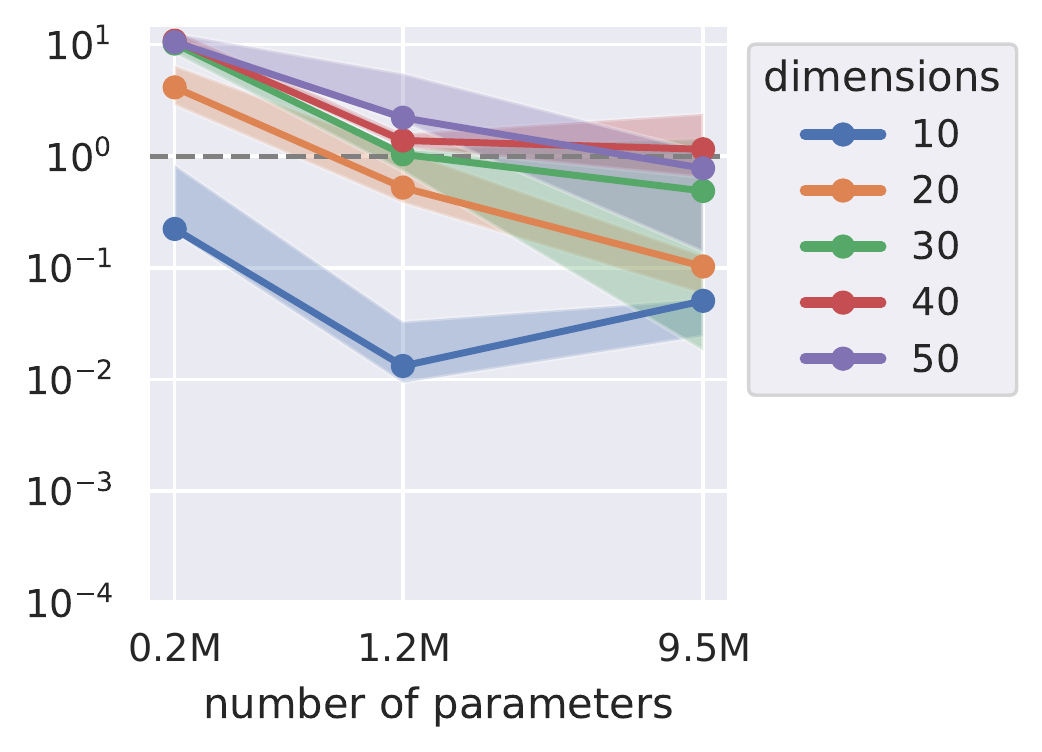}
        \caption{Skewed covariance}
        \label{fig:cap_skewed}
    \end{subfigure}
    \hfill
    \phantom{dsada}
    \caption{\emph{Understanding the effect of model capacity and problem dimension on in-context learning performance for in-distribution (a) and out-of-distribution (b,c) prompts.} We train Transformers to in-context learn linear functions and plot the error with $2d$ in-context examples as we vary problem dimension $d$ and model capacity.  Capacity helps with in-context learning in most cases, especially on out-of-distribution prompts (even when the absolute gains in the in-distribution setting are small). We train 3 models in each case with different random seeds, and show the median error (solid lines), and the minimum and maximum errors (shaded region). (See Appendix~\ref{app:variance} for training variance analysis.)}
    \label{fig:capacity}
\end{figure}

\paragraph{Problem Dimension and Capacity.}
In Section \ref{sec:linear_functions} and \ref{sec:distribution_shifts}, we saw that Transformer models can be trained to in-context learn 20-dimensional linear functions accurately and relatively robustly. 
To explore the interplay between problem dimensionality and capacity, we also consider models with fewer parameters (see Appendix~\ref{app:architecture}) and train each architecture on \{10, 30, 40, 50\}-dimensional problems.
In Figure~\ref{fig:capacity}, we plot the model error with $2d$ in-context examples as we vary the problem dimension $d$ and the model capacity. 
In the standard setting, i.e., when the training and inference time prompt distributions are the same, we observe that the  error decreases as we increase the capacity or reduce the problem dimensionality (see Figure~\ref{fig:cap_standard}). 
Thus, model capacity helps perform accurate in-context learning.
For out-of-distribution prompts, we observe that the settings where the input covariance is skewed or where in-context example inputs and query inputs lie in different orthants are particularly challenging, especially for higher dimensional problems.
However, the error decreases considerably (in most cases) as we increase the model capacity, even when absolute decrease in the standard error is small (see Figure \ref{fig:cap_random} and \ref{fig:cap_skewed}). 
See Appendix \ref{app:capacity}  for additional plots.

\paragraph{Curriculum.}
We train our models using curriculum learning. That is, we initially draw the prompt inputs from a fixed $5$ dimensional subspace (by setting some of the coordinates to $0$) with prompt length $11$ (number of input-output pairs), and increase the subspace dimension by $1$ and prompt length by $2$ every $2,000$ training steps, until the subspace dimension reaches the ambient dimension $d$ and prompt length reaches $2d+1$ (see Appendix \ref{app:training} for details).
This process can also be viewed as gradually increasing the complexity of the function class.
This speeds up training drastically, especially for higher dimensional problems: for dimension 50, the loss barely decreases through the 500k training steps without curriculum but reaches close to the optimum with curriculum. 
For the 20 dimensional problem where we were able to train the model without curriculum within the training (step count) budget, we did not observe any qualitative difference in accuracy or robustness compared to the model trained with curriculum. 
We include plots comparing the speed and accuracy of training with and without curriculum in Appendix \ref{app:curriculum}.

Notably, when training Transformers without curriculum, there is an initial---relatively long---period  in training where the loss does not decrease, followed by a period of sharp decrease.
The length of this period varies with training randomness and seems to increase on average with problem dimension.
Understanding the model just before and after this transition moment is a promising future direction, which can give insights into the emergence of  in-context learning. 
Interestingly,~\citet{olsson2022context} observe a similar jump in the in-context learning ability of a language model which they attribute to the formation of ``induction heads''.

\paragraph{Number of distinct prompts or functions seen during training.} 
To estimate the amount of training data required for in-context learning, we perform two ablation studies. In the first study, we limit the number of distinct prompts seen during training. That is, we create a set of $n_p$ randomly generated prompts (as described in Section~\ref{sec:setting}), and sample prompts from this set during training (here, we train without curriculum, as it would introduce additional prompts during the warmup phase). In the second study, we only limit the number of distinct functions used for training.
That is we create a set of $n_w$ randomly chosen vectors (corresponding to $n_w$ linear functions) and sample weight vectors uniformly from that set to generate the training prompts (the inputs are still sampled from $N(0, I_d)$ for each training prompt).
We find that the amount of training data required is relatively small: non-trivial in-context learning is possible with $n_p = 100\text{k}$  or $n_w = 1\text{k}$, and the error drops close to that of the unrestricted model (discussed in Section \ref{sec:linear_functions}) with $n_p = 1\text{M}$ or $n_w = 10\text{k}$ (details in Appendix \ref{app:distinct_funcs}).
For context, in Section \ref{sec:linear_functions}, the model is trained on fresh prompts each step, thus encountering 32M distinct functions and prompts (500k training steps with 64 prompts/batch). 

\section{Related work}
\label{sec:related}
\paragraph{In-context learning.} Since \citet{brown2020language} demonstrated the in-context learning ability of GPT-3, there has been a significant interest in improving and understanding this capability \citep{liu2021makes, min2021noisy, zhao2021calibrate, lu2021fantastically, rubin2021learning, min2021metaicl, chen2021meta, mishra2021reframing, lampinen2022can}.
The works most relevant to ours are as follows.
\citet{xie2022an} propose a Bayesian inference framework explaining how in-context learning works despite formatting differences between training and inference distributions. \citet{razeghi2022impact} show that in-context learning for numerical reasoning tasks is better for instances whose terms are more prevalent in training data. \citet{min2021noisy} demonstrate tasks where in-context learning  works  even when the prompt outputs are chosen randomly, questioning to what extent these models are truly  learning new tasks on-the-fly, while \citet{rong_2021} 
gives examples of novel tasks on which these models demonstrate on-the-fly learning ability. \citet{chan2022data} demonstrate that distributional properties such as long-tailedness are crucial for in-context learning on an image-based few-shot dataset. \citet{olsson2022context} and  \citet{elhage2021mathematical} consider a different framing of in-context learning, referring to any model behavior that utilizes information in a prompt to make predictions that improve with prompt size. They hypothesize the existence of special circuits inside Transformer models responsible for in-context learning, that can complete prompts by copying previous similar patterns in the prompt sequence. 
\citet{pesut2022who} and \citet[Table~16]{dinh2022lift} consider in-context learning for small tabular datasets and learning problems in one and two dimensions, and show that GPT-3 can obtain non-trivial accuracy.
Our work contributes to and complements this line of work, by posing in-context learning as a well-defined problem of learning function classes at inference time, and empirically investigating training models that in-context learn simple function classes.

\paragraph{Transformers.} There is a long line of work investigating the capabilities~\citep{vaswani2017attention,dehghani2018universal,yun2019transformers,perez2019turing,yao2021self,bhattamishra2020computational,zhang2022unveiling}, limitations~\citep{hahn2020theoretical, bhattamishra2020ability}, applications~\citep{lu2021pretrained,dosovitskiy2020image, parmar2018image}, and internal workings~\citep{elhage2021mathematical,snell2021approximating,weiss2021thinking,edelman2022inductive,olsson2022context} of Transformer models.
Most similar to our work, \citet{muller2021transformers} and \citet{nguyen2022transformer} demonstrate the ability of Transformer models to solve prediction tasks using the input context, albeit in different settings. \citet{muller2021transformers} introduce a ``Prior-data fitted transformer network'' that is trained to approximate Bayesian inference with priors such as Gaussian processes and Bayesian neural networks,  and use it to perform downstream tasks such as tabular dataset classification and few-shot image classification. \citet{nguyen2022transformer} introduce Transformer neural processes, building on prior work on neural processes~\citep{garnelo2018neural, garnelo2018conditional, kim2019attentive}, and show that they achieve state-of-the art performance on tasks such as image completion and contextual multi-armed bandits.
Our work complements these works, focusing on understanding the in-context learning ability of Transformers for various simple function classes and the extent to which this ability extrapolates beyond the training distribution.

\paragraph{Meta learning.} Training a model to perform in-context learning
can be viewed as an instance of the more general learning-to-learn or meta-learning paradigm \citep{schmidhuber1987evolutionary,naik1992meta,thrun2012learning}.
Typical approaches from this extensive line of work (see \citep{hospedales2020meta} for a survey) include:
training a meta-learner on how to update the parameters of a downstream learner \citep{bengio1995optimization, li2016learning}, learning parameter initializations from which one can quickly train for many downstream tasks \citep{finn2017model, ravi2016optimization},
learning latent embeddings that allow for effective similarity search~\citep{snell2017prototypical}.
Most relevant to our setting are approaches that directly  take as input examples from a downstream task and a query input and produce the corresponding output~\citep{hochreiter2001learning,mishra2018simple, santoro2016meta, garnelo2018neural, garnelo2018conditional,kirsch2021meta}.
Our work contributes to this line of work, by investigating the learning-to-learn abilities of Transformer models in a well-defined setting.

\paragraph{Data-driven algorithm design.}
Another line of work aims to discover algorithms that perform well on a distribution of inputs~\citep{horvitz2001bayesian,xu2008satzilla,vinyals2015pointer,bello2016neural,khalil2017learning,selsam2018learning,schwarzschild2021can} (as opposed to algorithms with guarantees on their worst-case performance).
See~\citet{balcan2020data} for a survey on advancements on the theoretical foundations of such algorithms.
Our work can be viewed as part of this line of work, as we train Transformer models to discover algorithms for different learning problems.

\section{Discussion}
\label{sec:discussion}
In this work, we formalize and study the question: can we train models that learn different classes of functions in-context? We show that Transformer models trained from scratch can in-context learn the class of linear functions, with performance comparable to the optimal least squares estimator, even under distribution shifts. Moreover, we  show that in-context learning is also possible for sparse linear functions, decision trees, and two-layer neural networks; learning problems which are solved in practice with involved iterative algorithms such as gradient descent. 

At the same time, understanding the implications of our results for language models requires further investigation. A pertinent question regarding the in-context learning capabilities of language models is how they leverage in-context examples \citep{min2022rethinking}. Our results demonstrate that Transformers can encode complex learning algorithms that utilize in-context examples in a far-from-trivial manner. In fact, this is the case for standard Transformer architectures trained with standard optimization procedures. The extent to which such non-trivial in-context learning behavior exists in large language models is still open, but we believe that our work takes a step towards formalizing and investigating this question.

Our work lays the groundwork for several future directions.

\paragraph{Complexity of in-context learning.} We empirically show that model capacity helps in performing in-context learning accurately and robustly. This raises the question: How does the in-context learning loss \eqref{eq:in-context_loss} depend on the complexity of the function class $\mathcal{F}$, the capacity of model $M$, and the number of prompts used to train $M$. Even the right notion of complexity of $\mathcal{F}$ is unclear and may depend on the model family. Understanding this question for models explicitly trained to perform in-context learning may suggest an upper bound for the in-context learning performance of models such as GPT-3 that have not been explicitly trained for this purpose.

\paragraph{Curriculum learning.} Within our framework, there is natural notion of curriculum learning where during training, we gradually increase the complexity of the function class learned in-context. This leads to drastic speed-ups in training. What is the reason behind such a speedup? Are similar speedups also possible for training large language models? Understanding these questions can have implications for training of models on large real-world datasets, potentially reducing the time and energy used for training.

\paragraph{Inductive bias of model families.} Our framework presents an opportunity to understand and compare the inductive biases of different model families (e.g., Transformers vs.\ LSTMs) in a well-defined setting. For instance, a concrete question is: Are there function classes that are easier to in-context learn using Transformers but harder for LSTMs and vice-versa?

\paragraph{Understanding the learning algorithms encoded in Transformers.}
The models we train are able to perform in-context learning, and are thus themselves encoding learning algorithms. A worthwhile research
direction would be to investigate the internal workings of these models and better understand the exact
learning algorithms that they encode.
Moreover, for settings such as decision trees, we do not have a good understanding of what the optimal learning algorithms are.
Nevertheless, in Section~\ref{sec:beyond_linear} we found that Transformers are able to discover sample efficient algorithms when being trained to perform in-context learning.
This suggests an intriguing possibility where we might be able to reverse engineer the Transformer to obtain better learning algorithms for such problems.

\section*{Acknowledgements}
We thank Niladri Chatterji, Micah Goldblum, Rohith Kuditipudi, Shibani Santurkar, Carmen Strassle,  Mirac Sugzun, and Li-Yang Tan for helpful conversations, and anonymous reviewers for helpful comments. 

SG was funded by a  Stanford Interdisciplinary Graduate Fellowship.
DT was funded by Open Philanthropy, and partially supported by NSF Award CCF-1813049.  GV was supported by NSF Awards CCF-1704417, CCF-1813049, Frontier Award 1804222 and DOE award DE-SC0019205.
We performed our experiments on the Stanford NLP cluster.

\bibliography{refs}


\clearpage
\appendix

\section{Experimental setup}
\label{app:setup}
Here, we provide additional details on our experimental setup.

\subsection{Model architecture}
\label{app:architecture}

We use architectures from the GPT-2 family~\citep{radford2018improving} as implemented by HuggingFace~\citep{wolf-etal-2020-transformers}
~\footnote{\url{https://huggingface.co/docs/transformers/model_doc/gpt2}} .
Specifically, we consider the following set of configurations.
\begin{center}
    \begin{tabular}{lrrrr}
    \toprule
        Model & Embedding size & \#Layers & \#Heads & (Total parameters) \\
    \midrule
        Tiny & 64 & 3 & 2 & 0.2M \\
        Small & 128 & 6 & 4 & 1.2M \\
        \textbf{Standard} & 256 & 12 & 8 & 9.5M \\
    \bottomrule
    \end{tabular}
\end{center}
We use the Standard model for the bulk of our experiments and only consider the smaller models for the capacity explorations in Section~\ref{sec:factors} and Appendix~\ref{app:capacity}.
Since we train on each input once (we sample new inputs at each training step), overfitting to the training data is not an issue. Therefore, we set the Dropout probability to 0.

Out of the box, these models take as input a sequence of vectors in embedding space and output a sequence of vectors in the same space.
However, the tasks we study are functions from a lower dimensional vector space (e.g., 10-50 dimensions) to a scalar value.
Thus, in order to use a prompt such as $x_1, f(x_1), x_2, f(x_2)\ldots$, we need to map $x_i$s and $f(x_i)$s to vectors in embedding space.
We do so by first turning the scalars $f(x_i)$ into vectors of the same dimension as $x_i$ by appending 0s and then applying a learnable linear transformation to map all these vectors into the embedding space.
Finally, we map the model output into a scalar value through a dot product with a learnable vector.

We treat the prediction of the model at the position corresponding to $x_i$ (that is absolute position $2i-1$) as the prediction of $f(x_i)$.
Due to the structure of these models, this prediction only depends on $(x_j, f(x_j))$ for $j<i$ and $x_i$.
We ignore the model predictions at positions corresponding to $f(x_i)$.

\subsection{Training}
\label{app:training}
Each training prompt is produced by sampling a random function $f$ from the function class we are training on, then sampling inputs $x_i$ from the isotropic Gaussian distribution $N(0, I_d)$ and constructing a prompt as $(x_1, f(x_1),\ldots,x_k, f(x_{k}))$.
Given a prompt, we obtain model predictions $\hat y_i$ (meant to approximate $f(x_i)$) for each input, and compute the loss
\[
    \frac{1}{k} \sum_{i=1}^k \left(\hat{y_i}-f(x_i)\right)^2.
\]
At each training step, we average the loss over a batch of randomly generated prompts (with different functions and prompt inputs), and perform an update step.
We use the Adam optimizer \citep{kingma2014adam}, and train for 500,000 total steps with a batch size of 64. We use a learning rate of $10^{-4}$ for all function classes and models.

\paragraph{Curriculum learning.} 
To accelerate training, we start by training on prompt inputs $x_i$ lying in a smaller dimensional subspace, and with fewer inputs per prompt, and gradually increase the subspace dimension and number of prompt inputs.
Specifically, we zero out all but the first $d_\text{cur}$ coordinates of $x_i$, sample prompts of size $k_\text{cur}$ and leave the rest of the training process the same.
We use the same schedule for all training runs for the function classes of linear functions and sparse linear functions, starting with $d_\text{cur} = 5,~ k_\text{cur} = 11$, and increasing $d_\text{cur}$ and $k_\text{cur}$ by $1$ and $2$ respectively, every 2000 steps, until $d_\text{cur} = d$, $k_\text{cur} = 2d+1$. We use a slightly different schedule for 2 layer neural networks and decision trees as we want prompts with more inputs for these function classes.
For these classes, we start with $d_\text{cur} = 5,~ k_\text{cur} = 26$, and increase $d_\text{cur}$ and $k_\text{cur}$ by $1$ and $5$ respectively, every 2000 steps, until $d_\text{cur} = d$, $k_\text{cur} = 5d+1$.

Overall, with curriculum, a training prompt $(x_1, f(x_1), \ldots, x_{k_\text{cur}}, f(x_{k_\text{cur}})$ is generated by sampling a random function $f$ from the function class,  drawing inputs $x_i$ by sampling i.i.d.  from $N(0, I_d)$ and zeroing out all but the first $d_\text{cur}$ coordinates. Given model predictions $\hat{y_i}$, the loss is given by 
\[
    \frac{1}{k_\text{cur}} \sum_{i=1}^{k_\text{cur}} \left(\hat{y_i}-f(x_i)\right)^2.
\]

\paragraph{Sampling random functions.} For the class of linear functions, we sample random function $f(x) = w^\top x$ by drawing $w \sim N(0, I_d)$. For our main setting (Section \ref{sec:linear_functions} and \ref{sec:distribution_shifts}), we set $d = 20$.

For the class of two-layer neural networks, we sample $f(x) = \sum_{i = 1}^r \alpha_i \sigma(w_i^\top x)$, where $\alpha_i$s and $w_i$s are drawn i.i.d. from $N(0, 2/r)$ and $N(0, I_d)$ respectively.  We set $d = 20$ and $r = 100$.

For the class of $k$-sparse linear functions, we sample $f(x) = w^\top x$ by drawing $w \sim N(0, I_d)$ and zeroing out all but $k$ coordinates of $w$ chosen uniformly at random from the first $d_\text{cur}$ coordinates (as defined in the curriculum learning description above).
We set $d = 20$ and $k = 3$. 

For the class of decision trees, the random function $f$ is represented by a decision tree of depth $4$ (with 16 leaf nodes), with 20 dimensional inputs. Each non-leaf node of the tree is associated with a coordinate selected uniformly at random from $\{1, 2, \ldots, d\}$, and each leaf node is associated with a value drawn randomly from $N(0, 1)$. To evaluate $f$ on an input $x$, we traverse the tree starting from the root node, and go to the right child if the coordinate associated with the current node is positive and go to the left child otherwise. $f(x)$ is given by the value associated with the leaf node reached at the end.

\paragraph{Computational resources.}
We train using a single NVIDIA GeForce RTX 3090 GPU and most training runs  take 5-20 hours depending on model size and context length. For instance, for the class of linear functions, training the standard model takes 17 hours for $d = 50$, 7 hours for $d = 20$ and 5.5 hours for $d = 10$. For decision trees, training the standard model takes 17 hours. The time it takes for decision trees and $50$ dimensional linear functions is higher due to larger context lengths (we train for $d$ dimensional linear functions with $2d+1$ input-output pairs per prompt).

\subsection{Baselines}
\label{app:baselines}
\paragraph{Least squares.} Minimum norm least squares is the optimal estimator for the linear regression problem. Given a prompt $P = (x_1, y_1, \ldots, x_k, y_k, x_{\text{query}})$, let $X$ be a $k \times d$ matrix with row $i$ given by $x_i$, and let $y$ be a $k$ dimensional vector with the $i^{\text{th}}$ entry $y_i$. Set $\hat{w}^T  = X^{+} y$, where $X^{+}$ denotes the Moore-Penrose pseudoinverse of $X$. The estimator predicts $M(P) = \hat{w}^T x_{\text{query}}$.

\paragraph{Averaging estimator.} This corresponds to $M(P) = \hat{w}^T x_{\text{query}}$ where $\hat{w} = \frac{1}{k} \sum_{i=1}^k x_i y_i$. This estimator is consistent (yet sub-optimal) when $x_i$s are drawn from $N(0, I_d)$. Unlike least squares, this estimator does not involve an inverse computation, and might be easier for a model to encode.

\paragraph{Nearest neighbors.} This corresponds to setting $M(P) = \frac{1}{n} \sum_{i \in S} y_i$. Here, $S$ is the set of indices of the $n$ nearest neighbors of $x_{\text{query}}$ among $x_1$ to $x_k$. For $k < n$, we average over all the $y_i$s from $1$ to $k$, and for $k=0$, we set $M(P) = 0$. 
We consider the nearest neighbors baselines as it might be easier for a Transformer model to encode using self-attention compared to least squares.

\paragraph{Lasso.} We use this baseline for sparse linear functions (Section \ref{sec:beyond_linear}). This corresponds to $M(P) = \hat{w}^T x_{\text{query}}$, where $\hat{w}$ minimizes the $\ell_1$-norm regularized least squares objective:
\[
\min_{\hat{w}} ~ \frac{1}{2k}|| y - X\hat{w}||_2^2 ~ + ~ \alpha ||\hat{w}||_1.
\]
 We try different values of $\alpha \in \{ 1, 10^{-1}, 10^{-2}, 10^{-3}, 10^{-4} \}$, and report the best solution (achieving the smallest error with $10$ in-context examples) corresponding to $\alpha = 10^{-2}$. To solve the optimization problem, we use the Lasso implementation from Scikit-learn \citep{scikit-learn}
 ~\footnote{\scriptsize\url{https://scikit-learn.org/stable/modules/generated/sklearn.linear_model.Lasso.html}}.

\paragraph{Greedy Tree Learning.} We use this baseline for the class of decision trees. This corresponds to greedily learning a decision tree using the in-context examples, and using it to classify the query input. To construct the tree, at each node (starting from a root node), we choose a coordinate for partitioning the examples into two sets, so as to minimize the variance of $y_i$s in each set, averaged across the two sets. The value associated with a leaf node is the average $y_i$ value of the examples belonging to it. We use Scikit-learn's decision tree regressor \citep{scikit-learn}
~\footnote{\scriptsize\url{https://scikit-learn.org/stable/modules/tree.html\#regression}} implementation for this, with all the arguments set to their default value except the max\_depth argument which is set to 2. 
We considered values $\{1, 2, 3, 4, 5, 6, \text{unbounded}\}$ for the maximum depth and chose the value that performs best at 100 in-context examples which was 2 (which differs from the decision trees sampled from the function class which have depth 4).
We also considered a baseline where we learn this tree using only the signs of each $x_i$ coordinate---after all, the decision tree we are trying to learn depends only on the signs of $x_i$.
In this case, we found the optimal depth to be 4.
 
\paragraph{Tree boosting.} For the class of decision trees, we also consider a tree boosting baseline that corresponds to learning an ensemble of decision trees (see \citet{friedman2001greedy} for a description of the general framework).
Specifically, we use the XGBoost library~\citep{chen2016xgboost}~\footnote{\scriptsize\url{https://github.com/dmlc/xgboost}}, an implementation commonly used for a wide range of real-world machine learning tasks.

We performed a hyperpameter search by considering \{1, 2, 5, 10, 50, 100, 200, 400\} estimators in the ensemble (equivalent to number of boosting rounds), a learning rate of \{0.001, 0.01, 0.1, 0.3, 0.6, 1, 3\}, and a maximum depth of \{1, 2, 3, 4, 6, 10, 16\}.
In general, we found the performance of the learning algorithm to be quite robust.
We chose the hyperparameters obtaining the best performance with 100 training examples, corresponding to 50 estimators, a maximum depth of 4, and a learning rate f 0.1.
We found these hyperparameters to also be optimal when learning based on the signs of each $x_i$.

\paragraph{Learning neural networks with gradient descent.} We use this baseline for the class of two-layer neural networks (Section \ref{sec:beyond_linear}). This corresponds to training a two-layer neural network on the in-context examples, and outputting its prediction on the query point. That is, $M(P) = \hat{f}(x_\text{query})$, where 
\[
\hat{f}(x_\text{query}) = \sum_{i=1}^r ~ \hat{\alpha}_i ~ \sigma(\hat{w}_i^\top x_\text{query}).
\]
Here, $\sigma(\cdot)$ is the ReLU activation.
We find parameters $\hat{\alpha}_i, \hat{w}_i$ by minimizing the squared error of the prediction for the in-context examples
\[
\sum_{i=1}^k \left(\hat{f}(x_i) - f(x_i)\right)^2,
\]
using the Adam optimizer. 
We use a batch size of 10 (we use full batch when the number of in-context examples is less than 10) with 5000 optimization steps, and set $r = 100$. We use a learning rate of $5\cdot 10^{-3}$ in the case when the data is generated using a neural network, and a learning rate of $5\cdot 10^{-2}$ when the data is generated using a linear function. We consider the setting with $100$ in-context examples, and do a hyperparameter grid search over learning rate $\in \{5\cdot 10^{-4}, 5\cdot 10^{-3}, 5\cdot 10^{-2}, 5\cdot 10^{-1}, 5\}$, $r \in \{100, 400\}$, batch size $\in \{10, 100\}$, optimization algorithm $\in \{\text{adam}, \text{sgd}\}$. All the hyperparameter settings in this grid led to a similar or worse performance compared to the hyperparameter setting we choose. 



\clearpage
\section{Additional experimental results}
\label{app:extra}

\subsection{Robustness to query scale}
\label{app:query}
\begin{figure}[htp]
        \centering
        \includegraphics[width=.6\textwidth]{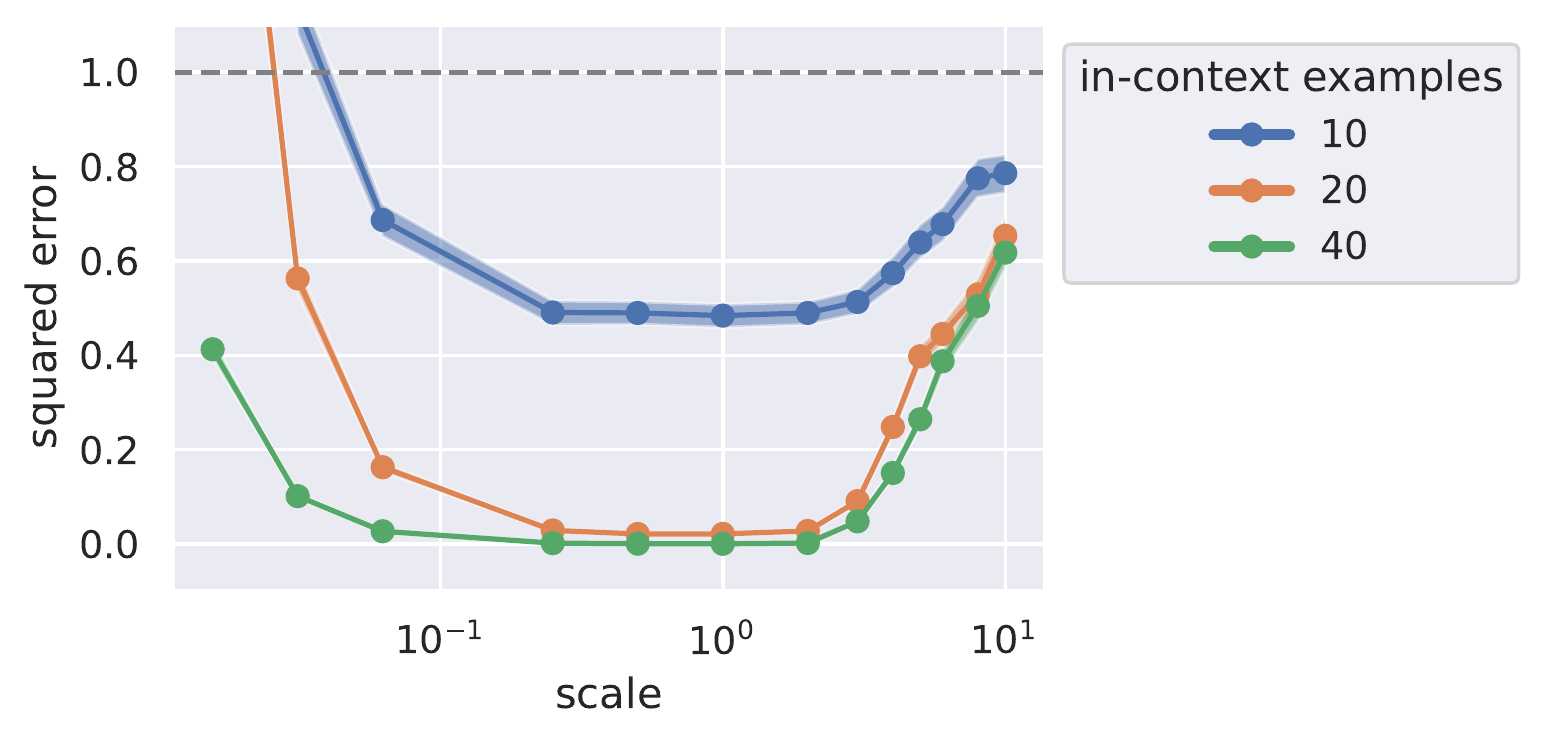}
    \caption{\emph{Robustness to the scale of query input.} For a fixed set of in-context examples, we measure the model's error as we scale the query input by a scalar.}
    \label{fig:query_scale}
\end{figure}

In Figure~\ref{fig:query_scale}, we show that the trained model is quite robust to scaling the query input (while keeping the in-context examples fixed): the error does not increase much as we scale up the query
input by a factor of up to 2, or scale down by a factor of up to 16, and degrades slowly after that.

\clearpage
\subsection{Out-of-distribution prompts}
\label{app:dist_shift}
\begin{figure}[htp]
    \begin{subfigure}[b]{0.33\textwidth}
        \centering
        \includegraphics[height=.75\textwidth]{figures/skewed.pdf}
        \caption{skewed covariance}
        \label{fig:skewed_app}
    \end{subfigure} 
    \begin{subfigure}[b]{0.33\textwidth}
        \centering
        \includegraphics[height=.75\textwidth]{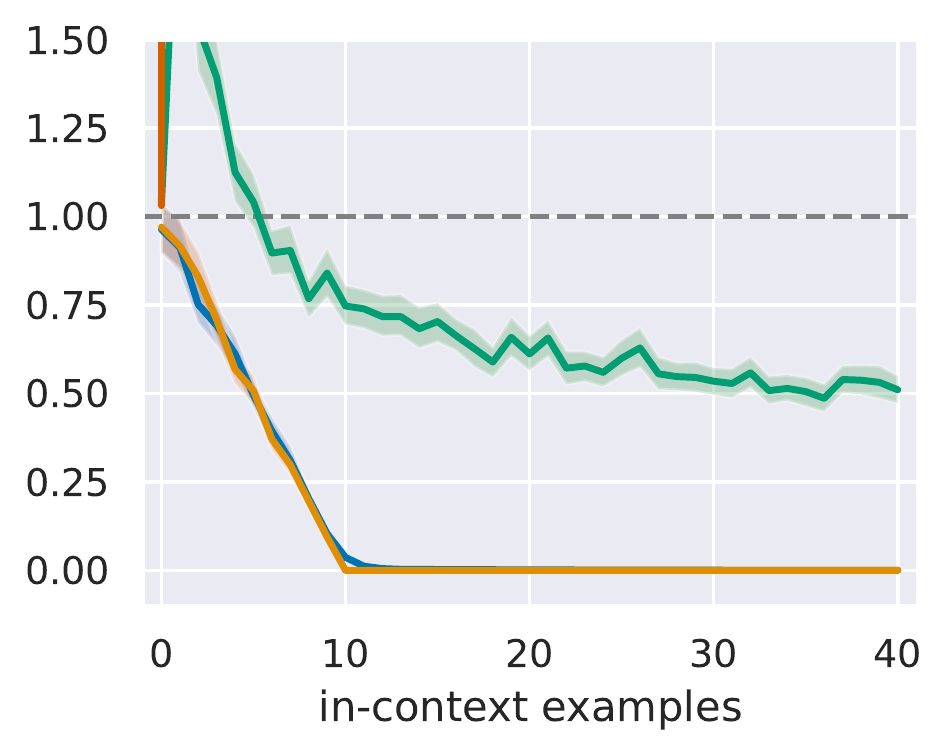}
        \caption{$d$/2-dimensional subspace}
        \label{fig:half_app}
    \end{subfigure}  
    \begin{subfigure}[b]{0.33\textwidth}
        \centering
        \includegraphics[height=.75\textwidth]{figures/noisyLR.pdf}
        \caption{noisy output}
        \label{fig:noisy_app}
    \end{subfigure} \\[.4em]
    
    \begin{subfigure}[b]{0.33\textwidth}
        \centering
        \includegraphics[height=.75\textwidth]{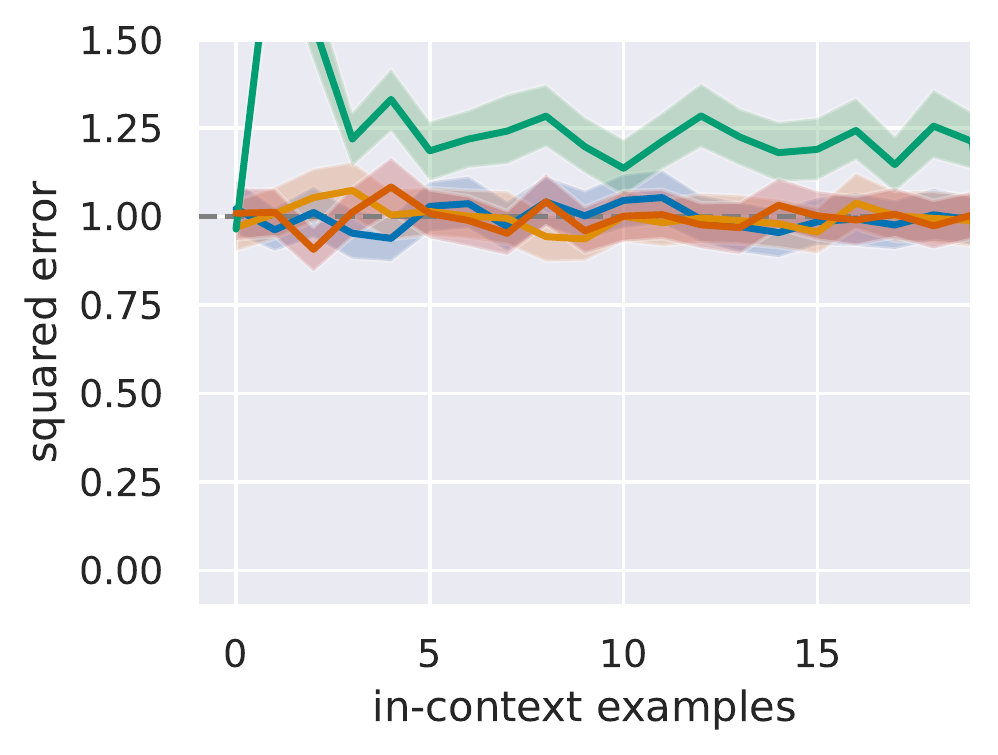}
        \caption{orthogonal query}
        \label{fig:orthogonal_app}
    \end{subfigure}
    \begin{subfigure}[b]{0.33\textwidth}
        \centering
        \includegraphics[height=.75\textwidth]{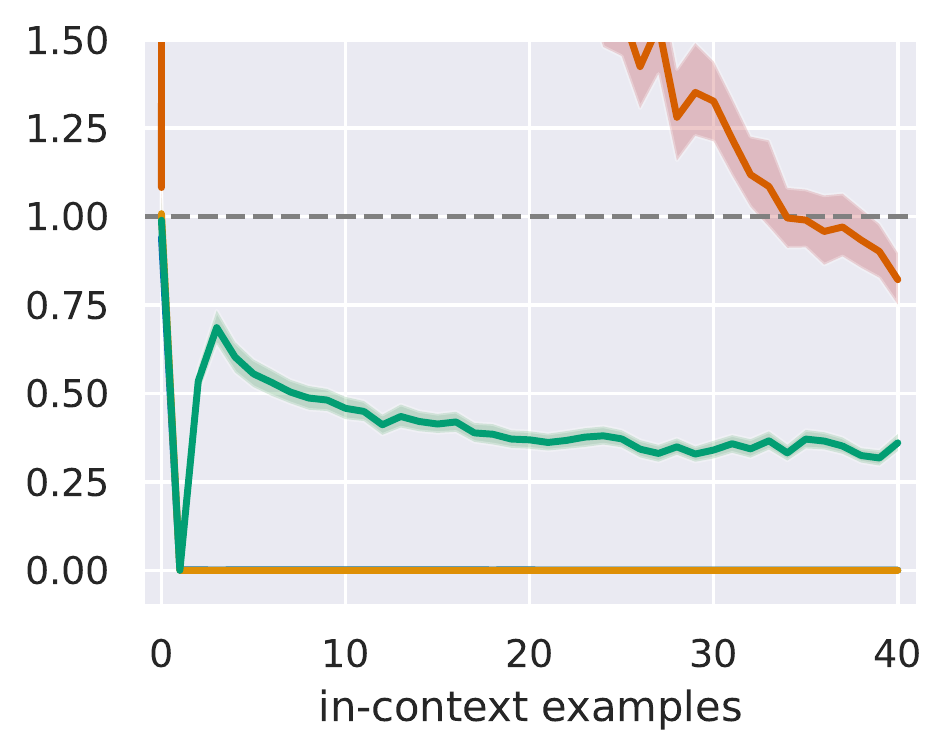}
        \caption{query matches in-context example}
        \label{fig:overlapping_app}
    \end{subfigure}
        \begin{subfigure}[b]{0.33\textwidth}
        \centering
        \includegraphics[height=.75\textwidth]{figures/random.pdf}
        \caption{different orthants}
        \label{fig:random_app}
    \end{subfigure}
    \caption{\emph{In-context learning on out-of-distribution prompts.}
    We evaluate the model trained to in-context learn linear functions on prompt distribution  that deviates from the training prompt distribution. 
    In general, the model error degrades gracefully and closely tracks that of the least squares estimator.
    }
    \label{fig:ood_prompts}
\end{figure}

Here, we describe the structure of our out-of-distribution prompts (cf.\  Section~\ref{sec:distribution_shifts}), and show the corresponding plots (Figure~\ref{fig:ood_prompts}). 
To avoid conflating factors, we normalize the prompt inputs such that their expected norm is equal to the expected norm of inputs during training and investigate the role of scaling these inputs separately.
We summarize how these prompts deviate from those seen during training in the table below.
\begin{center}
    \begin{tabular}{lccc}
    \toprule
        Prompting strategy
        & $D_\mathcal{X}^{\text{train}} \neq D_\mathcal{X}^{\text{test}}$
        & $D_\mathcal{F}^{\text{train}} \neq D_\mathcal{F}^{\text{test}}$
        & $D_{\text{query}}^\text{test} \neq D_\mathcal{X}^{\text{test}}$ \\
    \midrule
        Skewed covariance & \checkmark \\
        $d/2$-dimensional subspace  & \checkmark \\
        Scale inputs & \checkmark \\
    \midrule
        Noisy output &  & \checkmark \\
        Scale weights & & \checkmark \\
    \midrule
        Different Orthants & \checkmark & & \checkmark \\
        Orthogonal query & & & \checkmark \\
        Query matches example & & & \checkmark \\
    \bottomrule
    \end{tabular}
\end{center}

\paragraph{Skewed covariance.} (Figure \ref{fig:skewed_app}) We sample inputs from $N(0, \Sigma)$ where $\Sigma$ is a skewed covariance matrix with eigenbasis chosen uniformly at random and $i^{\text{th}}$ eigenvalue proportional to $\nicefrac{1}{i^2}$.

\paragraph{Low-dimensional subspace.} (Figure \ref{fig:half_app}) We sample prompt inputs from a random $d/2$ dimensional subspace. That is, we pick a random $d/2$ dimensional subspace, and draw the prompt inputs from an isotropic Gaussian distribution restricted to this subspace.
As a result, it is possible to achieve zero error after $d/2$ in-context examples.

\begin{figure}[htp]
\centering
    \begin{subfigure}[b]{0.42\textwidth}
        \centering
        \includegraphics[height=.7\textwidth]{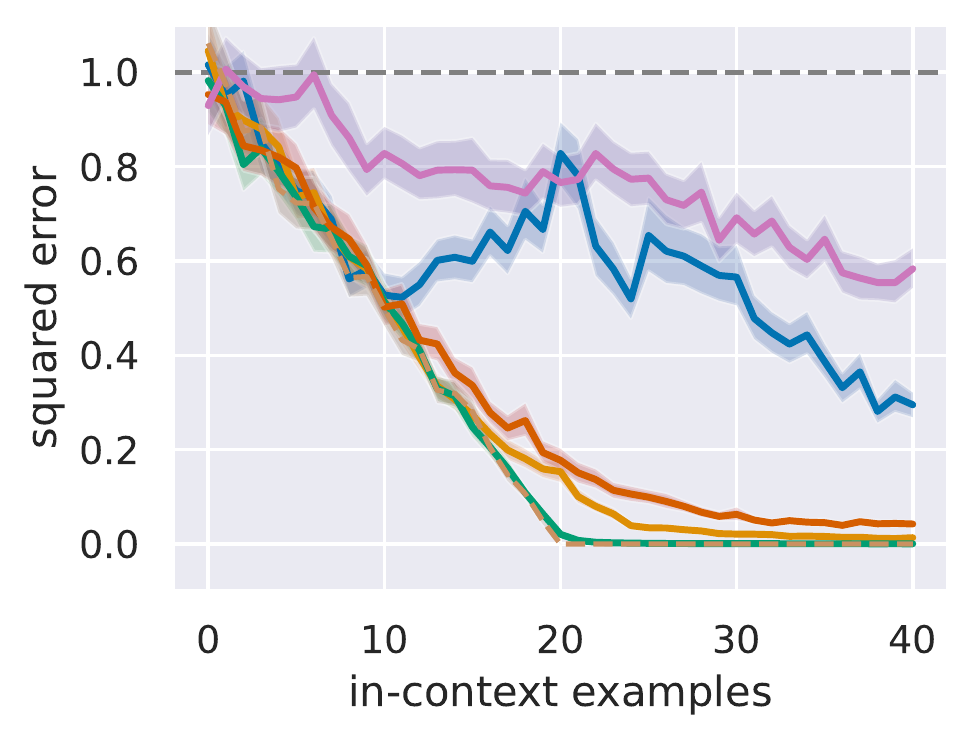}
        \caption{scaled $x$, Transformer}
        \label{fig:scalex_transformer_app}
    \end{subfigure}
    \begin{subfigure}[b]{0.42\textwidth}
        \centering
        \includegraphics[height=.7\textwidth]{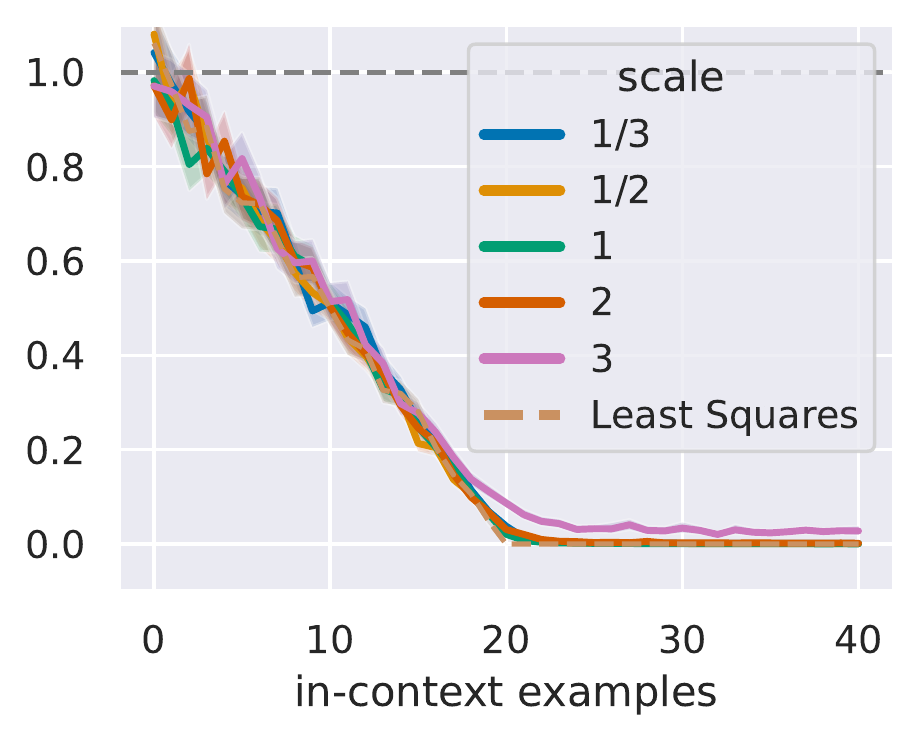}
        \caption{scaled $w$, Transfomer}
        \label{fig:scaley_transformer_app}
    \end{subfigure}
    \caption{\emph{In-context learning robustness to prompt scaling.}
    We evaluate the model trained to in-context learn linear functions when we scaled the prompt inputs $x$ or the weight of the function class $w$. 
    The model appear to be quite robust to scaling $w$ but their performance degrades when scaling the inputs up or down by a factor of 3.}
    \label{fig:scale_app}
\end{figure}

\paragraph{Prompt scale.} (Figure \ref{fig:scale_app}) We consider the setting where the prompt scale between training and inference is different. We either scale the prompt inputs or the weight vectors, by a factor $\{1/3, 1/2,  2, 3\}$.

\paragraph{Noisy linear regression.} (Figure \ref{fig:noisy_app}) We add noise to each prompt output, that is, the $i^{\text{th}}$ output is equal to $w^T x_i + \epsilon_i$ where $\epsilon_i \sim N(0, d/20)$. 

\paragraph{Different orthants for in-context and query inputs.}(Figure \ref{fig:random_app}) We fix the sign of each coordinate to be positive or negative for all in-context inputs $x_i$ (at random), and draw $x_\text{query}$ (as before) i.i.d. from $N(0, I_d)$. 
As a result, all in-context inputs lie in the same orthant, while the query input lies in another orthant with high probability.

\paragraph{Query input orthogonal to in-context inputs.}(Figure \ref{fig:orthogonal_app}) We choose the query input randomly in the space orthogonal to the space spanned by in-context example inputs.
That is, we  draw the query input from an isotropic Gaussian distribution restricted to the subspace orthogonal to the space spanned by the in-context examples. 
Thus, the optimal normalized error is 1 for any number of in-context examples (there can be at most $d-1$ in-context examples for an orthogonal query to exist).

\paragraph{Query input matches an in-context example.}(Figure \ref{fig:overlapping_app})
We set the query input equal to one of the in-context examples chosen uniformly at random.
Thus it's possible to achieve zero error since the in-context examples include the correct prediction for the query input already.

\clearpage
\subsection{Effect of problem dimension and model capacity}
\label{app:capacity}
\begin{figure}[htp]
    \centering
    \includegraphics[width=.4\textwidth]{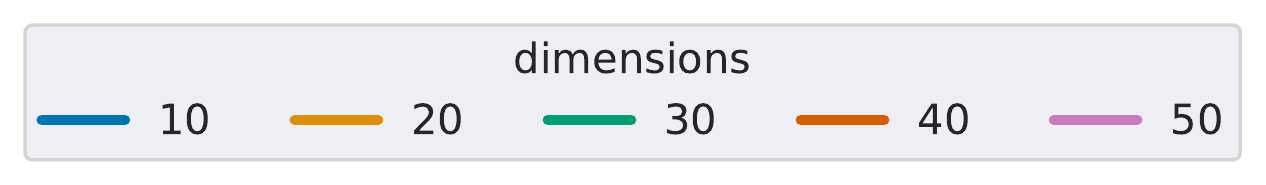}
    
    \centering
    \begin{subfigure}[b]{0.325\textwidth}
        \centering
        \includegraphics[height=.8\textwidth]{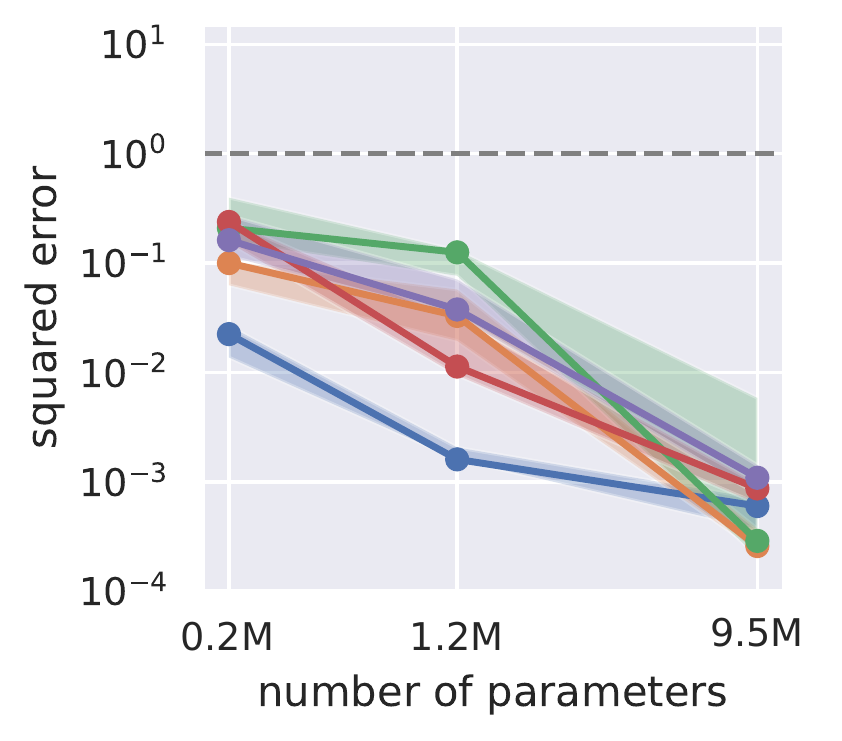}
        \caption{$d/2$-dimensional subspace}
        \label{fig:capacity_half}
    \end{subfigure} 
    \begin{subfigure}[b]{0.325\textwidth}
        \centering
        \includegraphics[height=.8\textwidth]{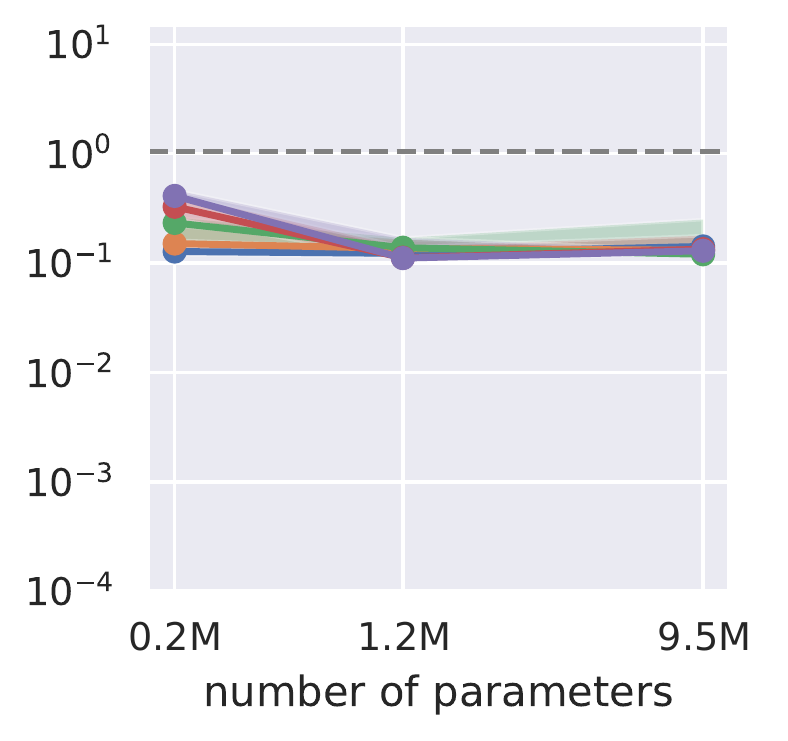}
        \caption{noisy output}
        \label{fig:capacity_noisy}
    \end{subfigure} 
    \begin{subfigure}[b]{0.325\textwidth}
        \centering
        \includegraphics[height=.8\textwidth]{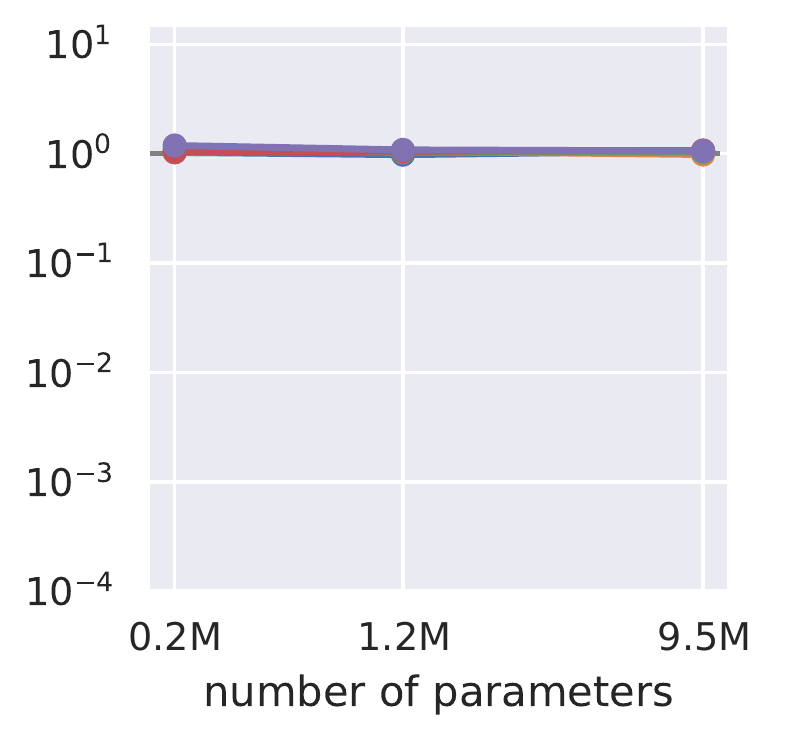}
        \caption{orthogonal query}
        \label{fig:capacity_orthogonal}
    \end{subfigure}
    
    \centering
    \begin{subfigure}[b]{0.325\textwidth}
        \centering
        \includegraphics[height=.8\textwidth]{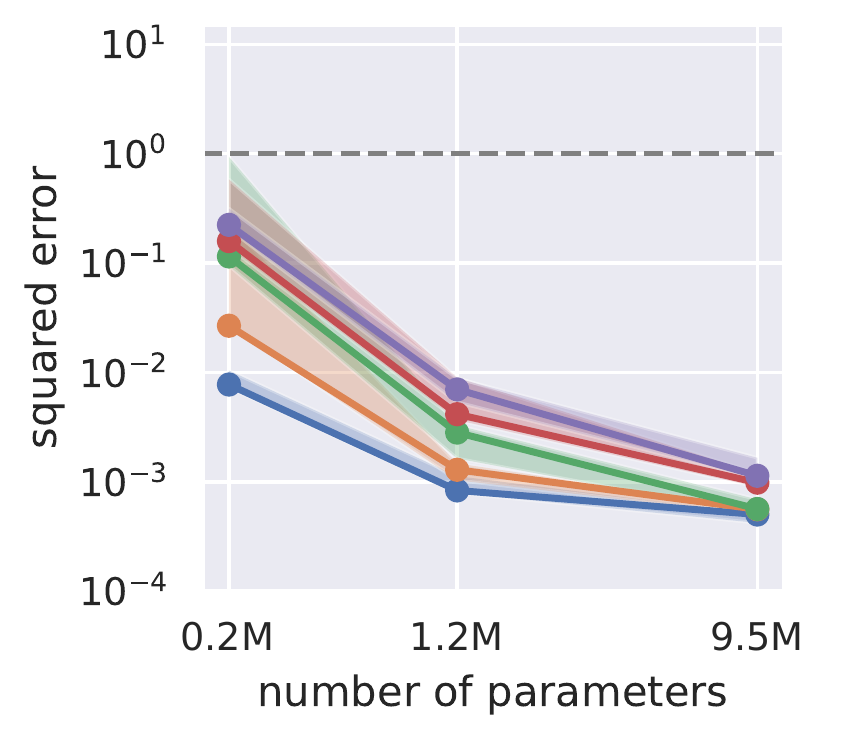}
        \caption{query matches in-context examples}
        \label{fig:capacity_overlapping}
    \end{subfigure}
    \begin{subfigure}[b]{0.325\textwidth}
        \centering
        \includegraphics[height=.8\textwidth]{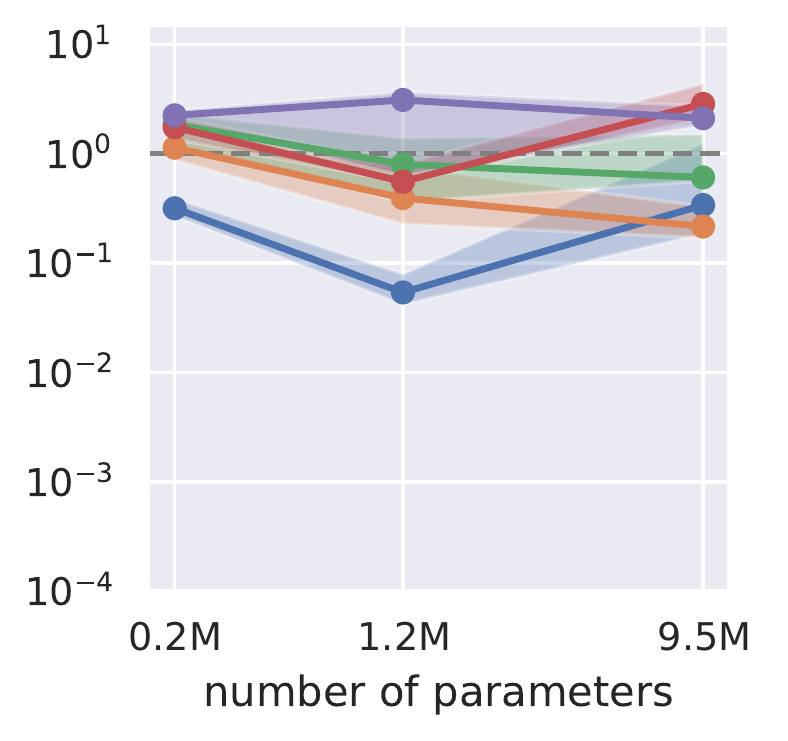}
        \caption{scale $x$ by a factor of 2}
        \label{fig:capacity_scale_x_2}
    \end{subfigure}
    \begin{subfigure}[b]{0.325\textwidth}
        \centering
        \includegraphics[height=.8\textwidth]{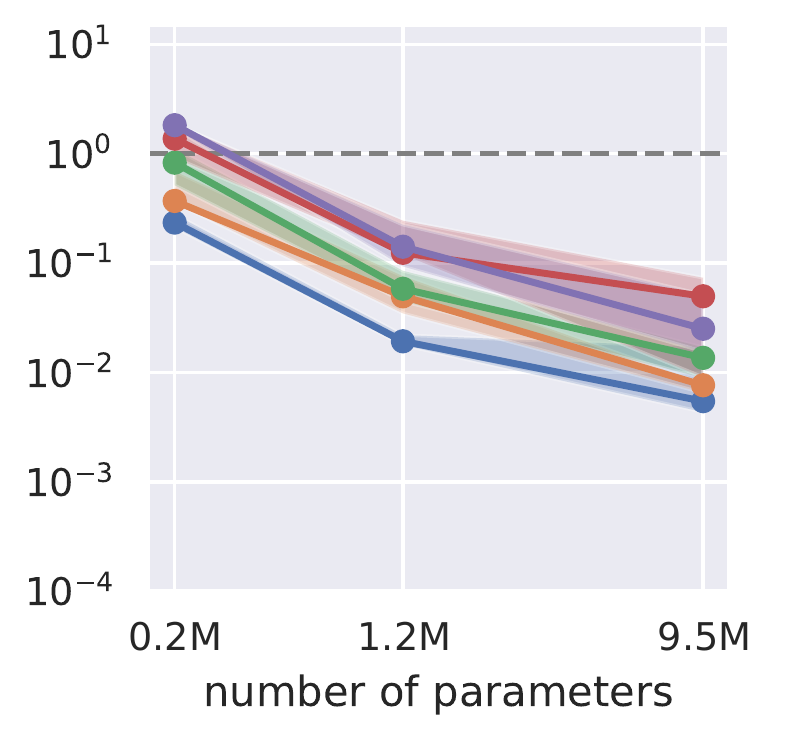}
        \caption{scale $w$ by a factor of 2}
        \label{fig:capacity_scale_y_2}
    \end{subfigure}
    \hfill
    \phantom{}
      \caption{\emph{The effect of model capacity and problem dimension for in-context learning performance on out-of-distribution prompts.} We train Transformer models of varying capacity to in-context learn linear function in varying dimensions $d$. We plot the error with $2d$ in-context examples (or $d-1$ for orthogonal queries). We find that capacity helps in most cases, with the exception of scaling $x$ where we find no clear trend. For each setting, we train 3 models with different random seeds, and show the median error (solid lines), and the minimum and maximum errors (shaded region). (See Figures~\ref{fig:cap_random},~ \ref{fig:cap_skewed} in the main text for the corresponding plots on different-orthants and skewed-covariance.)}
    \label{fig:capacity_app}
\end{figure}

We plot the model error for additional out-of-distribution prompts in Figure~\ref{fig:capacity_app} for $2d$ in-context examples (with the exception of orthogonal queries where we use $d-1$ in-context examples). 

Similar to the settings in Section \ref{sec:factors} (skewed covariance and different orthants), accuracy improves with capacity in most cases. One exception is scaling $x$ (Figure \ref{fig:capacity_scale_x_2}), in which case we do not see any clear trend. In the case of noisy output (Figure \ref{fig:capacity_noisy}), the accuracy almost saturates at 1.2M parameters, close to the error of the least squares estimator.
In the case of orthogonal query input (Figure \ref{fig:capacity_orthogonal}), the model achieves the optimum error of $1$ even with the tiny model with 0.2M parameters.

\clearpage
\subsection{Training variance}
\label{app:variance}
In Figure~\ref{fig:trials}, we show the variance in error across training runs for the standard Transformer model (9.5M parameters). We plot the squared error for 3 models (with different random seeds) each for $d \in\{10, 20, 30, 40, 50\} $, trained to in-context learn linear functions.
The error is quite concentrated in the standard setting as well as for most out-of-distribution prompts. In the different-orthants and skewed-covariance settings, we observe a high variance for higher dimensional problems $(d \geq 30)$. However, in Section \ref{sec:factors}, we saw that the error in these settings usually decreases as we increase the model size. 
In the setting where we scale $x$, there is high variance even when $d=10$.

\begin{figure}[htp]
    \begin{subfigure}[b]{\textwidth}
        \centering
        \includegraphics[width=.8\textwidth]{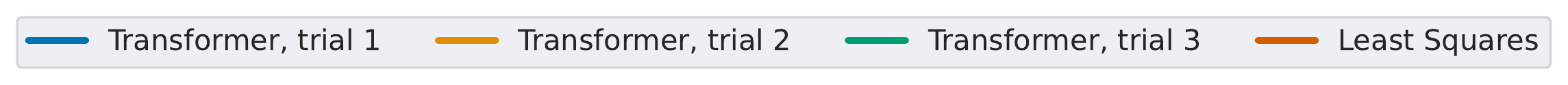}
        
        \includegraphics[height=.18\textwidth]{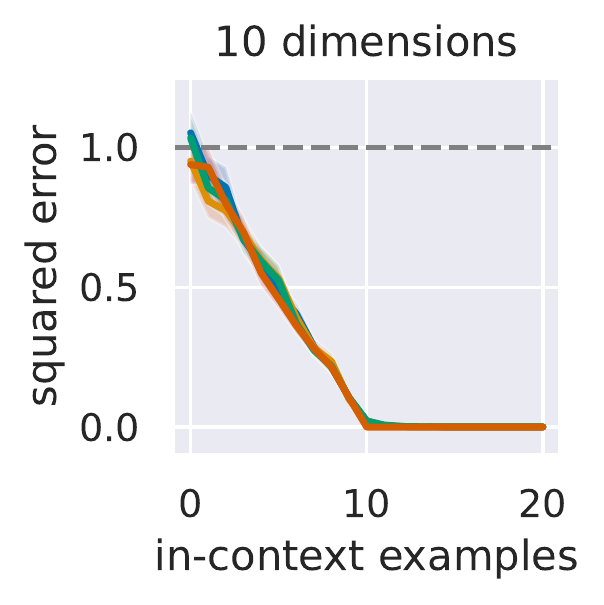}
        \includegraphics[height=.18\textwidth]{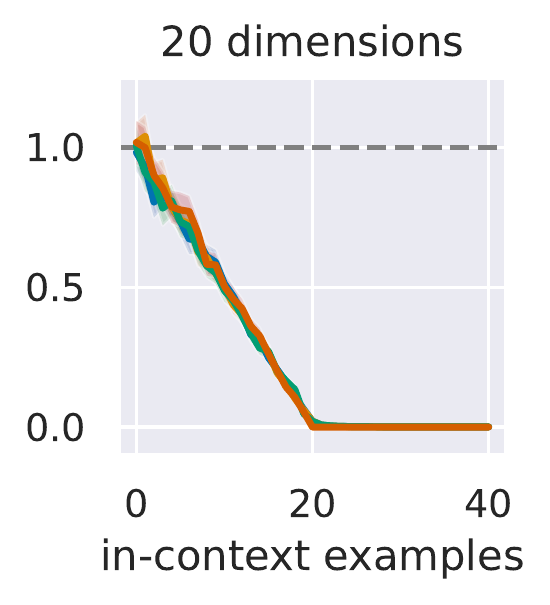}
        \includegraphics[height=.18\textwidth]{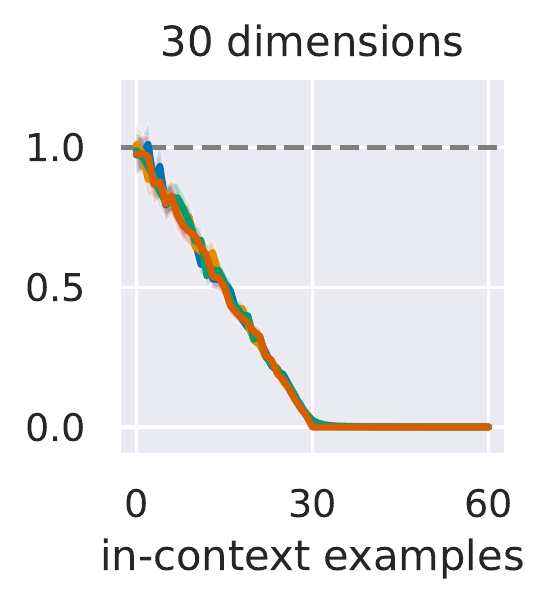}
        \includegraphics[height=.18\textwidth]{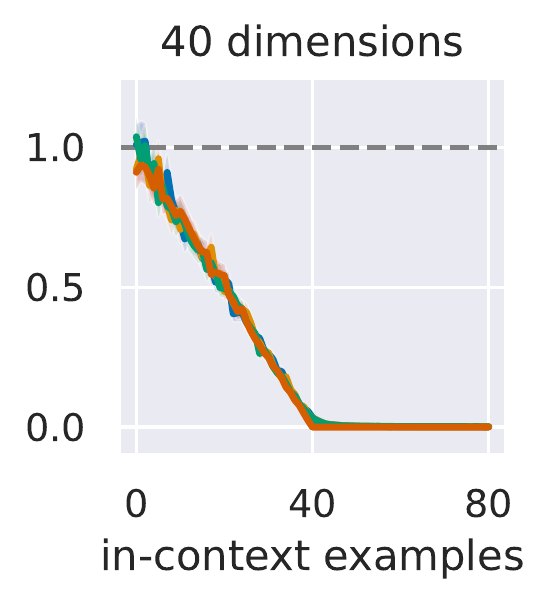}
        \includegraphics[height=.18\textwidth]{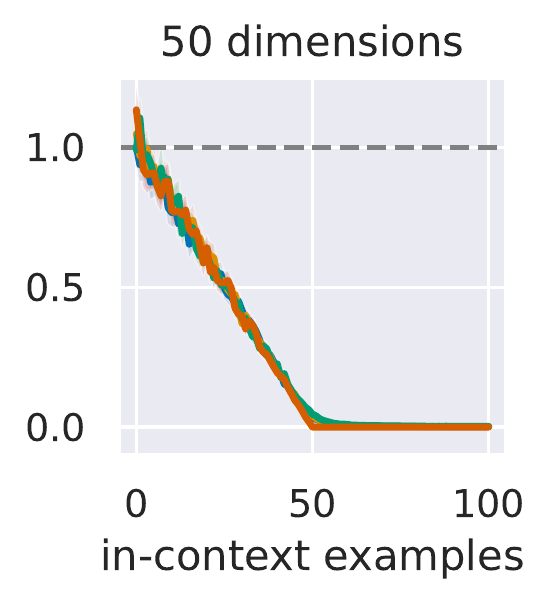}
        \caption{standard}
        \label{fig:trial_standard}
    \end{subfigure} \\[.6em]
    \begin{subfigure}[b]{\textwidth}
        \centering
        \includegraphics[height=.18\textwidth]{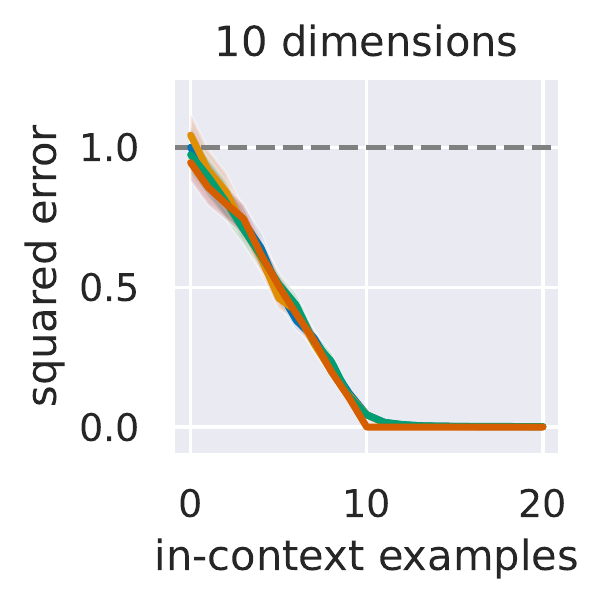}
        \includegraphics[height=.18\textwidth]{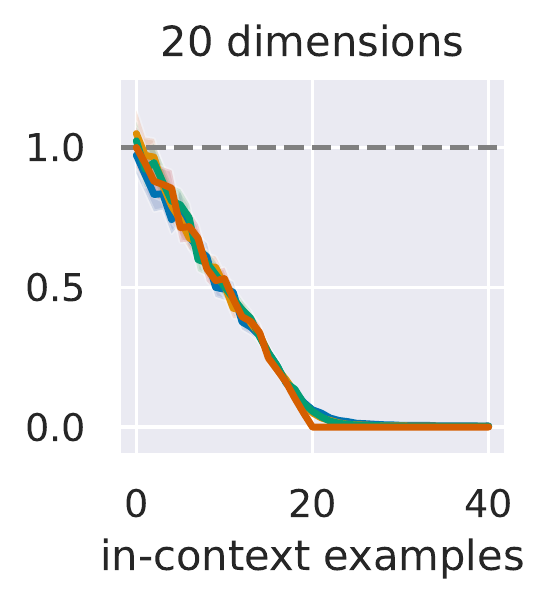}
        \includegraphics[height=.18\textwidth]{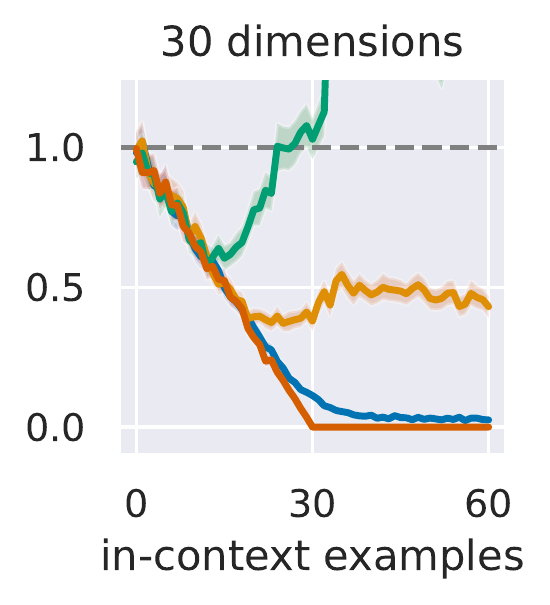}
        \includegraphics[height=.18\textwidth]{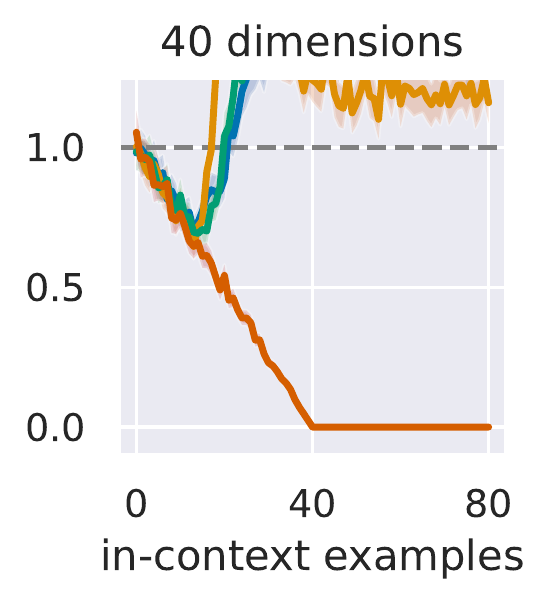}
        \includegraphics[height=.18\textwidth]{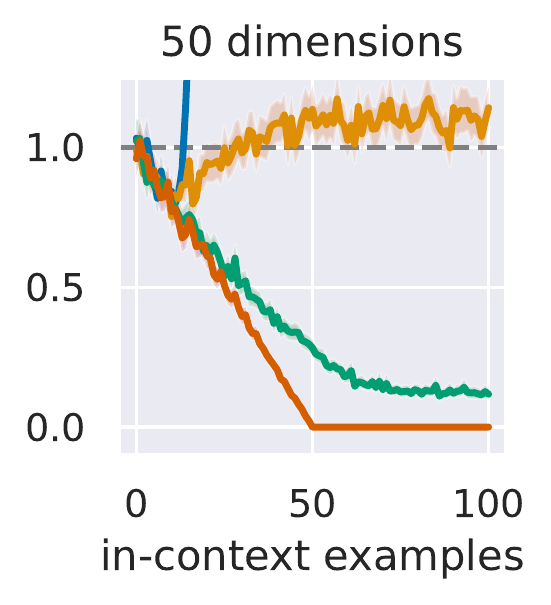}
        \caption{different orthants}
        \label{fig:trial_random}
    \end{subfigure} \\[.6em]
    \begin{subfigure}[b]{\textwidth}
        \centering
        \includegraphics[height=.18\textwidth]{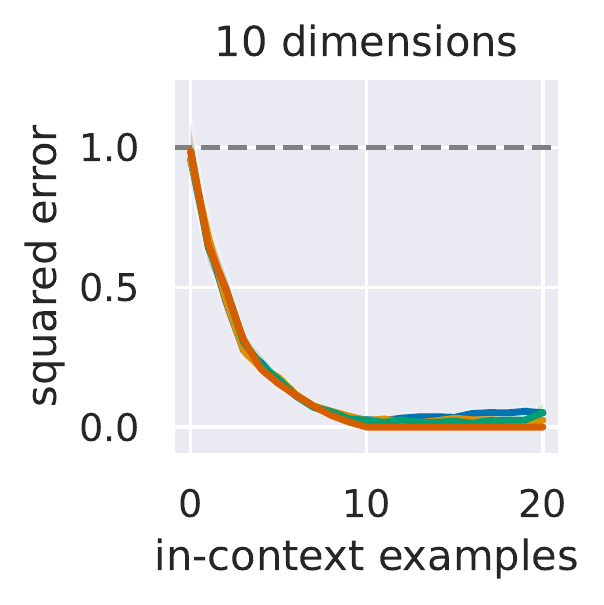}
        \includegraphics[height=.18\textwidth]{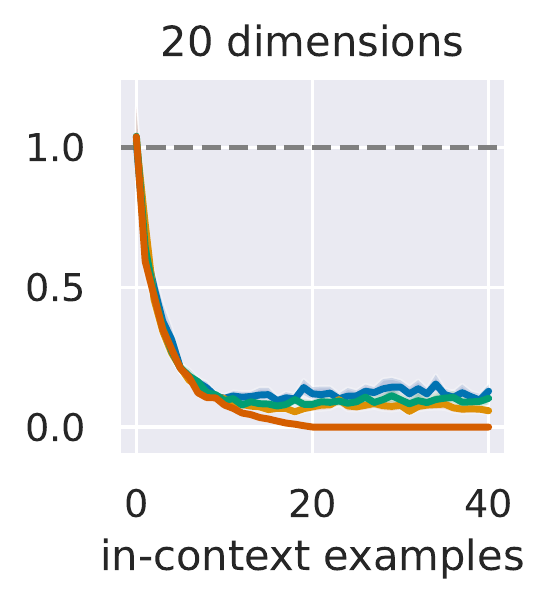}
        \includegraphics[height=.18\textwidth]{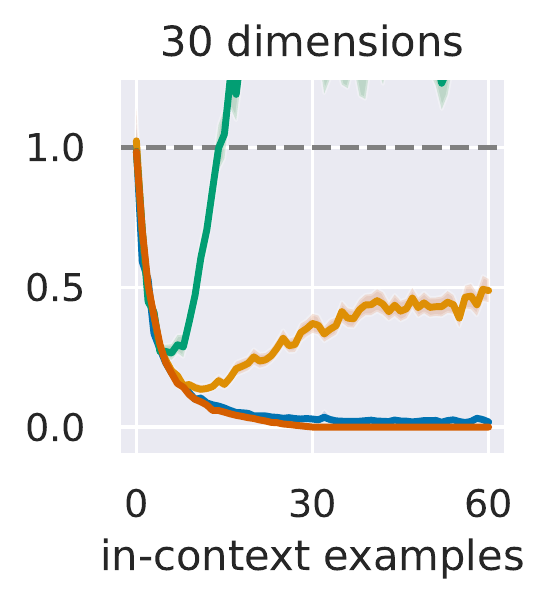}
        \includegraphics[height=.18\textwidth]{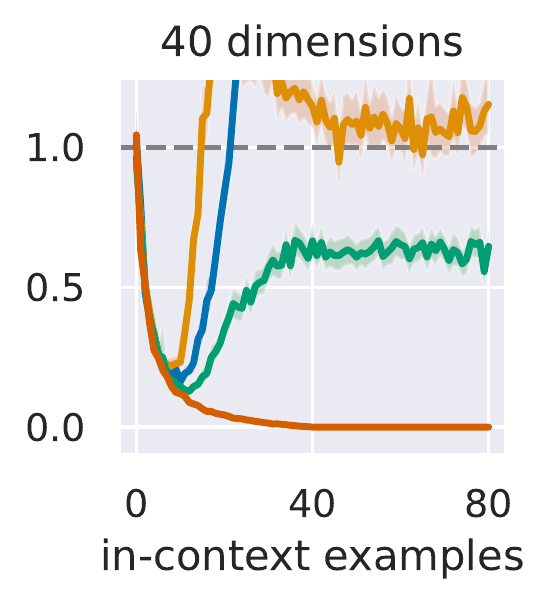}
        \includegraphics[height=.18\textwidth]{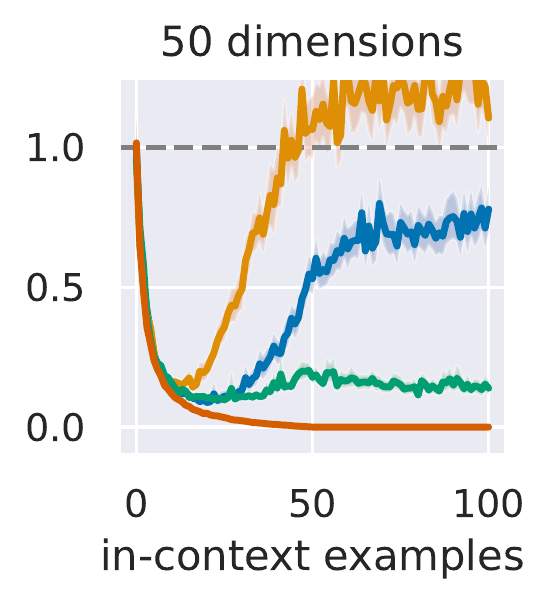}
        \caption{skewed covariance}
        \label{fig:trial_skewed}
    \end{subfigure} \\[.6em]
    \begin{subfigure}[b]{\textwidth}
        \centering
        \includegraphics[height=.18\textwidth]{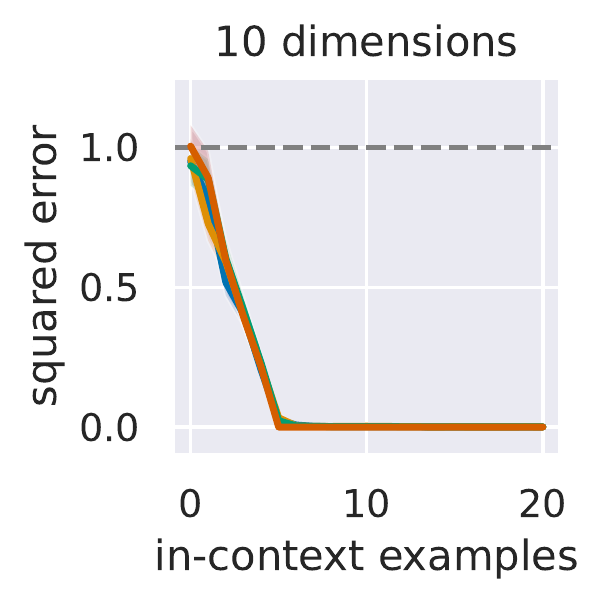}
        \includegraphics[height=.18\textwidth]{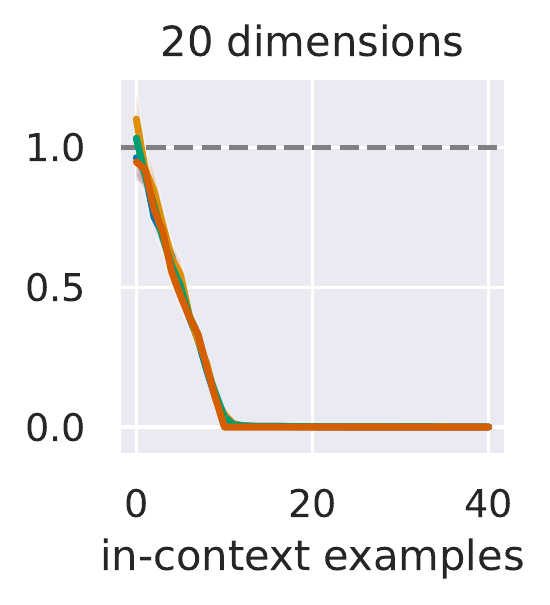}
        \includegraphics[height=.18\textwidth]{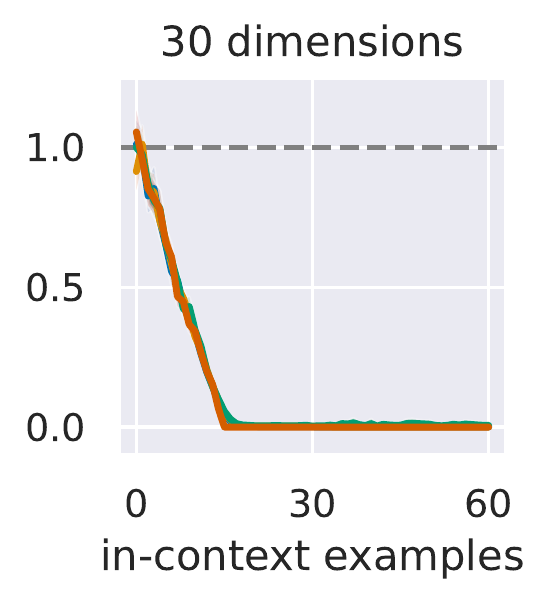}
        \includegraphics[height=.18\textwidth]{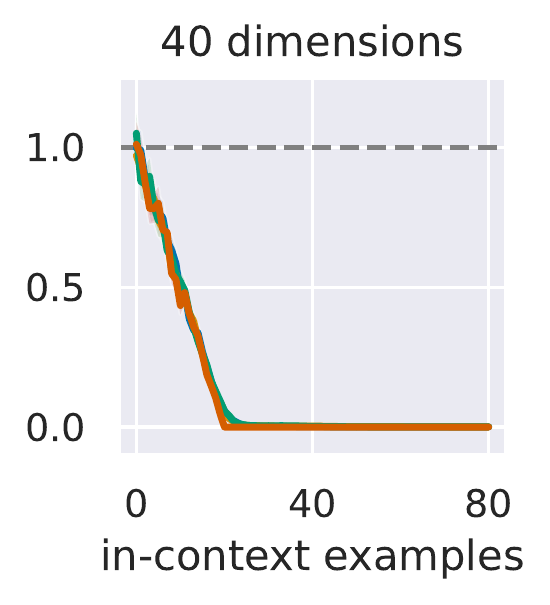}
        \includegraphics[height=.18\textwidth]{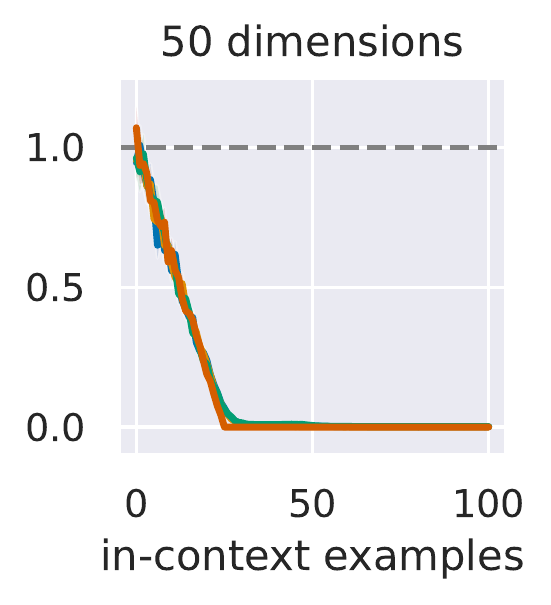}
        \caption{$d/2$-dimensional subspace}
        \label{fig:trial_half}
    \end{subfigure}
    \caption{\emph{Errors for models trained with different random seeds.} For each dimension, we train three models with different random seeds and show the corresponding error curves.}
    \end{figure}
    \begin{figure}[htp]\ContinuedFloat
    \begin{subfigure}[b]{\textwidth}
        \centering
        \includegraphics[width=.8\textwidth]{figures/trials/trial_legend.pdf}
        
        \includegraphics[height=.18\textwidth]{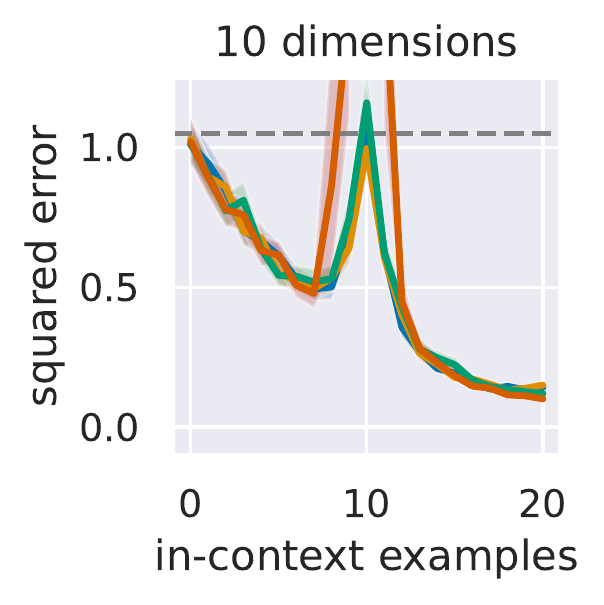}
        \includegraphics[height=.18\textwidth]{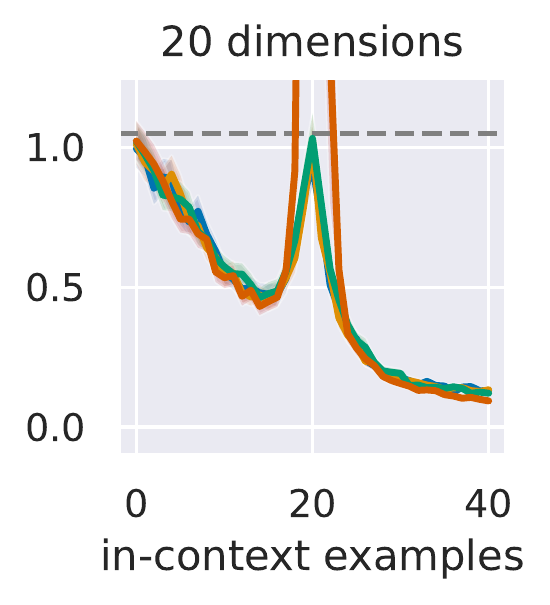}
        \includegraphics[height=.18\textwidth]{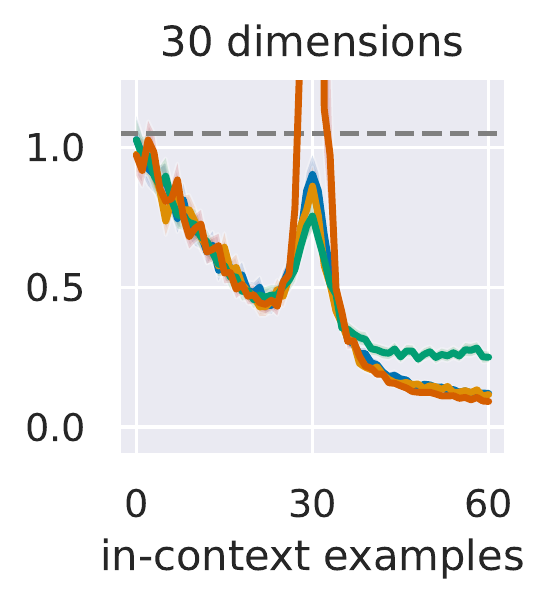}
        \includegraphics[height=.18\textwidth]{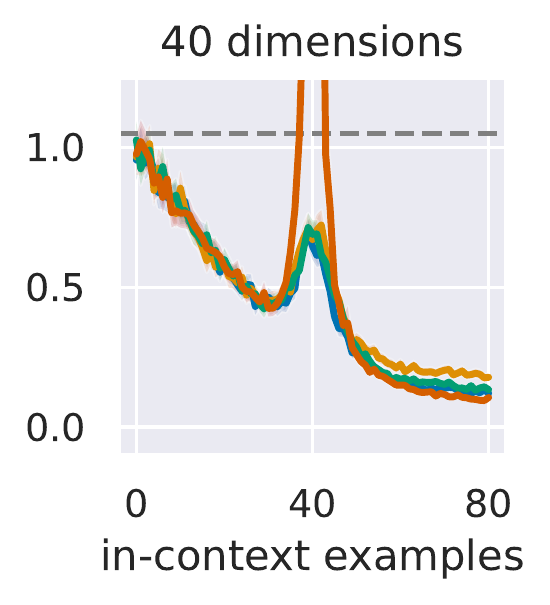}
        \includegraphics[height=.18\textwidth]{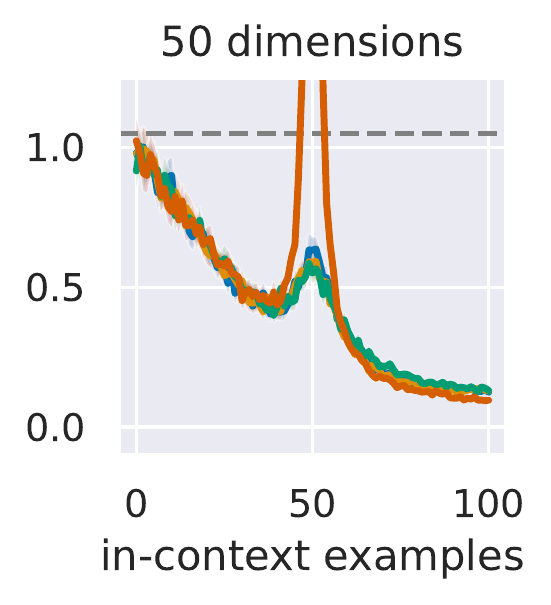}
        \caption{noisy output}
        \label{fig:trial_noisy}
    \end{subfigure} \\[.7em]
    \begin{subfigure}[b]{\textwidth}
        \centering
        \includegraphics[height=.18\textwidth]{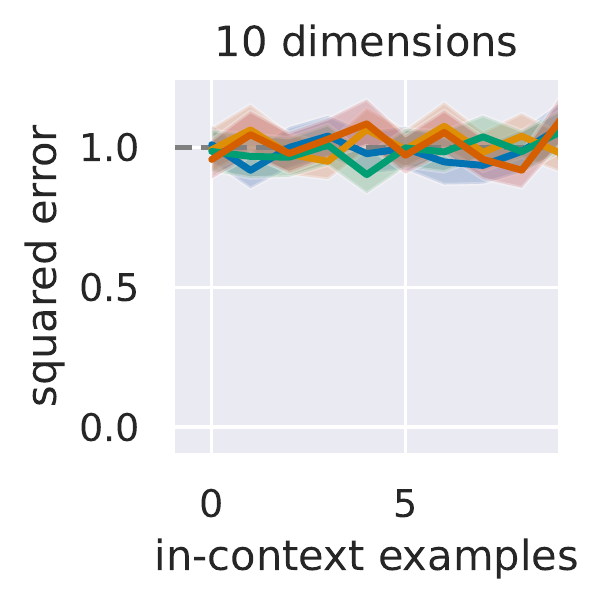}
        \includegraphics[height=.18\textwidth]{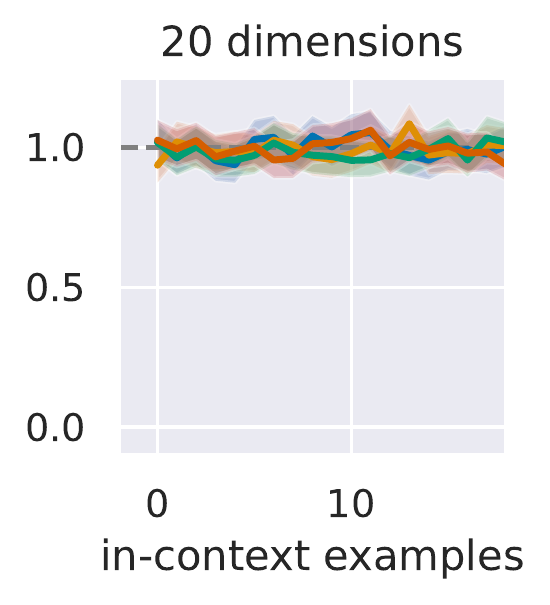}
        \includegraphics[height=.18\textwidth]{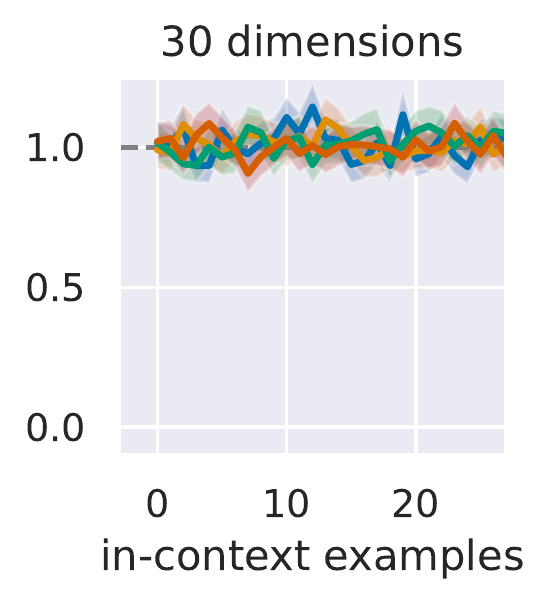}
        \includegraphics[height=.18\textwidth]{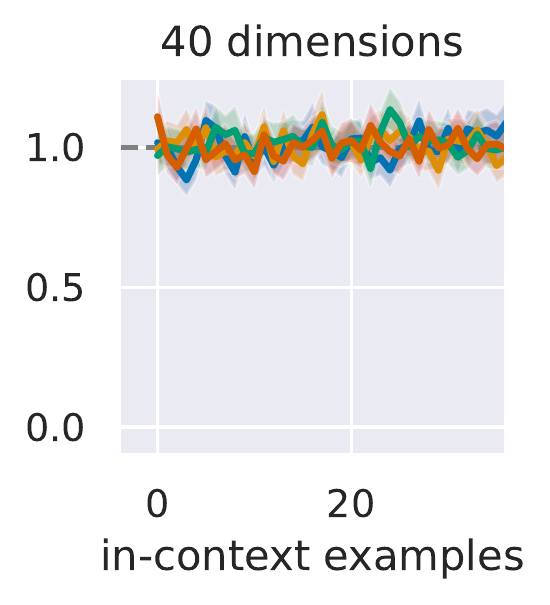}
        \includegraphics[height=.18\textwidth]{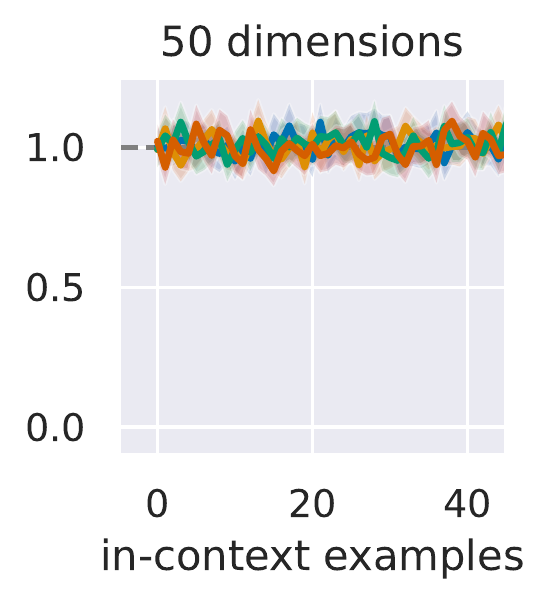}
        \caption{orthogonal query}
        \label{fig:trial_orthogonal}
    \end{subfigure} \\[.7em]
    \begin{subfigure}[b]{\textwidth}
        \centering
        \includegraphics[height=.18\textwidth]{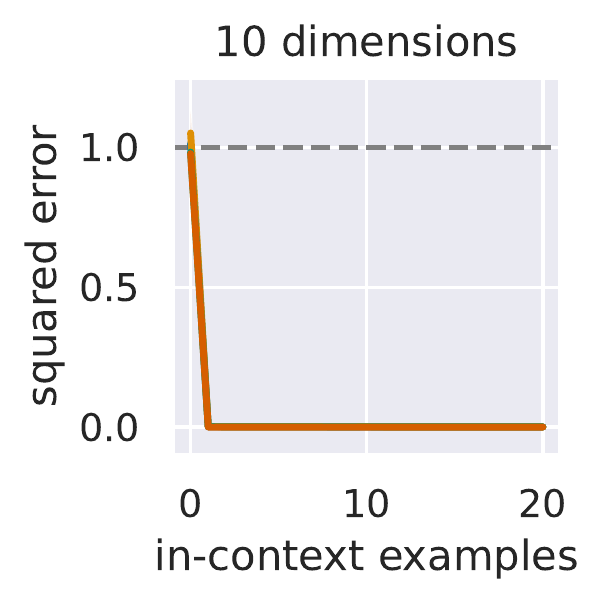}
        \includegraphics[height=.18\textwidth]{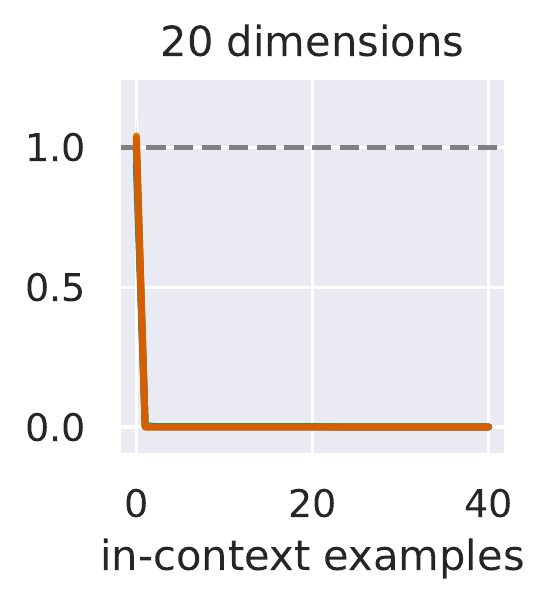}
        \includegraphics[height=.18\textwidth]{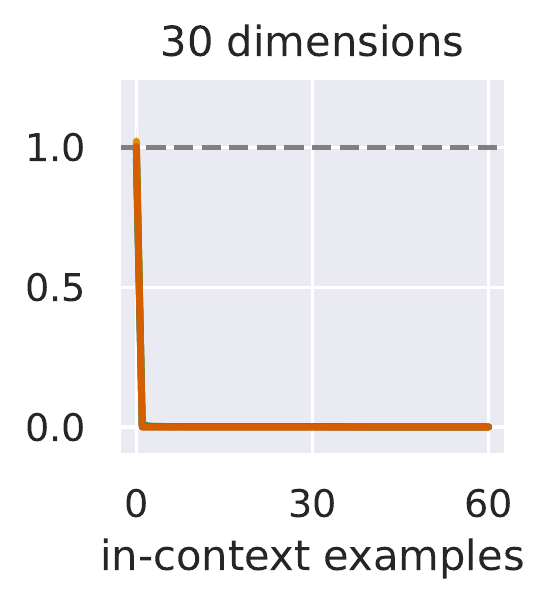}
        \includegraphics[height=.18\textwidth]{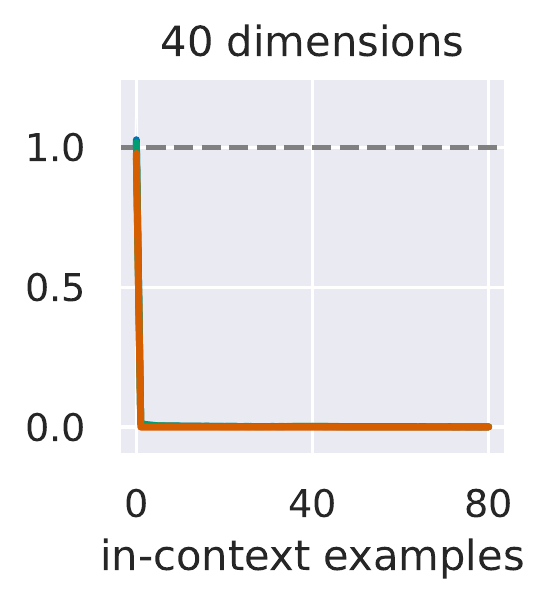}
        \includegraphics[height=.18\textwidth]{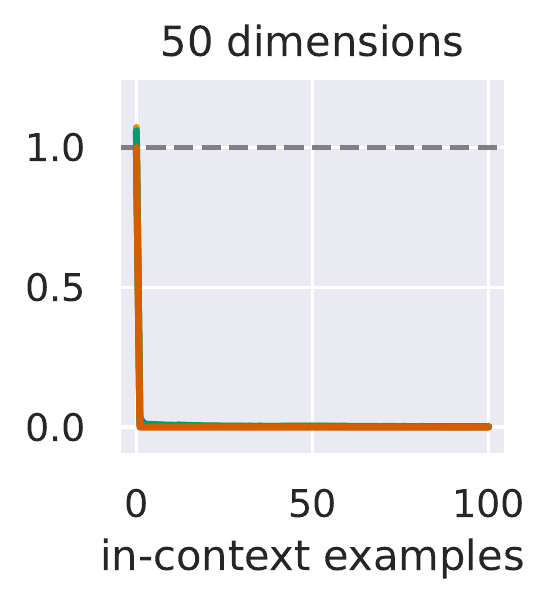}
        \caption{query matches in-context example}
        \label{fig:trial_overlapping}
    \end{subfigure} \\[.7em]
    \begin{subfigure}[b]{\textwidth}
        \centering
        \includegraphics[height=.18\textwidth]{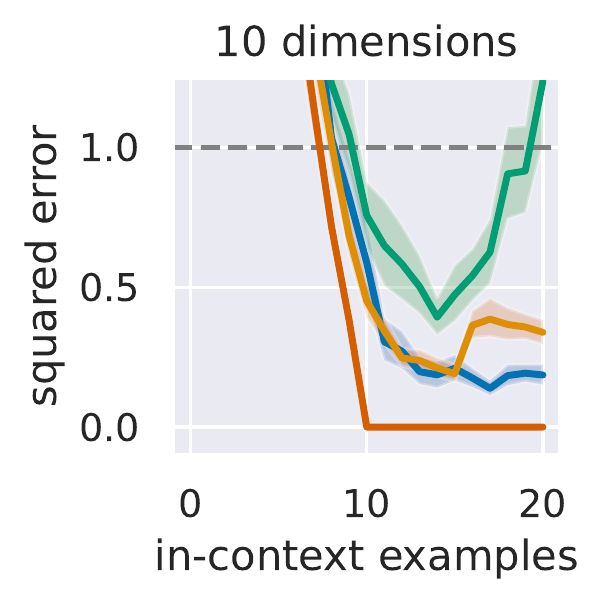}
        \includegraphics[height=.18\textwidth]{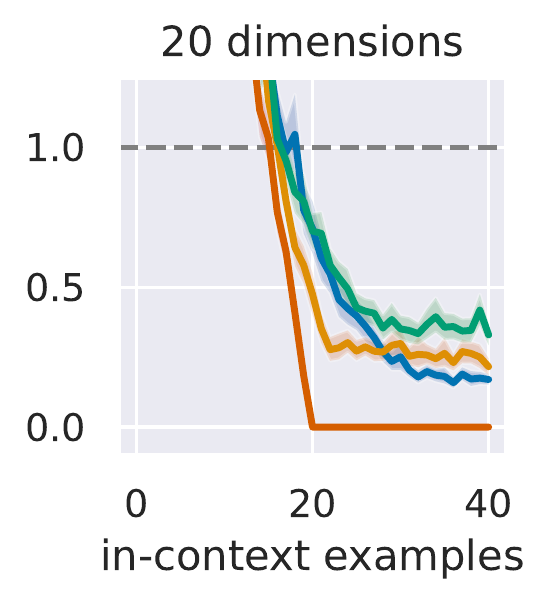}
        \includegraphics[height=.18\textwidth]{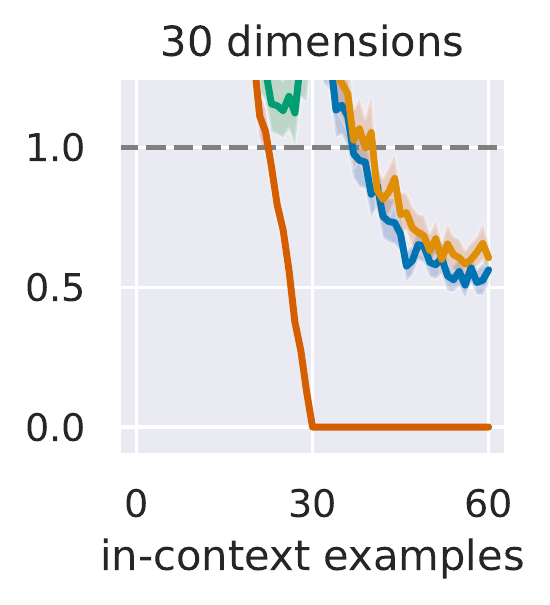}
        \includegraphics[height=.18\textwidth]{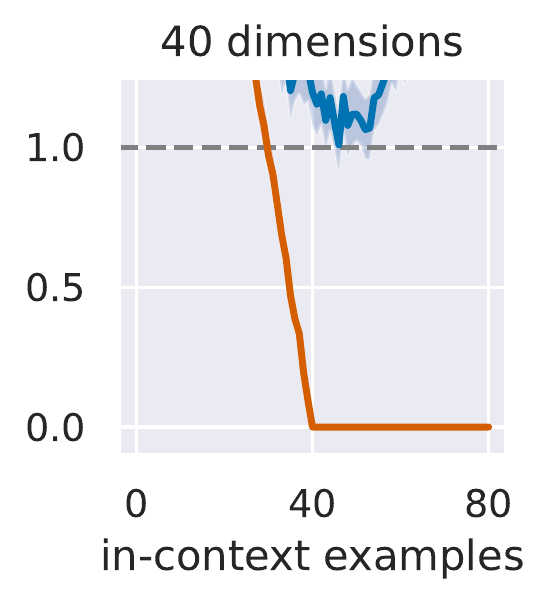}
        \includegraphics[height=.18\textwidth]{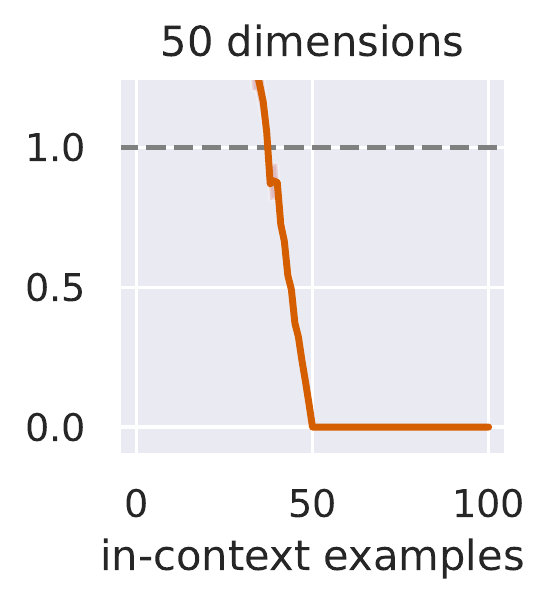}
        \caption{scaled $x$ by a factor of $2$}
        \label{fig:trial_scalex}
    \end{subfigure} \\[.7em]
    \begin{subfigure}[b]{\textwidth}
        \centering
        \includegraphics[height=.18\textwidth]{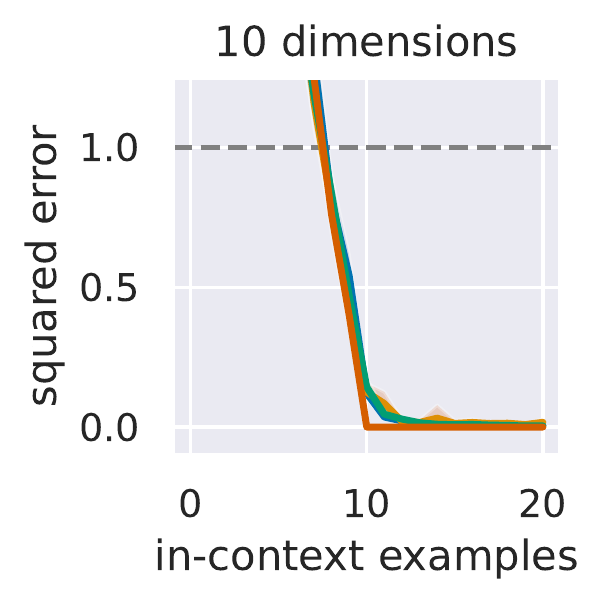}
        \includegraphics[height=.18\textwidth]{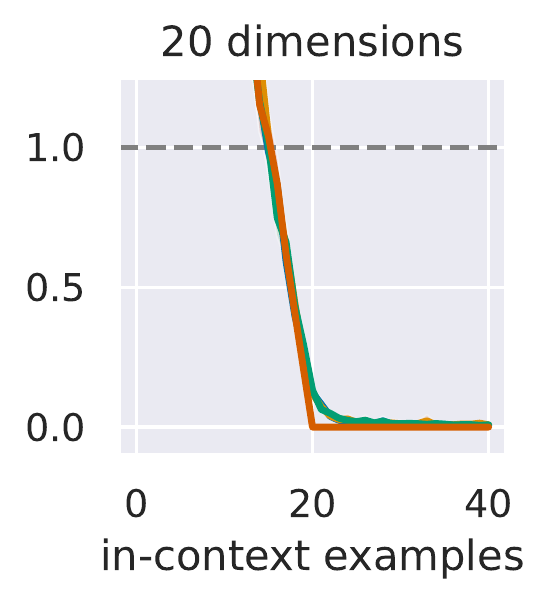}
        \includegraphics[height=.18\textwidth]{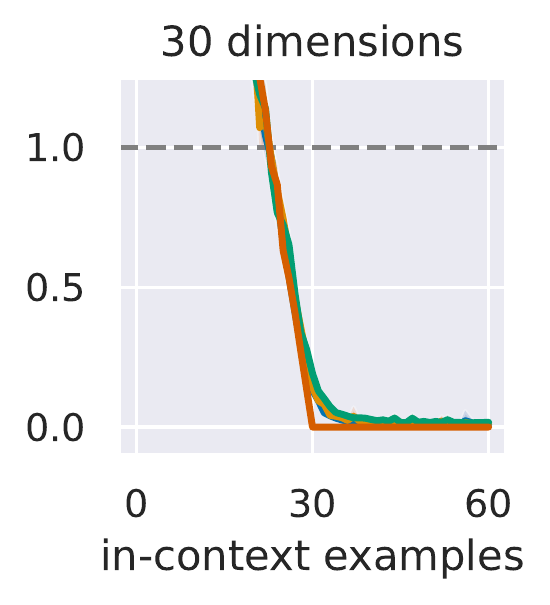}
        \includegraphics[height=.18\textwidth]{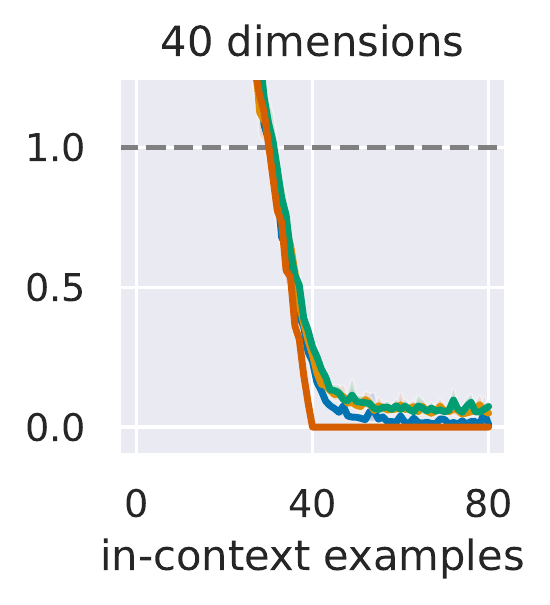}
        \includegraphics[height=.18\textwidth]{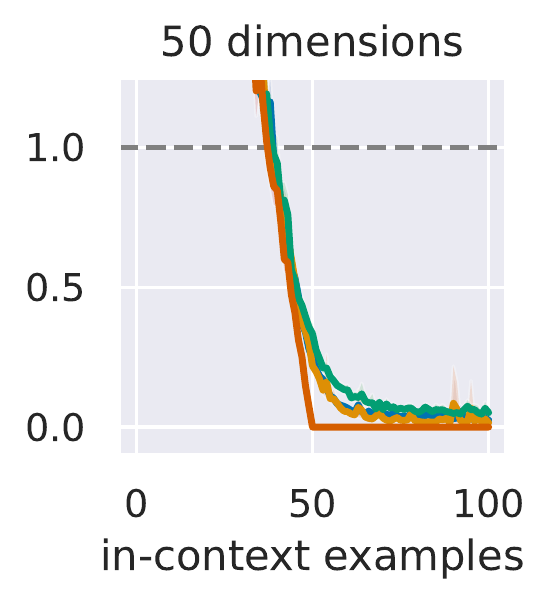}
        \caption{scaled $w$ by a factor of $2$}
        \label{fig:trial_scaley}
    \end{subfigure}
    \caption{(continued) \emph{Errors for models trained with different random seeds.} For each dimension, we train three models with different random seeds and show the corresponding error curves.}
    \label{fig:trials}
\end{figure}

\clearpage
\subsection{Curriculum}
\label{app:curriculum}
\begin{figure}[htp]
    \centering
    \begin{subfigure}[b]{.32\textwidth}
        \includegraphics[height=.75\textwidth]{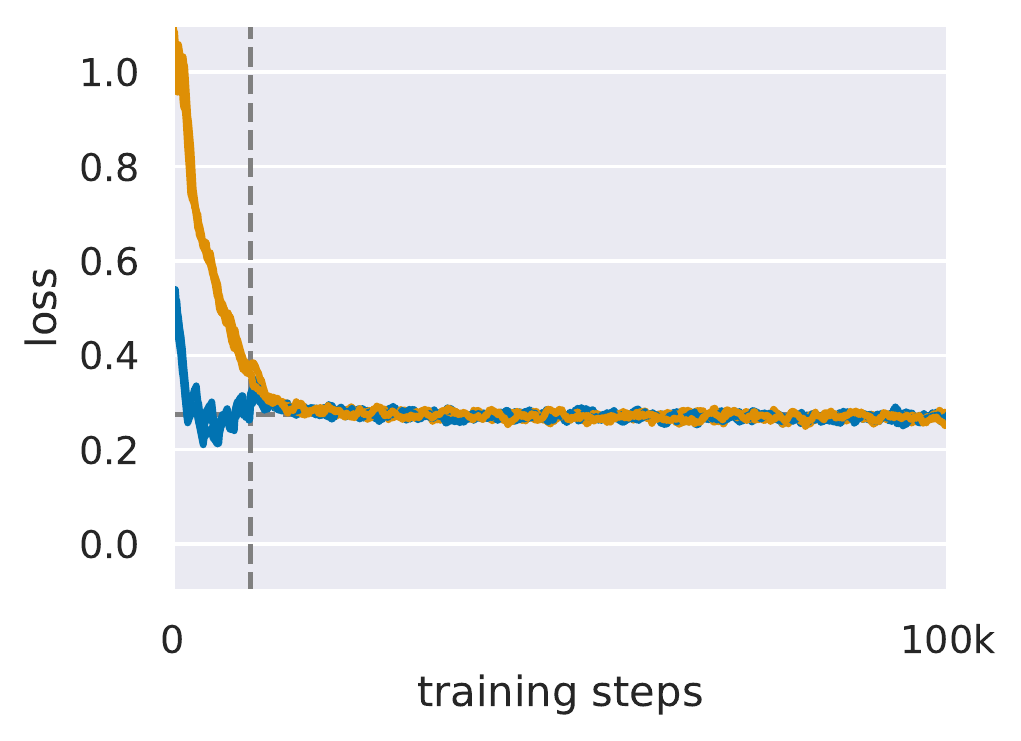}
        \caption{10 dimensions}
        \label{fig:curr_10}
    \end{subfigure}
    \begin{subfigure}[b]{.32\textwidth}
        \includegraphics[height=.75\textwidth]{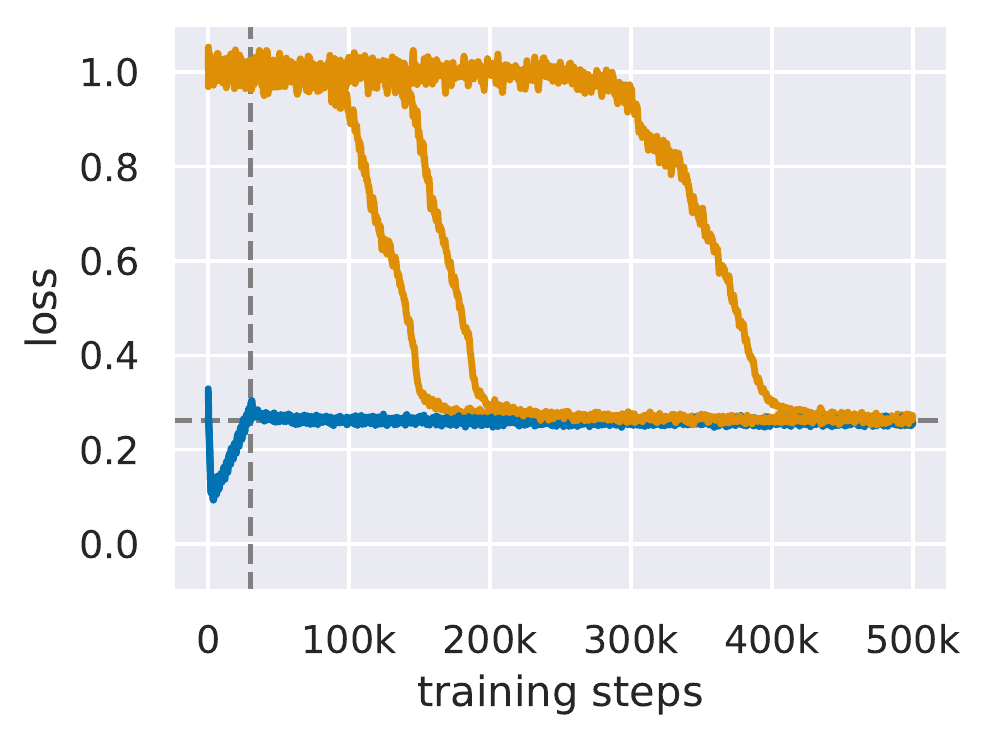}
        \caption{20 dimensions}
        \label{fig:curr_20}
    \end{subfigure}
    \begin{subfigure}[b]{.32\textwidth}
        \includegraphics[height=.75\textwidth]{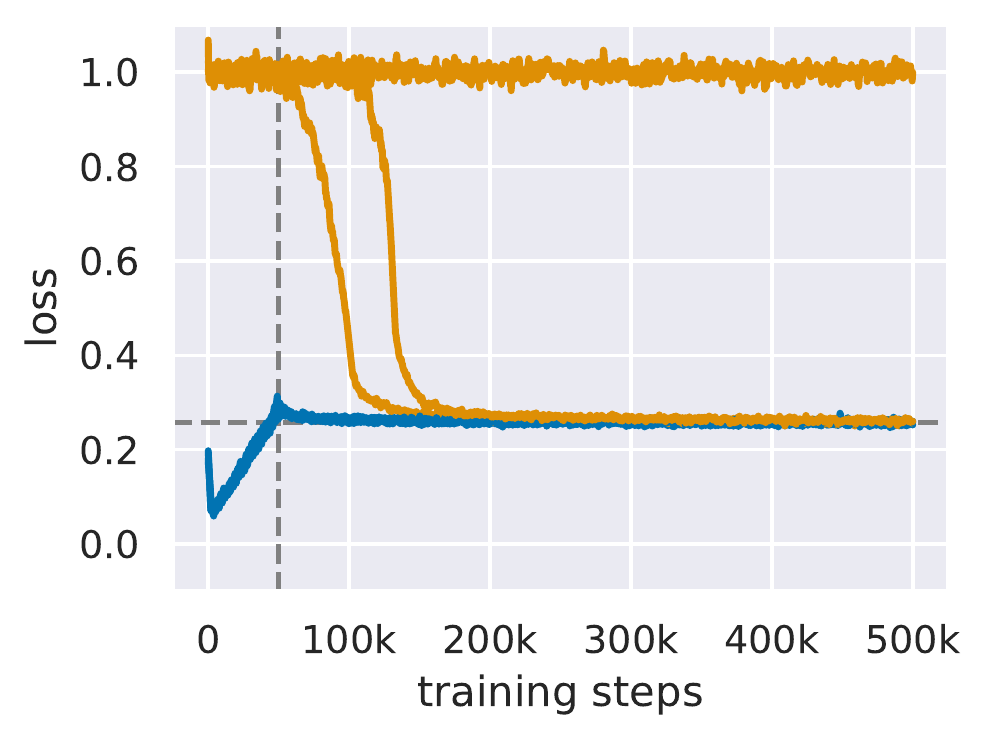}
        \caption{30 dimensions}
        \label{fig:curr_30}
    \end{subfigure}\\[1em]
    
    \begin{subfigure}[b]{.32\textwidth}
        \includegraphics[height=.75\textwidth]{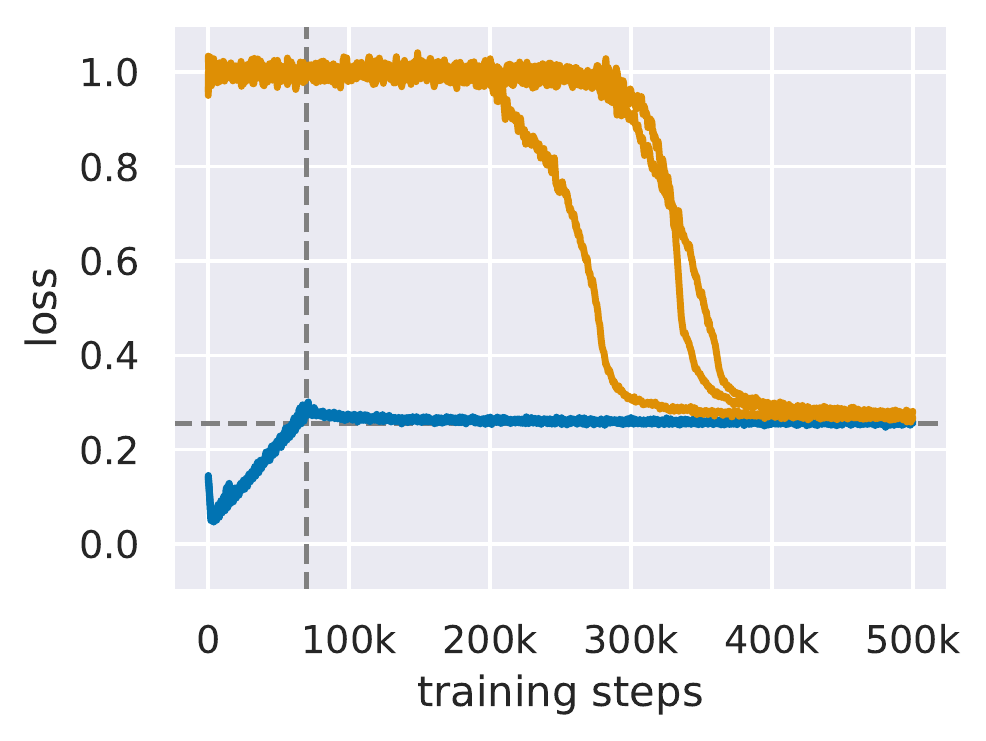}
        \caption{40 dimensions}
        \label{fig:curr_40}
    \end{subfigure}
    \begin{subfigure}[b]{.32\textwidth}
        \includegraphics[height=.75\textwidth]{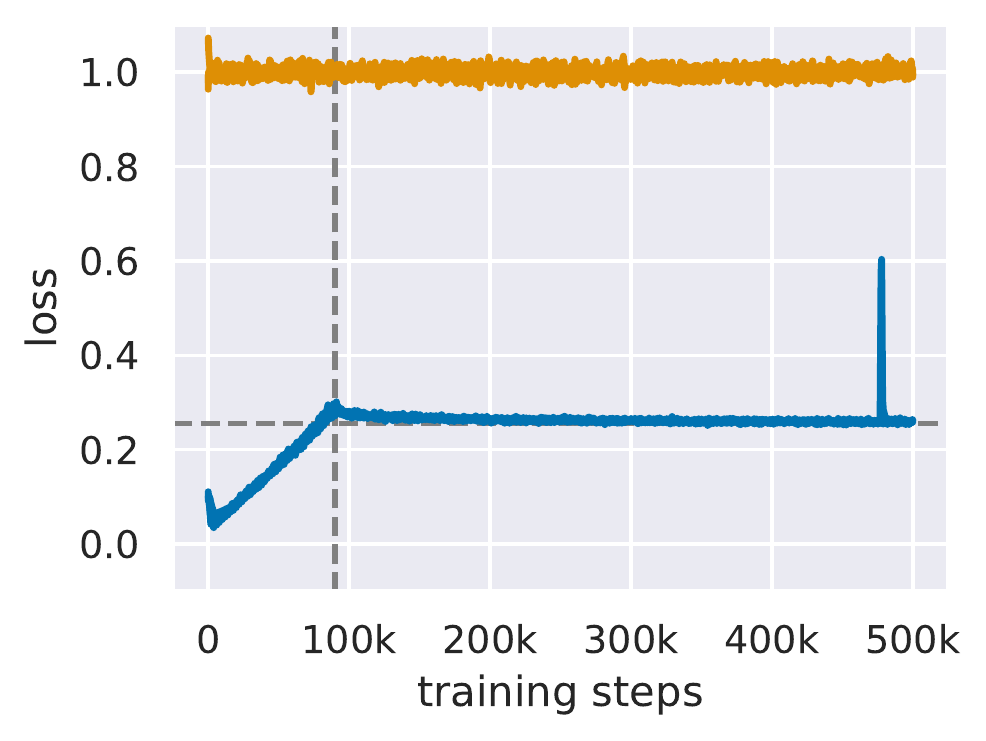}
        \caption{50 dimensions}
        \label{fig:curr_50}
    \end{subfigure}
    \begin{subfigure}[b]{.32\textwidth}
        \centering
        \includegraphics[width=.8\textwidth]{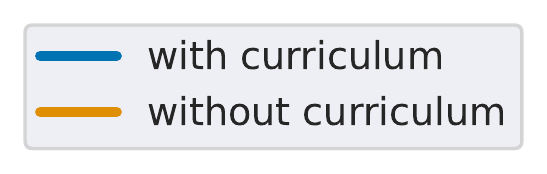}
        \vspace{2cm}
    \end{subfigure}
    \caption{\emph{Loss progression during training with and without curriculum.} For each dimension, we show the loss progression with 3 random seeds each for training with and without curriculum. The vertical dashed line shows the point at which the effective dimension of prompt inputs $d_\text{cur}$ reaches the actual dimension $d$, after which training with and without curriculum have the same prompt distribution. The horizontal dashed line shows the optimum expected loss.
    There is a drastic speedup in training with curriculum. Without curriculum, there is an initial relatively long period where the loss does not decrease. For each dimension, there is a large variance in the length of this period, and the average length seems to increase with dimension.
    }
    \label{fig:curriculum}
\end{figure}

In Figure \ref{fig:curriculum}, we show the training loss of the Transformer model trained to in-context learn linear functions, with and without a curriculum. Specifically, given a random training prompt sequence $(x_1, f(x_1)$, $x_2, f(x_2)$, $\ldots$, $x_{k_\text{cur}}, f(x_{k_\text{cur}}))$, let $\hat{y_i}$ be the model's prediction for the $i^\text{th}$ input (meant to approximate $f(x_i)$). For each such prompt, we consider the loss given by the normalized squared error averaged over all prompt prefixes
\[
\frac{1}{k_{\text{cur}}}  \sum_{i=1}^{k_{\text{cur}}} 
\frac{\left(
\hat{y}_i - f(x_i)
\right)^2}{d}.
\]
At each training step, we plot the loss averaged over a batch of $64$ random prompts. For training with curriculum, $k_{\text{cur}}$ is gradually increased to $2d+1$  as described in Section \ref{app:training}. For training without curriculum $k_{\text{cur}} = 2d+1$ at all times. 

Note that the loss often increases in the beginning as we train the model with curriculum. This is due to a sharp increase in the loss at steps where we increase the effective dimensionality ($d_\text{cur}$) of prompt inputs ($x_i$). There are two reasons for this increase: (i) variance of the target output ($f(x_i) = w^\top x_i$) increases, so even the optimum loss is larger, (ii) the model performance is worse for the prompt inputs with increased effective dimension. After each such step where we increment $d_\text{cur}$, the loss starts to decrease again until the next increment. The overall trend in the loss looks upward when the sharp increase dominates the decrease that follows. Some observations worth highlighting are as follows.

\paragraph{Curriculum drastically speed-ups training.} For functions in 20 or more dimensions, curriculum allows us to train a low-error model often 4 times faster. Moreover, training without curriculum does not always succeed within our training budget (500k steps), e.g., for one run with $d=30$  and \emph{all} runs with $d=50$, the loss does not decrease at all without curriculum.
    
\paragraph{Initial lull without curriculum.}     For training without curriculum, we observe that the loss does not decrease for relatively a long period in the beginning, and starts to decrease sharply thereafter. There is a large variance in the length of this period for any fixed dimension, and  the average length  seems to increase with dimension. This period is almost non-existent for smaller dimensions (e.g., see the plot for $d=10$), and therefore we do not observe such a period while training with curriculum where we start training with inputs lying in a 5 dimensional subspace.


\begin{figure}[htp]
    \centering
    \begin{subfigure}[b]{.32\textwidth}
        \centering
        \includegraphics[height=.85\textwidth]{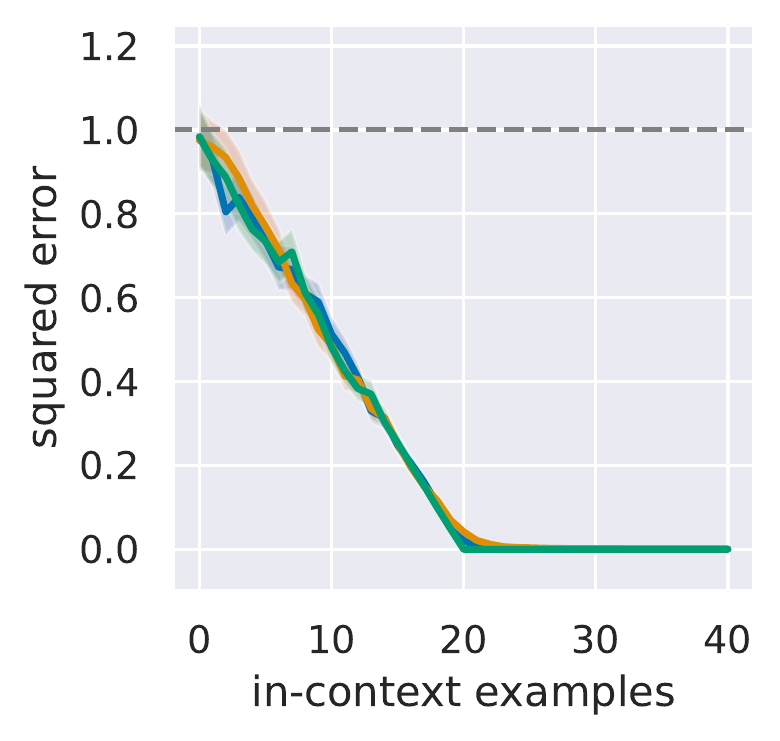}
        \caption{standard}
    \end{subfigure}
    \begin{subfigure}[b]{.32\textwidth}
        \centering
        \includegraphics[height=.85\textwidth]{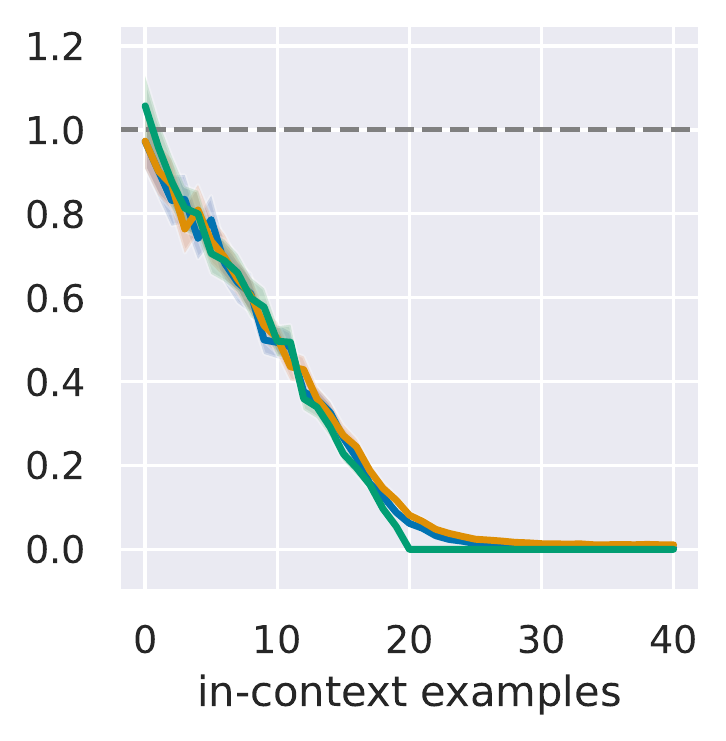}
        \caption{different orthants}
    \end{subfigure}
    \begin{subfigure}[b]{.32\textwidth}
        \centering
        \includegraphics[height=.85\textwidth]{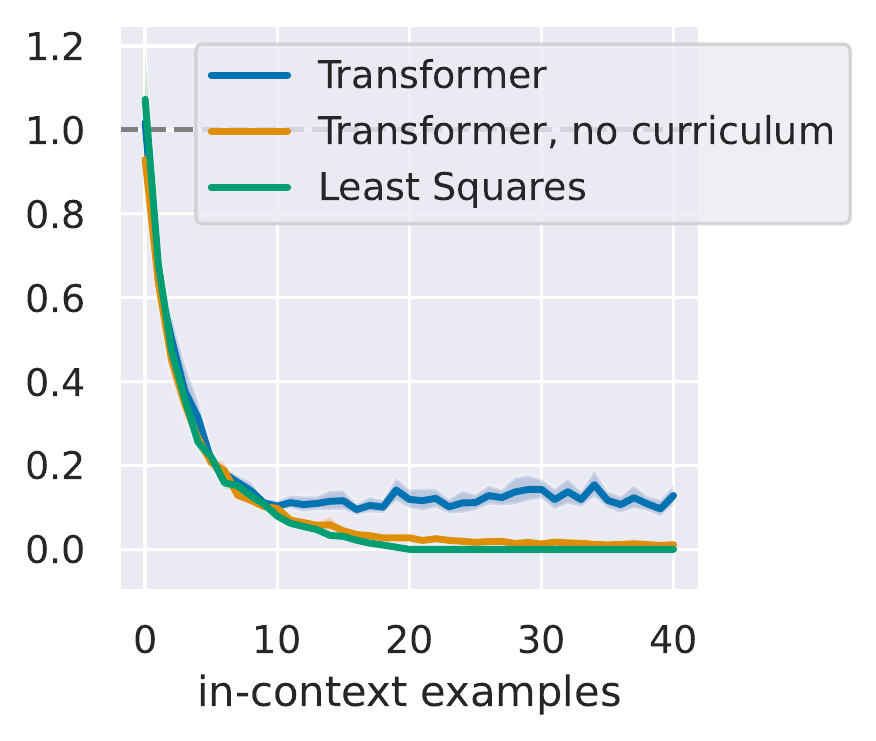}
        \caption{skewed}
    \end{subfigure}
    \caption{\emph{In-context learning performance for models trained with and without curriculum}. We show the  performance for models trained with and without curriculum for in-context learning linear functions ($d=20$). We did not observe any major qualitative difference in performance between the two settings in most cases. One exception is the case with skewed covariance where the model trained without curriculum does better. }
    \label{fig:curr_ood}
\end{figure}

\paragraph{Curriculum does not affect final performance significantly.} For our core setting  ($d=20$), where we are able to train the model to  low error even without curriculum, we do not observe any qualitative differences in the error in most cases (both with and without distribution shifts). One exception is the case with skewed covariance, where the model trained without curriculum seems to do slightly better. We plot the error curves for the standard, different orthants and skewed covariance cases in Figure \ref{fig:curr_ood}.

\clearpage
\subsection{Effect of number of distinct prompts/functions seen during training}
\label{app:distinct_funcs}
Here, we investigate the effect of amount of training data required for in-context learning linear functions.

\begin{figure}[htp]
    \centering
    \includegraphics[height=.25\textwidth]{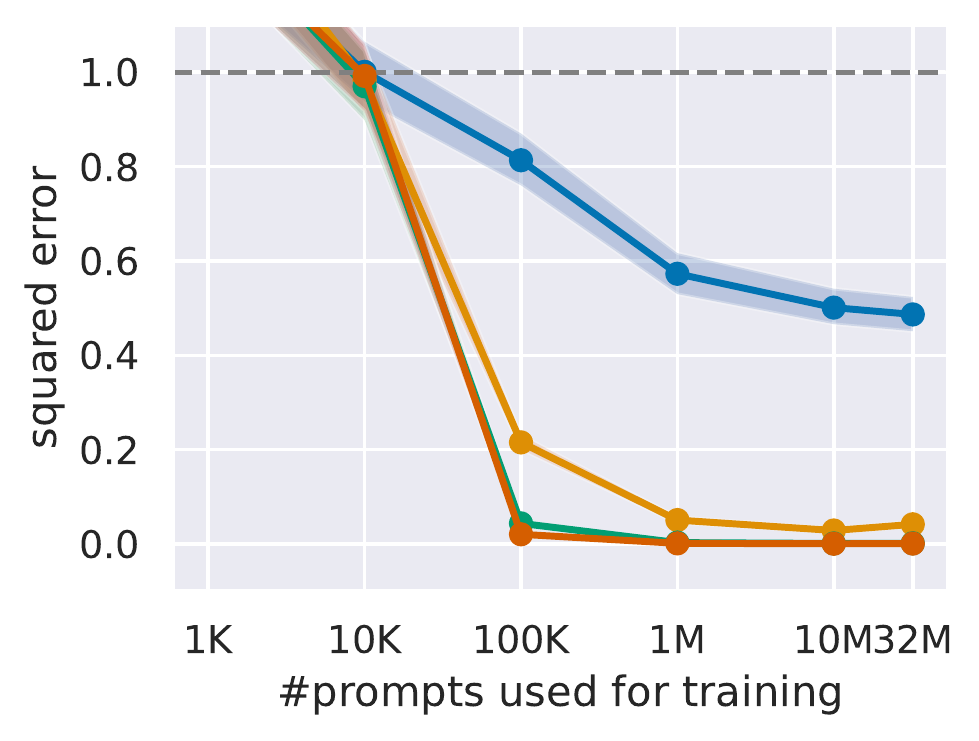}
    \includegraphics[height=.25\textwidth]{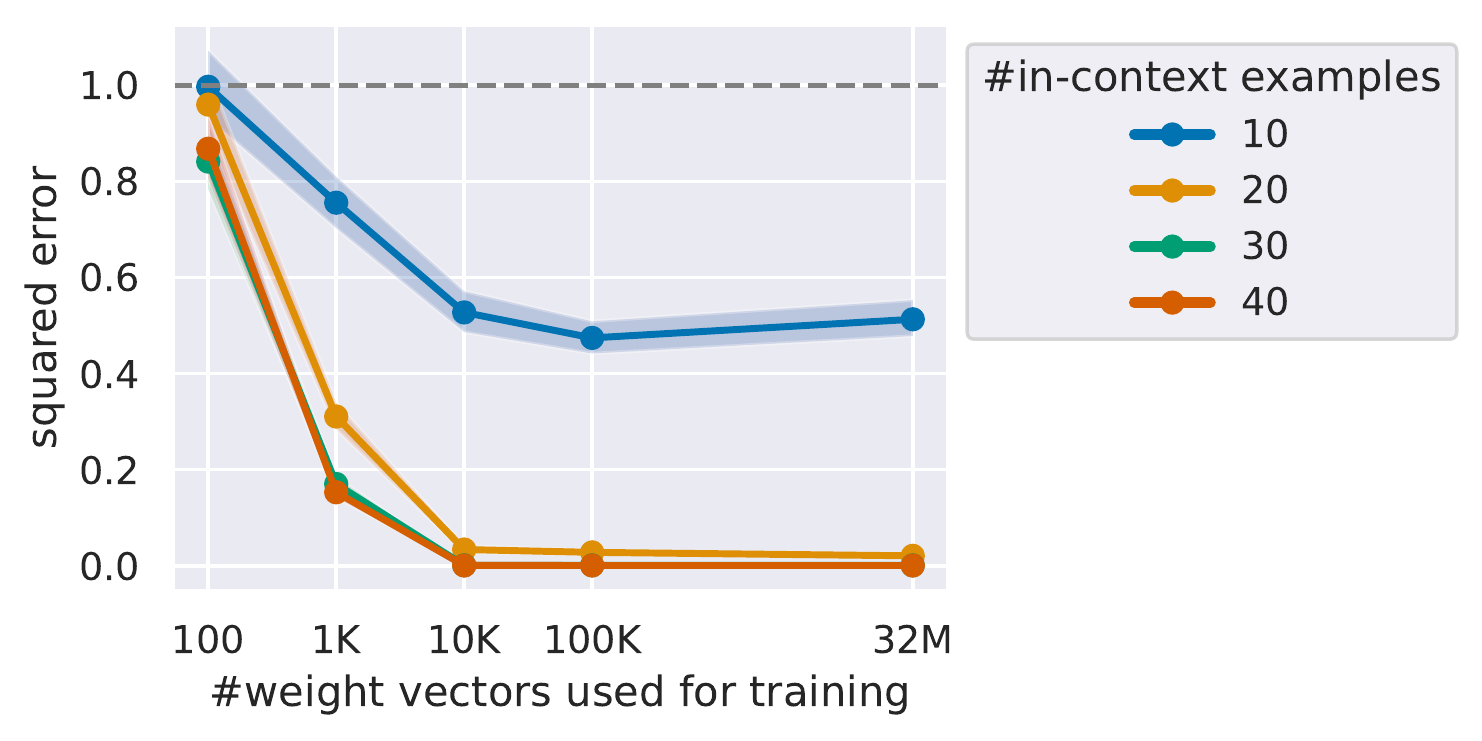}
    \caption{\emph{Effect of number of distinct prompts/functions seen during training.} We plot the squared error for models trained to in-context linear functions, as we increase the number of 
    distinct prompts and distinct weight vectors (equivalently, distinct functions) seen during training. (Note that 32M corresponds to the unrestricted model where we sample fresh prompts at each training step.) The models are able to achieve error close to that of the unrestricted model with  $1M$ distinct prompts or $10k$ distinct weight vectors.
    }
    \label{fig:finite}
\end{figure}

First, we consider the effect of number of distinct prompts encountered during training. For this, we create a set $S_p$ of $n_p$ randomly generated prompts, where each prompt in $S_p$ is generated by sampling a weight vector and prompt inputs from $N(0, I_d)$. We generate random prompts during training by sampling uniformly from this set. As before, we train the model for $500k$ steps with a batch size of $64$. We observe (see Figure \ref{fig:finite}) that a model trained with $n_p = 100k$ is able to achieve non-trivial error and a model trained with $n_p = 1M$ achieves error close to that of the unrestricted model (trained with $32M$ distinct prompts). Recall that with curriculum learning, we zero out some of the coordinates of prompt inputs in the beginning of training, which will increase the total number of prompts the model sees during training.
Therefore we do not use curriculum learning for this study to avoid inflating the number of distinct prompts seen during training.

Second, we consider the effect of number of distinct weight vectors  (equivalently, distinct functions) encountered during training. For this, we create a set $S_w$ of $n_w$ weight vectors where each weight $w$ is drawn i.i.d. from $N(0, I_d)$. To generate a training prompt, $(x_1, w^\top x_1, \ldots, x_{k}, w^\top x_{k})$, we draw prompt inputs ($x_i$s) i.i.d. from $N(0, I_d)$ as in the unrestricted setting,  and sample $w$ uniformly at random from $S_w$. Thus while we sample from a finite set of weight vectors, we sample fresh inputs at each step. As before, we train the model for $500k$ steps with a batch size of $64$. Here, we observe (see Figure \ref{fig:finite}) that the model trained with as few as $10k$ distinct weight vectors achieves error close to the unrestricted model (trained with $32M$ distinct functions). We use curriculum learning for this study as in our standard setting. Recall that with curriculum learning, we only zero out some coordinates of prompt inputs in the beginning, so this does not change the number of distinct weight vectors seen by the model during training.

\clearpage
\subsection{Can memorization explain model performance?} 
\label{app:memorization}
In Section \ref{sec:linear_functions_1}, we discussed that memorization of prompts seen during training cannot explain model performance. 
This is because the probability of the model encountering a training prompt similar to the one used for testing is astronomically low---the prompt inputs alone lie in a 800-dimensional space when predicting with $2d$ in-context examples ($d=20$).

Moreover, even considering the possibility that the model encountered a similar \emph{weight vector} during training cannot explain its performance.
Let $S_w$ be the set of weight vectors used to generate training prompts. At inference time, given a prompt with in-context examples generated using a weight vector $w_\star$, suppose the model is somehow able to find the best weight vector $\hat{w}$ in $S_w$ minimizing the normalized squared error on query inputs:
\begin{align*}
\hat{w} &= \argmin_{w \in S_w} \ \mathds{E}_{x_{\text{query}} \sim N(0, I_d)} \left[\frac{(w^\top x_{\text{query}} - {w_\star}^\top x_{\text{query}}  )^2}{d} \right] \\
&= \argmin_{w \in S_w} \ \frac{ \lVert w - w* \rVert ^2_2}{d}
\end{align*}
Taking expectation over the weight vector $w_\star$, we get the expected normalized squared error of the model (with respect to randomly drawn in-context examples and query inputs):
\[
\mathds{E}_{{w_\star \sim N(0, I_d)}} \left[ \min_{w \in S_w} \ \frac{ \lVert w - w_\star \rVert ^2_2}{d} \right].
\]
To empirically estimate this quantity, we sample $n_w$ weight vectors from $N(0, I_d)$ (with $d = 20$) that form the set $S_w$, and $500$ weight vectors from $N(0, I_d)$ to estimate the outer expectation. We do this 20 times, freshly sampling the $500$ weight vectors and the vectors comprising $S_w$ each time, and compute the mean of the 20 estimates obtained. 
When $n_w = 32M$ (number of weight vectors encountered in our standard training setup), we get a mean of 0.216 (standard deviation 0.004). However, our model is able to achieve an expected error of less than $0.001$ for prompts with $2d$ in-context examples.
Similarly, when $n_w = 10,000$, we get a mean of 0.505 (standard deviation 0.006), while a model trained on prompts generated using $10,000$ distinct weight vectors is able to achieve a much smaller error (see Figure \ref{fig:finite}). 

Thus we can conclude that the model cannot be relying on memorization
of the training prompts or weight vectors, and is encoding a more sophisticated algorithm capable of in-context
learning linear functions that are very different from those seen during training.

\end{document}